\documentclass[10pt,twocolumn,letterpaper]{article}

\usepackage[pagenumbers]{cvpr} % To force page numbers, e.g. for an arXiv version

\usepackage[dvipsnames]{xcolor}

\definecolor{cvprblue}{rgb}{0.21,0.49,0.74}
\usepackage[pagebackref,breaklinks,colorlinks,citecolor=cvprblue]{hyperref}

\def\paperID{*****} % *** Enter the Paper ID here
\def\confName{****}
\def\confYear{****}

\def\Name{PPNet}

\title{Deep Planar Parallax for Monocular Depth Estimation}
\author{Haoqian Liang$^1$ ~ ~ Zhichao Li$^2$ ~ ~ Ya Yang$^1$\thanks{Corresponding author.} ~ ~ Naiyan Wang$^2$ \\
\normalsize $^1$Beijing University of Posts and Telecommunications ~ ~ $^2$TuSimple \\
{\tt\small \{lianghq, yangya\}@bupt.edu.cn, \{leeisabug, winsty\}@gmail.com}
}

\begin{document}
\maketitle

%%%%%%%%% ABSTRACT
\begin{abstract}
Recent research has highlighted the utility of Planar Parallax Geometry in monocular depth estimation. However, its potential has yet to be fully realized because networks rely heavily on appearance for depth prediction. Our in-depth analysis reveals that utilizing flow-pretrain can optimize the network's usage of consecutive frame modeling, leading to substantial performance enhancement. Additionally, we propose Planar Position Embedding (PPE) to handle dynamic objects that defy static scene assumptions and to tackle slope variations that are challenging to differentiate. Comprehensive experiments on autonomous driving datasets, namely KITTI and the Waymo Open Dataset (WOD), prove that our Planar Parallax Network (PPNet) significantly surpasses existing learning-based methods in performance.

\end{abstract}
%%%%%%%%% BODY TEXT
%%%%%%%%% BODY TEXT

\section{Introduction}
\label{sec:intro}

\begin{figure}[tb]
  \centering
  \includegraphics[width=1\linewidth]{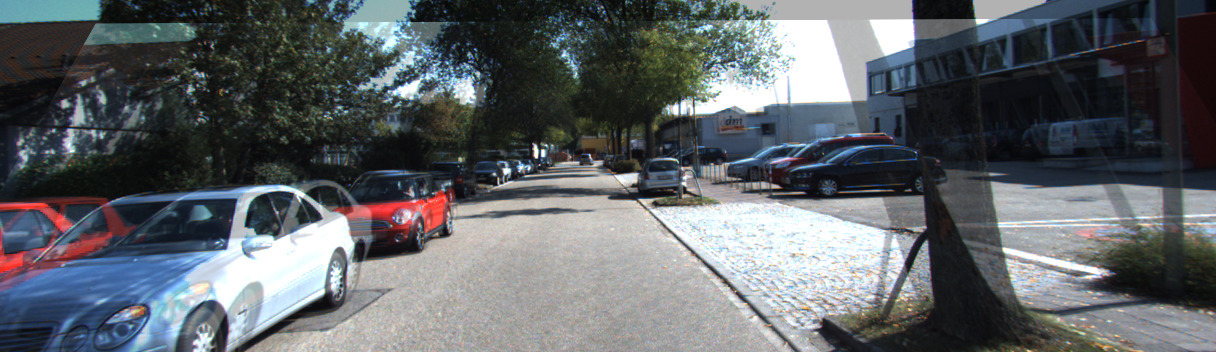}
  \includegraphics[width=1\linewidth]{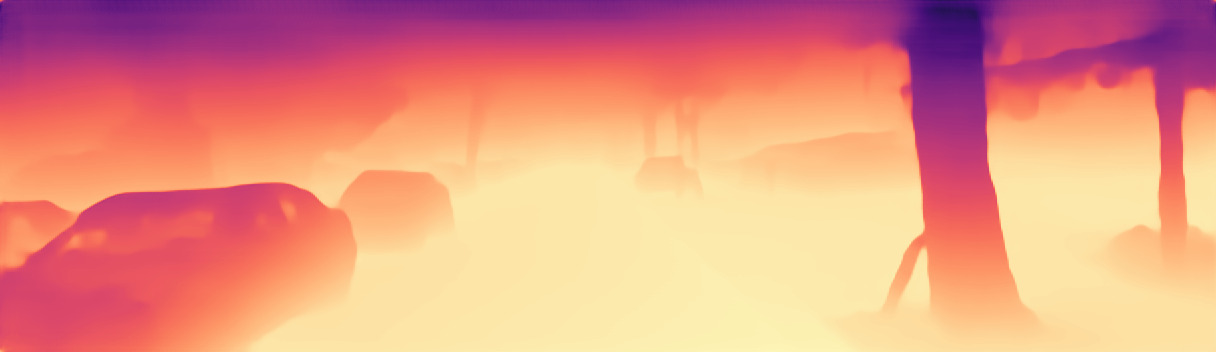}
  \includegraphics[width=1\linewidth]{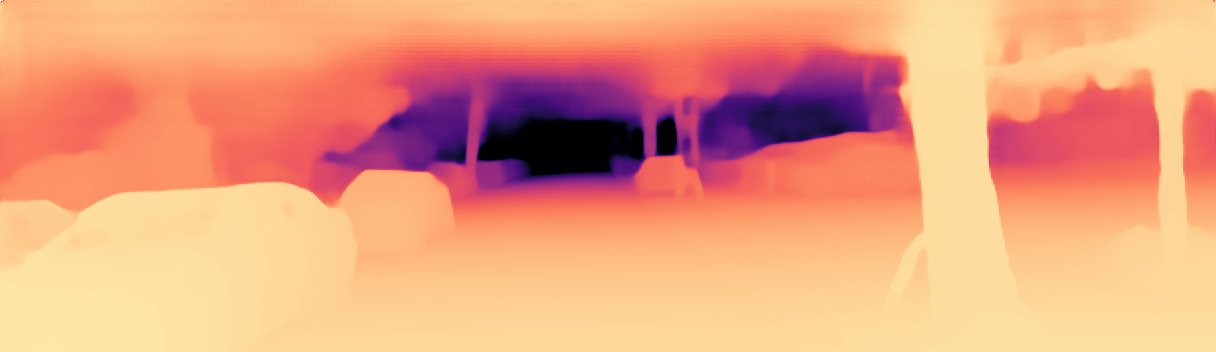}
  \caption{The first row shows the current frame superimposed by the homography warping of the previous frame based on planar parallax geometry, in which if the pixel is higher than the plane, the shearing is heavier. The second row shows the $\gamma$ map predicted by our model using the two images in the first row. Furthermore, the last row displays our depth estimation results based on the predicted $\gamma$ map.}
  \label{fig:idea}
\end{figure}

Depth perception is at the core of 3D computer vision tasks. Recent studies have shown that learning-based methods~\cite{bian2019, li2021deepi2p, aoki2019pointnetlk, bian2021, donati2020deep, luo2020consistent, yin2021virtual, kopf2021robust} can estimate geometric properties such as depth or pose in a single forward pass by taking advantage of the data-driven approach.
For the depth estimation problem in autonomous driving systems, several crucial assumptions exist: objects and obstacles are placed on a known road plane, and multiple consecutive frames with known ego motion between them are available. These assumptions naturally lead to the planar parallax (P+P) geometry~\cite{shashua1994relative, sawhney19943d}. In particular, the P+P geometry shows that, after the alignment of consecutive frames to the reference plane, the 3D scene structure can be derived from the residual pixel displacements caused by the camera's motion. Consequently, ~\cite{yuan2021monocular, xing2022joint} follows the P+P framework, using a neural network to predict pixel-wise $\gamma$, the ratio of height to depth, rather than a depth map. Nevertheless, our analysis indicates that the performance boosts achieved by these methods are somewhat restrained. A deeper dive revealed that these networks couldn't fully harness the benefits of consecutive frame inputs combined with planar warping.

This paper endeavors to bridge this gap, targeting further performance enhancements. Inspired by Eq.~\ref{eq:res2gamma}, we introduce flow prediction to help establish connections between consecutive frames. Our experiments show that using flow as an additional input can effectively enhance performance, but this requires a separate network to predict the flow. Furthermore, we find that a good flow pre-training can effectively preserve the connections between frames and bring significant performance improvements without additional computational overhead. 
To address the challenge of moving objects that violate the assumption of P+P and slope change that are hard to differentiate, we propose Planar Position Embedding (PPE), which informs the network of the relative position of each pixel with respect to the reference plane. We also incorporate a single view path with learnable weights following DfM~\cite{wang2022dfm}, which helps the network balance the importance of geometry and statistics.

To summarize, our main contributions are three-fold:
\begin{itemize}
  \item We propose a deep planar parallax depth estimation network, which effectively incorporates planar parallax geometry pipeline and data-driven methods. For the first time, we demonstrate that optical flow pre-training is crucial for utilizing connections between consecutive frames.
  \item We also propose a Planar Position Embedding, which introduces the pixel position related to the reference plane into the network, which can effectively compress the error space.
  \item We test our proposed method on large-scale self-driving datasets, KITTI~\cite{geiger2012we} and Waymo Open Dataset (WOD)~\cite{sun2020scalability}. Extensive results demonstrate that our method beats the state-of-the-art methods by a large margin. %about 28.8\%.
\end{itemize}

\section{Related Work}
\label{sec:related_work}

\subsection{Monocular Depth Estimation}

Eigen~\etal~\cite{eigen2014depth} are the first to use Convolutional Neural Networks (CNN) for the monocular depth estimation task. They propose a method that combines local and global information to predict depth from a single image. Since then, depth estimation methods based on neural networks have made significant progress~\cite{fu2018deep, yin2019enforcing, lee2019big}.
Recently, Vision Transformers have been introduced to depth estimation tasks. Adabin~\cite{bhat2021adabins} uses a transformer-based architecture to divide the depth range into bins. NeWCRFs~\cite{yuan2022newcrfs} further improves depth estimation accuracy by adopting a Swin Transformer~\cite{liu2021Swin} as the encoder and a fully-connected Conditional Random Field (CRF) as the decoder. This new architecture has shown considerable improvement in depth estimation performance~\cite{agarwal2023attention, liu2022va, piccinelli2023idisc, yang2023gedepth}.
While some works focus on learning from a single image, others concentrate on monocular videos.
~\cite{gu2020cascade, bae2022multi, teeddeepv2d, long2021multi} use geometric cues between frames directly to estimate depth. \cite{zhu2023lighteddepth, li2023learning} enable the multi-frame depth to benefit more from single-frame depth. However, these methods often rely on future frames or suffer from low speed, which makes them unsuitable for real-time prediction in scenarios such as autonomous driving.

\subsection{Flow Estimation}
Starting from the FlowNet series~\cite{dosovitskiy2015flownet, flownet2},  end-to-end optical flow networks have shown their superiority in flow estimation tasks. After that, PWC-Net~\cite{pwcnet} introduces the pyramid(P), warping(W), and cost volume(C) into network design and significantly improves the performance. MaskFlowNet~\cite{zhao2020maskflownet} resolves the occluded areas during warping with a self-learned occlusion mask. The success of RAFT~\cite{raft} lies in the iterative refinement of the cost volumes. GMFlow~\cite{xu2022gmflow} first uses a transformer in flow estimation. Besides the supervised methods, photometric loss-based unsupervised flow~\cite{janai2018unsupervised,meister2018unflow, zhong2019unsupervised,wang2018occlusion,liu2020learning} attracts researchers' attention, but there still exists a performance gap compared with supervised methods. Optical flow is a fundamental low-level task that plays an essential role in many downstream problems. We use it as a geometric prior so that the network can more efficiently use the geometric information to estimate the depth.

\subsection{Planar Parallax Methods}
The planar parallax methods are first proposed in the mid-90s~\cite{sawhney19943d, shashua1994relative, sawhney1994motion}. This method decomposes motion between multiple frames into planar homography and residual pixel displacement from unaligned pixels by choosing a reference plane~\cite{irani1996parallax}. 
The planar homography is obtained through camera z-axis translation and plane normal vector. The residual pixel displacements can be obtained by feature-matching methods such as optical flow. In this way, P+P geometry significantly reduces the estimation space. Many applications~\cite{sawhney1994simplifying,yuan2007detecting, cross1999parallax, irani2002direct} depend on a plane in the scene. For instance, the height of points from the ground plane is critical in robotics~\cite{baehring2005detection, lourakis1999using}. Additionally, the planar parallax has been used in various applications such as quantifying image stitching~\cite{jung2021quantitative} and camera calibration~\cite{vaish2004using}.
However, finding a suitable reference plane is a critical challenge in these methods. Naturally, in driving scenes, the road plane can easily be extracted from LiDAR points or high-precision maps. Taking these advantages, Yuan~\etal~\cite{yuan2021monocular} proposes a new solution combining traditional planar parallax geometry with a deep neural network for road environments. Xing~\etal~\cite{xing2022joint} use the dense structure information provided by P+P as depth hints. During our experiments, we find that previous learning-based P+P methods do not fully utilize the geometry structure. In this paper, we will pursue this goal.

\section{Method}
\label{sec:method}

\subsection{Preliminary}
\label{subsec:PP} 

This paper's derivation of the Planar Parallax Geometry primarily follows ~\cite{yuan2021monocular}. Here, we will briefly explain the formulas we have used, while detailed derivations can be found in the appendix. The ratio of height to depth $\gamma$ plays an important role in our model:
\begin{equation}
	\gamma = \frac{h}{z},
\end{equation}
where $h$ and $z$ denote the height and depth of a pixel, respectively. Following ~\cite{yuan2021monocular}, our model predict $\gamma$ instead of depth $z$. 

We define $\mathbf{T}=(t_x, t_y, t_z)^T$ as the translation vector between the two camera views. Let $\mathbf{t}=\mathcal{K}\mathbf{T}$, where $\mathcal{K}$ is intrinsic matrix of the camera.

\begin{figure}[tb]
  \centering
  \includegraphics[width=1\linewidth]{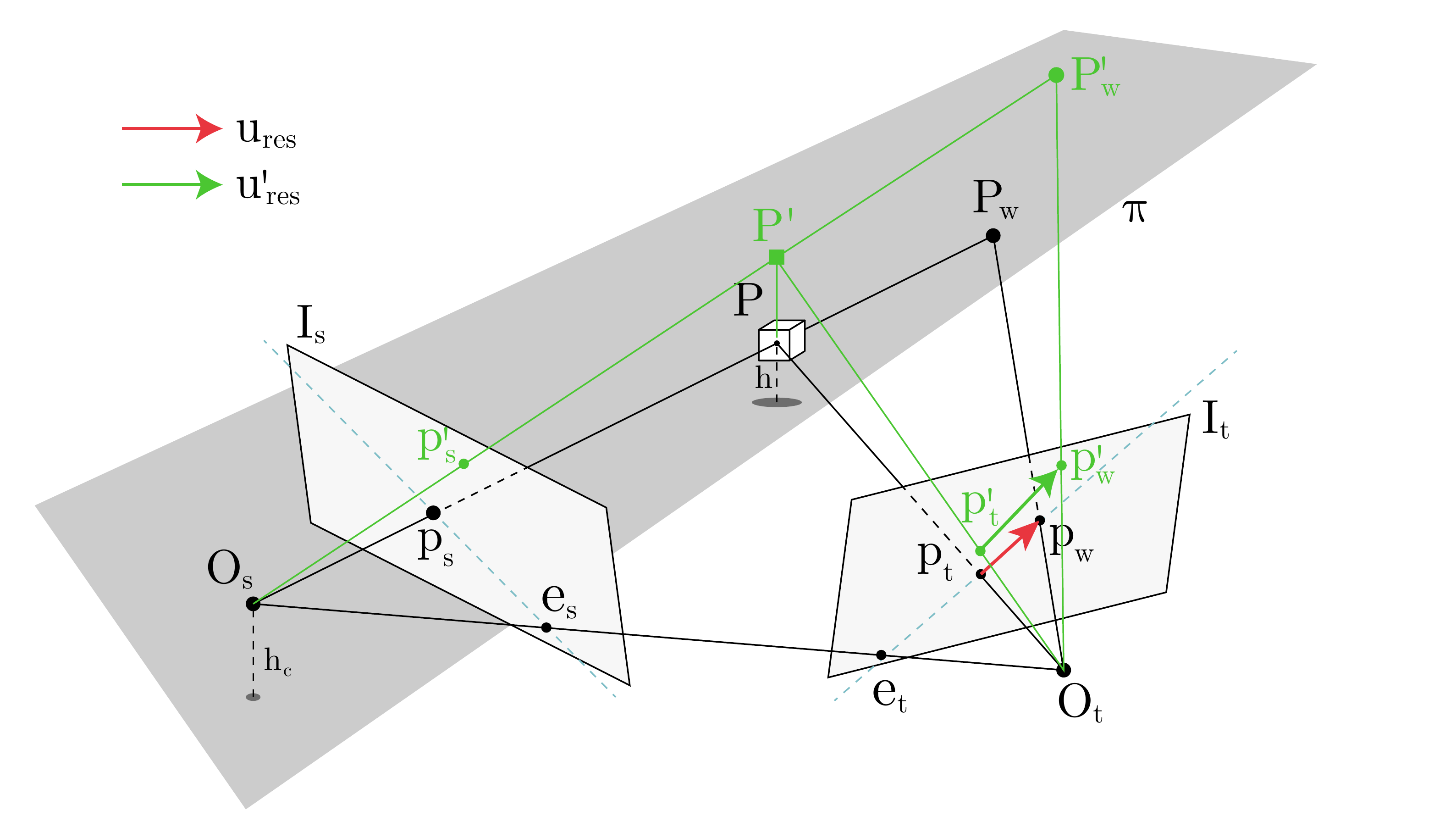}
  \caption{The illustration of planar parallax geometry.}
  \label{fig:pp}
\end{figure}

In Fig.~\ref{fig:pp}, we present the geometry visually. 
$\mathbf{u}_{res}=\mathbf{p}_w-\mathbf{p}_t$ is the residual flow. Following the mathematical derivation in ~\cite{yuan2021monocular, irani1996parallax}, when $t_z \neq 0$, we can obtain:
\begin{equation}
\label{eq:res}
  \begin{split}
  \mathbf{u}_{res} = \frac{\gamma\frac{t_z}{h_c}}{1 - \gamma\frac{t_z}{h_c}}(\mathbf{p}_t - \mathbf{e}_t).
  \end{split}
\end{equation}

The epipole $\mathbf{e}_t=\frac{1}{t_z}\mathbf{t}$ in the target view represents the point that remains stationary after the image warping. $h_c$ is the height of the camera. 

Fig.~\ref{fig:pp} illustrates a scenario where $\mathbf{P}^{'}$ is a 3D point located higher than $\mathbf{P}$, resulting in a larger residual flow $\mathbf{u}_{res}^{'}$ compared to $\mathbf{u}_{res}$ for points with the same depth $z$. Thus, the amount of residual flow is a key geometric clue for determining 3D structure.

From Eqn.~\ref{eq:res},  we can infer that the residual flow always moves towards or away from the epipole and has a strong correlation with $\gamma$. We can also convert Eqn.~\ref{eq:res} to
\begin{equation}
\label{eq:res2gamma}
	\gamma = \frac{h_c}{t_z(1 + \frac{\mathbf{p}_t - \mathbf{e}_t}{\mathbf{u}_{res}})}.
\end{equation}

Except for the relationship with the residual flow, $\gamma$ can also be used for 3D reconstruction. 
Following the mathematical derivation in ~\cite{yuan2021monocular, irani1996parallax}, we can obtain an important formula discussed in the proposed Planar Position Embedding(Sec.~\ref{sec:model}):

\begin{equation} \label{eq:nkp}
	\vec{\mathbf{N}}^T(\mathcal{K}^{-1}\mathbf{p}_t) = \frac{h_c - h}{z},
\end{equation}
where $\vec{\mathbf{N}}^T$ is the normal of the plane.

Eqn.~\ref{eq:nkp} can be finally transformed into
\begin{equation}	\label{eq:gamma2Z}
	z = \frac{h_c}{\gamma + \vec{\mathbf{N}}^T(\mathcal{K}^{-1}\mathbf{p}_t)}.
\end{equation}

We could use it to convert predicted $\gamma$ to depth results given the plane and camera height above the plane. Compared with epipolar geometry, the superiority of the P+P geometry is twofold: First, as shown in the first row in Fig.~\ref{fig:idea}, the road pixel in the image is aligned without disparity after warping by road homography. It saves the network capacity for estimating the depth of the ground. Second, Eqn.~\ref{eq:res2gamma} is only affected by $t_z$, thus removing the dependence on rotation and reducing the errors caused by ego-motion noise.

\begin{table}[tb]
    \setlength{\tabcolsep}{0.2mm}{
	\begin{center}  
	\small
	\begin{tabular}{lccccccc}  
	\hline  
	Backbone & Method & Frame & Abs Rel$\downarrow$ & Sq Rel$\downarrow$ & RMSE$\downarrow$ & $\delta_1$$\uparrow$    \\  
    \hline
	GMFlow~\cite{xu2022gmflow} & - & 1 & 0.079 & 0.482 & 3.517 & 0.913	\\
	GMFlow~\cite{xu2022gmflow} & - & 2 & 0.077 & 0.445 & 3.494 & 0.915	\\
	GMFlow~\cite{xu2022gmflow} & FI & 2 & 0.062 & 0.398 & 2.976 & 0.948	\\
	GMFlow~\cite{xu2022gmflow} & FP & 1 & 0.064 & 0.329 & 3.036 & 0.939 \\
	GMFlow~\cite{xu2022gmflow} & FP & 2 & \textbf{0.047} & \textbf{0.218} & \textbf{2.362} & \textbf{0.967}	\\
	\hline
	MaskFlownet~\cite{zhao2020maskflownet} & - & 1 & 0.072 & 0.422 & 3.538 & 0.920	\\
	MaskFlownet~\cite{zhao2020maskflownet} & - & 2 & 0.070 & 0.392 & 3.384 & 0.928	\\
	MaskFlownet~\cite{zhao2020maskflownet} & FI & 2 & 0.053 & 0.234 & 2.644 & 0.958	\\
	MaskFlownet~\cite{zhao2020maskflownet} & FP & 1 & 0.066 & 0.376 & 3.178 & 0.936 \\
	MaskFlownet~\cite{zhao2020maskflownet} & FP & 2 & \textbf{0.047} & \textbf{0.233} & \textbf{2.429} & \textbf{0.964}	\\
	\hline
	ARFlow~\cite{liu2020learning} & - & 1 & 0.085 & 0.513 & 3.814 & 0.896	\\
	ARFlow~\cite{liu2020learning} & - & 2 & 0.081 & 0.516 & 3.687 & 0.908	\\
	ARFlow~\cite{liu2020learning} & FI & 2 & 0.066 & 0.540 & 3.034 & 0.947 \\
	ARFlow~\cite{liu2020learning} & FP & 1 & 0.070 & 0.398 & 3.320 & 0.928 \\
	ARFlow~\cite{liu2020learning} & FP & 2 & \textbf{0.054} & \textbf{0.260} & \textbf{2.651} & \textbf{0.957}	\\
	\hline
	\end{tabular}
	\end{center}
	}
	\caption{\label{tab:flow_result} Results using simply modified flow models. FP means flow pretrain, FI means adding flow as input.}
\end{table}

\begin{figure*}[tb]
  \centering
  \includegraphics[width=1\linewidth]{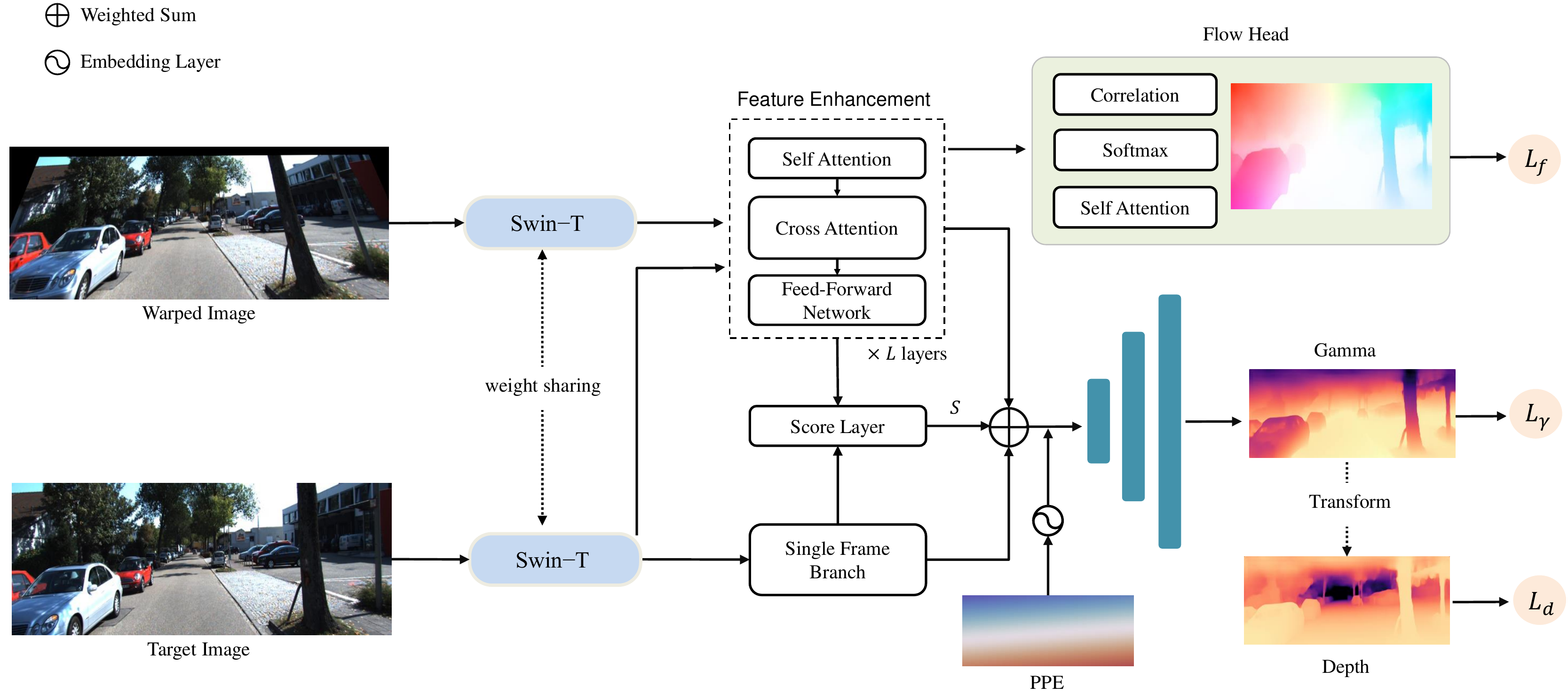}
  \caption{Overview of the Proposed Planar Parallax Network. Given two plane-aligned images, we first extract features using a Swin-Tiny backbone. We then divide the network into two streams: a flow branch and a single frame branch. In the flow branch, we use the Feature Enhancement module and Flow Head from GMflow~\cite{xu2022gmflow}. The Flow Head and flow loss $L_f$ are only used for flow pre-training. The Single Frame Branch is a simple convolutional network that is fused with the flow branch using a weighted sum (see Eqn.~\ref{eq:fuse}). The network is supervised by gamma loss $L_{\gamma}$ and depth loss $L_d$.}
  \label{fig:network}
\end{figure*}

\subsection{Planar Parallax Network} 
\label{sec:model}
\noindent
\textbf{Learning from Flow}
As demonstrated in Eqn.~\ref{eq:res2gamma}, we can derive $\gamma$ from the residual flow obtained from plane-aligned images, given the camera height and translation along the forward axis. The flow can then reconstruct the height and depth information.
Based on this, the most straightforward approach is to use a well-predicted flow to determine gamma according to Eqn.~\ref{eq:res2gamma}. However, due to the absence of direct optimization for the target, errors in flows can result in significant $\gamma$ errors, leading to less-than-ideal outcomes, as seen in Tab.~\ref{tab:flow}.
Therefore, we propose two methods to incorporate a flow prior to the P+P task to establish the connection between continuous frames. In the first approach, we use flow as an additional input, concatenated with the input image. This leads to a noticeable enhancement but demands additional computational resources to predict the flow. Subsequently, we discovered that flow-pretraining can effectively help the network maintain the continuity between frames without superfluous overhead. With flow-pretraining, the $\gamma$ prediction becomes a much easier assignment. Tab.\ref{tab:flow_result} presents our experimental results using modified versions of GMFlow~\cite{xu2022gmflow}, MaskFlownet~\cite{zhao2020maskflownet}, and ARFlow~\cite{liu2020learning}. The findings across all models are consistent, highlighting the substantial improvement brought about by flow pre-training. In contrast, without the incorporation of flow, there is only a marginal performance increase between single and multiple frames.

\noindent
\textbf{Planar Position Embedding (PPE)}
In P+P methods, there are two parameters the camera's intrinsic $\mathcal{K}$ and the normal vector of the plane $\vec{\mathbf{N}}^T$. The network should be aware of these two variables. Inspired by position embedding from transformer models, we propose Planar Position Embedding, which combines these two into one formula, $\vec{\mathbf{N}}^T(\mathcal{K}^{-1}\mathbf{p})$.
As shown in Eqn.~\ref{eq:nkp}, the PPE is the projection of the point in the normalized image plane in the normal direction of the reference plane. Fig.~\ref{fig:network} visually illustrates the PPE and how it captures changes in the plane's orientation relative to the camera.
We use it as the pixel's position related to the reference plane and build embedding as follows:
\begin{equation}    \label{eq:embed}
    \mathcal{F} = \phi(\mathcal{E})
\end{equation}
\begin{equation}    \label{eq:embed2}
    \mathcal{E}_{ij} = \vec{\mathbf{N}}^T(\mathcal{K}^{-1}\mathbf{p}_{ij})
\end{equation}
where $\phi$ is a simple convolutional network. It lets the network be aware of the relative position of the plane, suppressing some absurd errors. The results from Tab.\ref{tab:ablation} and Fig.\ref{fig:ppe} indicate that PPE effectively suppresses the high-error regions caused by dynamic objects or changes in slope. We investigate its effectiveness in Sec.~\ref{sec:ablation}.

\noindent
\textbf{Single Frame Branch(SFB)}
To further suppress the errors brought about by inaccurate observations across multiple frames, such as moving objects, we follow the recent work DfM~\cite{wang2022dfm} that introduces a single frame branch to assist the network in deciding whether to rely more on multi-frame observations or more on single-frame prediction.
Our single frame branch utilizes only the target image as input and helps predict $\gamma$, mainly benefiting from the training data and network's generalization power. We integrate this branch with the primary branch as follows:
\begin{equation}
     \mathcal{S} = \sigma[\phi(\mathcal{F}_{m}, \mathcal{F}_{s})],
\end{equation}
\begin{equation}    \label{eq:fuse}
     \mathcal{F} = \mathcal{S} \circ \mathcal{F}_{m} + (1 - \mathcal{S}) \circ \mathcal{F}_{s},
\end{equation}
where $\sigma$ is the sigmoid function. $\circ$ is element-wise multiplication. $\phi$ is a simple convolutional layer. The fusion score $\mathcal{S}$ is used to guide the fusion of the single-frame feature $\mathcal{F}_{s}$ and multi-frame feature $\mathcal{F}_{m}$.

\noindent
\textbf{Overall Structure}
Combining the abovementioned methods, we propose a new framework named \Name. As shown in Fig.~\ref{fig:network}, the network takes two consecutive images $I_{t-1}$ and $I_{t}$ as input. Image $I_{t-1}$ is warped using road plane homography. The network outputs $\gamma$ map of image $I_{t}$, which is then transformed into the depth map using Eqn.~\ref{eq:gamma2Z}. We adopt the feature extraction and feature enhancement component introduced in GMFlow~\cite{xu2022gmflow} and leave the feature matching and flow propagation only for flow pre-training.

We replace the CNN backbone in GMFlow with a Swin Transformer~\cite{liu2021Swin} for greater model capacity.
The $\frac{1}{8}$ downsampled feature map is fed into the feature enhancement transformer, where the P+P geometry can be analyzed. The $\frac{1}{32}$ downsampled feature map is used for single frame estimation, which is then fused with the output of the feature enhancement transformer. After the fusion, planar position embedding will be introduced. The final feature is upsampled to the original size to output $\gamma$.

\noindent
\textbf{Training Loss.}
Following previous work~\cite{yuan2022newcrfs, bhat2021adabins, lee2019big, yuan2021monocular, eigen2014depth}, we use an L1 loss to supervise $\gamma$ and a Scale-Invariant Logarithmic (SILog) loss to supervise the depth obtained by $\gamma$ using Eqn.~\ref{eq:gamma2Z}. If the total number of pixels with ground truth is $N$, the loss for $\gamma$ can be defined as
\begin{equation}
  L_{\gamma} =  \frac{1}{N}\sum_i|\gamma_i - \gamma^*_i|,
\end{equation}
where $\gamma_i$ and $\gamma^*_i$ are the predicted $\gamma$ value and corresponding ground-truth.

The logarithm difference is defined as
\begin{equation}
  \Delta d_i =  \log d_i - \log d_i^*,
\end{equation}
where $d_i$ and $d_i^*$ are the predicted depth value obtained by $\gamma$ using Eqn.~\ref{eq:gamma2Z} and its corresponding ground-truth depth value.
Then the depth loss is defined as:
\begin{equation} \label{eq:lossd}
  L_{d} =  \alpha \sqrt{\frac{1}{N}\sum_i{\Delta d_i^2} - \frac{\lambda}{N^2}(\sum_i{\Delta d_i})^2},
\end{equation}
where $\lambda$ is a variance minimizing factor, and $\alpha$ is a scale constant. Following the previous works \cite{yuan2022newcrfs, lee2019big}, we set $\lambda=0.85$ and $\alpha=10$.

The total loss function is defined as the summation of the loss for $\gamma$ and depth with weight $w_{\gamma}$ and $w_{d}$:

\begin{equation}
  L = w_{\gamma} L_{\gamma} + w_{d}L_{d}.
\end{equation}

Here we set $w_{\gamma}=1$ and $w_{d}=10^{-2}$ since the depth produced by Eqn.~\ref{eq:gamma2Z} may be very unstable. We choose the weight of depth loss by grid search, which is shown in appendix.  Also, the network should focus more on the geometric advantage that $\gamma$ brings.

\begin{table*}[tb]
    \setlength{\tabcolsep}{1.1mm}{
	\begin{center}  
	\small
	\begin{tabular}{lccccccccccc}  
	\toprule[1pt]
	Method & Frame & Backbone & Abs Rel $\downarrow$ & Sq Rel $\downarrow$ & RMSE $\downarrow$ & RMSE log  $\downarrow$ & $\delta_1$ $\uparrow$ & $\delta_2$ $\uparrow$ & $\delta_3$ $\uparrow$ & Params & FPS  \\  \hline
	
	Eigen~\etal~\cite{eigen2014depth} & 1(0) & - & 0.190 & 1.515 & 7.156 & 0.270 & 0.692 & 0.899 & 0.967 & 83 M & -	\\
	DORN~\cite{fu2018deep} & 1(0) & ResNet-101 & 0.072 & 0.307 & 2.727 & 0.120 & 0.932 & 0.984 & 0.995 & 100 M & -	\\
	BTS~\cite{lee2019big} & 1(0) & ResNext-101 & 0.059 & 0.245 & 2.756 & 0.096 & 0.956 & 0.993 & \underline{0.998} & 113 M & 21.4	\\
	DPT~\cite{Ranftl_2021_ICCV} & 1(0) & VIT-Hybrid & 0.062 & 0.222 & 2.575 & 0.092 & 0.959 & 0.995 & \textbf{0.999} & 123 M & 20.1 \\
	Adabin~\cite{bhat2021adabins} & 1(0) & EfficientNet-B5 & 0.058 & 0.190 & 2.360 & 0.088 & 0.964 & 0.995 & \textbf{0.999} & 78 M & 20.6	\\  %17.0
	NeW CRFs~\cite{yuan2022newcrfs} & 1(0) & Swin-Large & 0.052 & 0.155 & 2.129 & 0.079 & 0.974 & \underline{0.997} & \textbf{0.999} & 270 M & 25.5	\\
	PixelFormer~\cite{agarwal2023attention} & 1(0) & Swin-Large & 0.051 & 0.149 & 2.081 & 0.077 & 0.976 & \underline{0.997} & \textbf{0.999} & 271 M & 30.0	\\
    VA-DepthNet~\cite{liu2022va} & 1(0) & Swin-Large & 0.050 & 0.148 & 2.093 & 0.076 & 0.977 & \underline{0.997} & \textbf{0.999} & 263 M & 7.0	\\
    iDisc~\cite{piccinelli2023idisc} & 1(0) & Swin-Large & 0.050 & 0.144 & 2.067 & 0.077 & 0.977 & \underline{0.997} & \textbf{0.999} & 209 M & 7.7	\\
	\hline
	
	MaGNet~\cite{bae2022multi} & 3(-2,0,2) & - & 0.054 & 0.162 & 2.158 & 0.083 & 0.971 & 0.996 & \textbf{0.999} & 76 M & 8.1 \\
	*DeepV2D~\cite{teeddeepv2d} & 5(-2,-1,0,1,2) & - & 0.037 & 0.174 & 2.005 & 0.074 & 0.977 & 0.993 & 0.997 & 51 M & 1.5 \\
	DeepV2D~\cite{teeddeepv2d} & 5(-2,-1,0,1,2) & - & 0.046 & 0.191 & 2.107 & 0.080 & 0.975 & 0.993 & 0.997 & 51 M & 1.5 \\
	\hline
	
	Ours(MP) & 2(-1, 0) & Swin-Tiny & 0.044 & 0.127 & 1.986 & 0.069 & 0.981 & \underline{0.997} & \textbf{0.999} & 52 M & 23.9	\\
	Ours(EP) & 2(-1, 0) & Swin-Tiny & 0.037 & 0.109 & 1.815 & 0.062 & 0.983 & \underline{0.997} & \textbf{0.999} & 52 M & 23.9	\\
	\hline
	Ours(EP) & 3(-2, -1, 0) & Swin-Tiny & \underline{0.035} & \underline{0.101} & \underline{1.741} & \underline{0.059} & \underline{0.985} & \textbf{0.998} & \textbf{0.999} & 52 M & 11.3	\\
	Ours(EP) & 3(-1, 0, 1) & Swin-Tiny & \textbf{0.033} & \textbf{0.092} & \textbf{1.667} & \textbf{0.056} & \textbf{0.986} & \textbf{0.998} & \textbf{0.999} & 52 M & 11.3	\\
	\bottomrule[1pt]
	\end{tabular}
	\end{center}  
	}
	\caption{\label{tab:kitti}Quantitative results on the Eigen split of KITTI dataset. The best results are in \textbf{bold} and second best are \underline{underlined}. Frame 0 is the current frame. FPS is tested in RTX 3090. The input size for DeepV2D is $1088 \times 192$ and $1216 \times 352$ for others. * means the results are rescaled by ground-truth.}
\end{table*}

\section{Experiments}
\label{sec:experiments}

\subsection{Datasets}
\label{sec:dataset}
We use KITTI dataset~\cite{geiger2012we} and Waymo Open Dataset~\cite{sun2020scalability}  to evaluate the performance of the proposed network.

\noindent
\textbf{KITTI.}
The KITTI dataset is widely used as a benchmark for Monocular Depth Estimation. We follow the data split proposed by Eigen~\etal~\cite{eigen2014depth}, which includes 23488 training samples and 697 testing samples. 
We utilize the RANSAC algorithm to extract the road plane from the ground-truth depth map by reprojecting all the points back to 3D space. The homography transformation is calculated using the odometry data provided by KITTI, following previous work~\cite{yuan2021monocular}.
Moreover, we rectify some inaccurate pose matrices using the point-to-plane ICP algorithm~\cite{chen1992object}. 
To demonstrate the universality of our approach, we utilize two distinct settings for the reference plane:

\begin{itemize}
  \item \textbf{Estimated Plane(EP)}. In this setting, we aim to simulate scenarios with a High-Definition Map (HD Map) where road surface information can be predetermined and stored. The road planes are extracted using the RANSAC algorithm to fit a flat plane in the scene. For the images without a road plane, we use the mean plane method below. 
  \item \textbf{Mean Plane(MP)}. In scenarios without an HD map, the relationship between the camera and the ground can be obtained from the extrinsic parameters between the camera and the vehicle using methods such as IPM~\cite{ipm}. However, KITTI does not provide precise values for these extrinsic parameters. We use mean plane to simulate the plane derived from the extrinsic parameters.
  The mean plane is acquired by computing the mean plane normal of the estimated plane from all the frames in the training set.
\end{itemize}

\noindent
It's worth noting that our EP setting follows ~\cite{yuan2021monocular}. Compared to the monocular depth estimation method, it additionally utilizes a pre-stored ground plane. Both settings do not require additional LiDAR input and planar calculation during runtime.

\noindent
\textbf{Waymo Open Dataset.}
To compare with RPANet~\cite{yuan2021monocular}, we reproduced the RP2-Waymo dataset, which is currently unpublished. Our reproduced dataset comprises 12,894 training samples and 1,345 test samples.
Note that WOD lacks dense depth supervision, and the ground truth $\gamma$ is built from sparse observations gathered by LiDAR. While experiments conducted on WOD demonstrate the generalization of the methods, it is worth considering that the ground-truth depth obtained from LiDAR may contain some unexpected errors if not processed deliberately.
We refer readers to the appendix for specific examples.

\begin{table*}[tb]
	\begin{center} 
	\begin{tabular}{lccccccccc}  
	\toprule[1pt]
	Method & Height & Frame & Abs Rel $\downarrow$ & Sq Rel $\downarrow$ & RMSE $\downarrow$ & RMSE log  $\downarrow$ & $\delta_1$ $\uparrow$ & $\delta_2$ $\uparrow$ & $\delta_3$ $\uparrow$  \\  \hline
	RPANet~\cite{yuan2021monocular} & $<1$m & 2(-1, 0) & 0.036 & 0.198 & 2.707 & 0.080 & 0.974 & 0.992 & 0.997	\\
	BTS~\cite{lee2019big} & $<1$m & 1(0) & 0.044 & 0.166 & 2.383 & 0.075 & 0.980 & 0.995 & 0.998	\\
	DPT~\cite{Ranftl_2021_ICCV} & $<1$m & 1(0) & 0.042 & 0.156 & 2.280 & 0.072 & 0.981 & \textbf{0.996} & 0.998 	\\
	NeW CRFs~\cite{yuan2022newcrfs} & $<1$m & 1(0) & 0.043 & 0.155 & 2.321 & 0.072 & 0.981 & \textbf{0.996} & \textbf{0.999}	\\
	\hline
	Ours(EP) & $<1$m & 2(-1, 0) & \textbf{0.029} & \textbf{0.131} & \textbf{2.200} & \textbf{0.065} & \textbf{0.984} & 0.995 & 0.998	\\
	\midrule[1pt]
	RPANet~\cite{yuan2021monocular} & - & 2(-1, 0) & 0.086 & 1.089 & 5.623 & 0.187 & 0.903 & 0.968 & 0.987	\\
	BTS~\cite{lee2019big} & - & 1(0) & 0.071 & 0.531 & 4.105 & 0.119 & 0.939 & 0.984 & 0.995	\\
	DPT~\cite{Ranftl_2021_ICCV} & - & 1(0) & 0.068 & 0.501 & 3.930 & 0.114 & 0.943 & 0.986 & 0.995	\\
	NeW CRFs~\cite{yuan2022newcrfs} & - & 1(0) & 0.067 & 0.459 & 3.866 & 0.112 & 0.945 & \textbf{0.987} & \textbf{0.996}	\\
	\hline
	Ours(EP) & - & 2(-1, 0) & \textbf{0.056} & \textbf{0.450} & \textbf{3.853} & \textbf{0.108} & \textbf{0.950} & \textbf{0.987} & 0.995	\\
	\bottomrule[1pt]
	\end{tabular}
	\end{center}  
	\caption{\label{tab:waymo}Quantitative results on the Waymo Open Dataset. }
\end{table*}

\subsection{Implementation Details}
\label{sec:implementation}
We implement the proposed network using Pytorch~\cite{paszke2019pytorch}. AdamW optimizer\cite{loshchilov2017decoupled} with a weight decay of $10^{-2}$ is adopted. All experiments are performed on Nvidia RTX 3090 GPUs. Following ~\cite{lee2019big}, the learning rate decrease from $10^{-4}$ to $10^{-5}$ using polynomial decay with power $p=0.9$. Our model is trained for $20$ epochs with a total batch size of $8$. We use augmentation techniques such as horizontal flipping and brightness jittering.
Moreover, since the P+P geometry is built on the condition $t_z \neq 0$, we randomly replace the warped image $I_{t-1}$ with image $I_{t}$ to improve the robustness of the static scene. Our model is first pre-trained on the flow estimation task. We follow the training strategy of KITTI dataset in GMFlow~\cite{xu2022gmflow}, the model is first traind on FlyingChairs(Chairs)~\cite{dosovitskiy2015flownet} and FlyingThings3D (Things)~\cite{mayer2016large} datasets, then fine-tuned on Sintel~\cite{butler2012naturalistic} and KITTI~\cite{menze2015object} datasets.

\begin{figure*}[tb]
	\centering
	
	\begin{minipage}{0.24\linewidth}
		\includegraphics[width=1\textwidth]{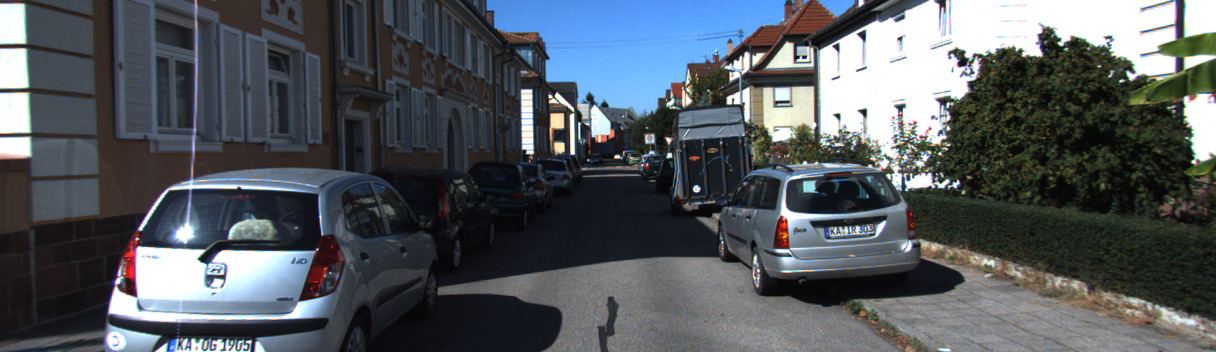}
		\includegraphics[width=1\textwidth]{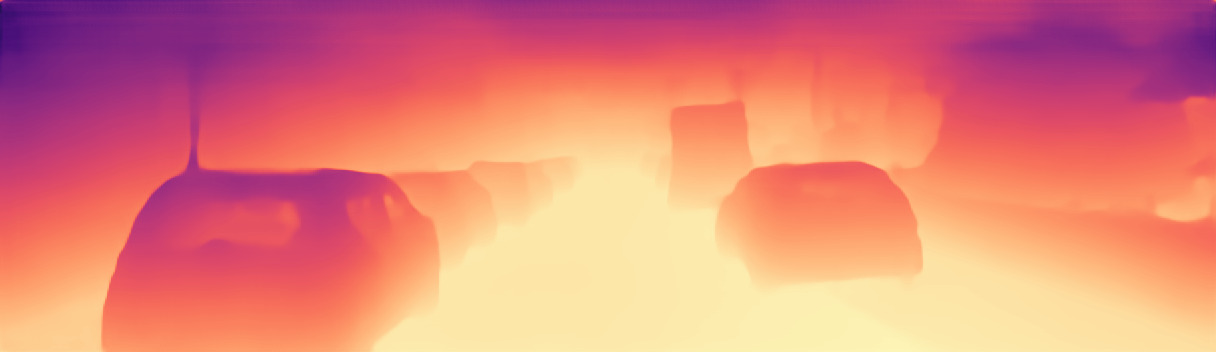}
	\end{minipage}
	\begin{minipage}{0.24\linewidth}
		\includegraphics[width=1\textwidth]{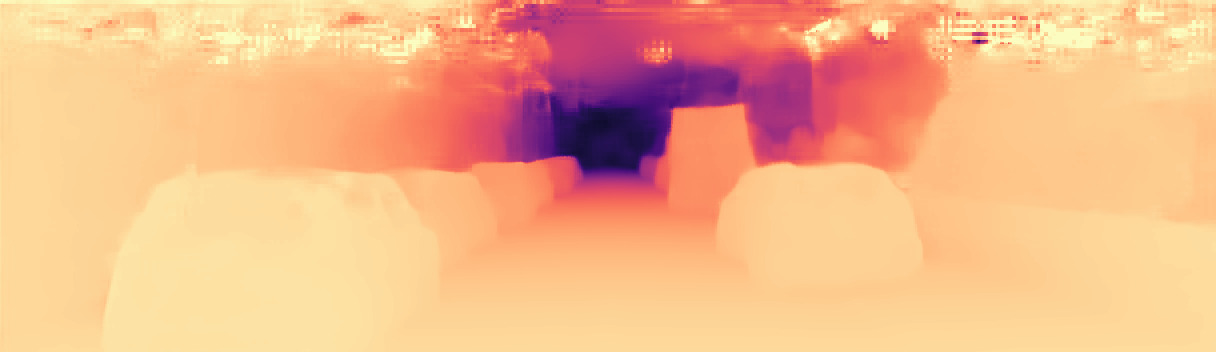}
		\includegraphics[width=1\textwidth]{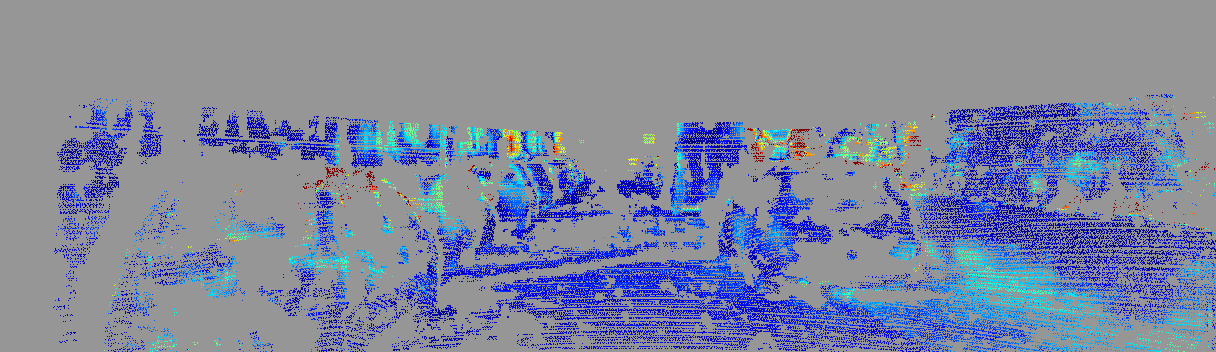}
	\end{minipage}
	\begin{minipage}{0.24\linewidth}
		\includegraphics[width=1\textwidth]{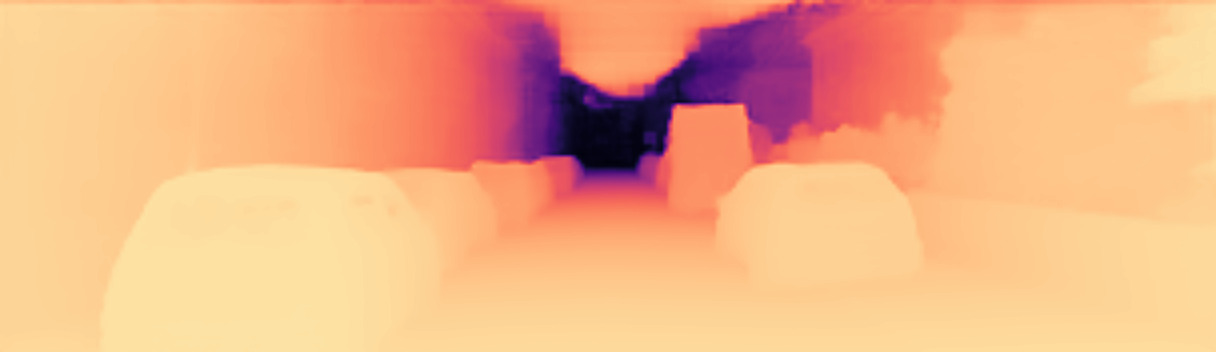}
		\includegraphics[width=1\textwidth]{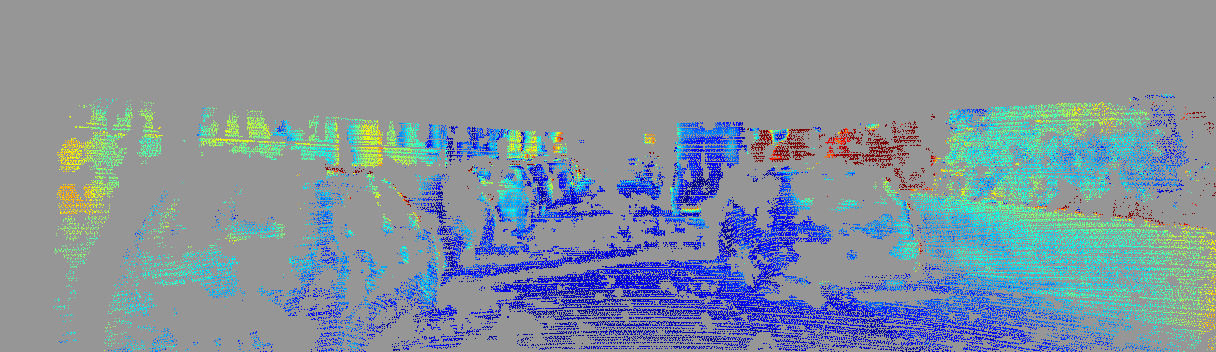}
	\end{minipage}
	\begin{minipage}{0.24\linewidth}
		\includegraphics[width=1\textwidth]{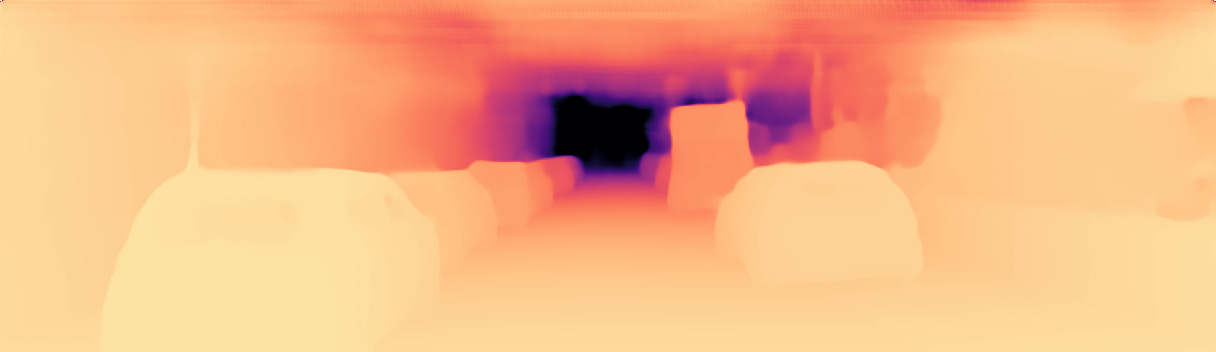}
		\includegraphics[width=1\textwidth]{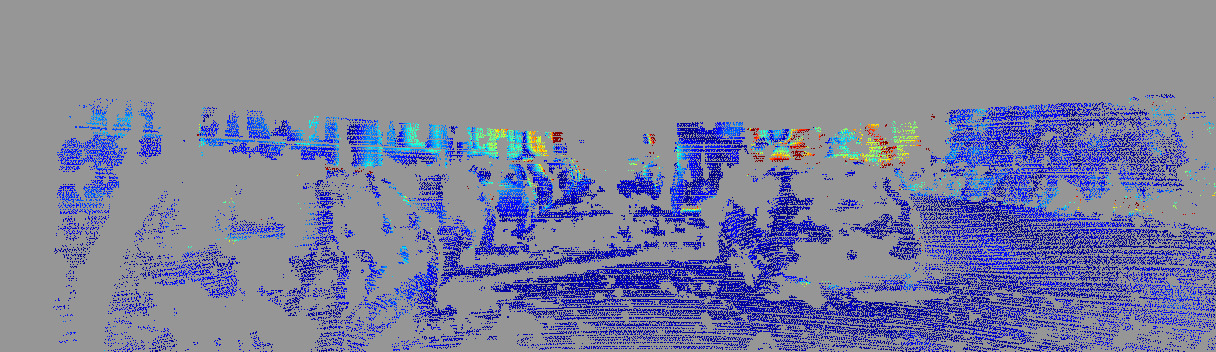}
	\end{minipage}
	
	\begin{minipage}{0.24\linewidth}
		\includegraphics[width=1\textwidth]{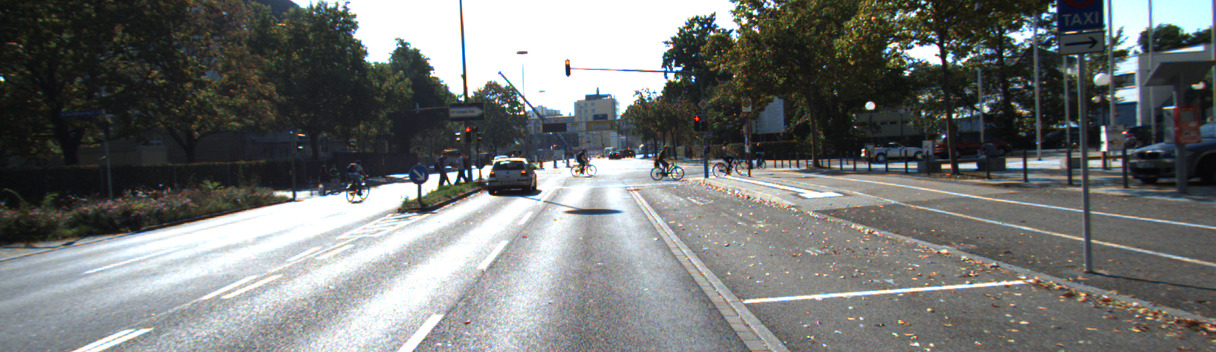}
		\includegraphics[width=1\textwidth]{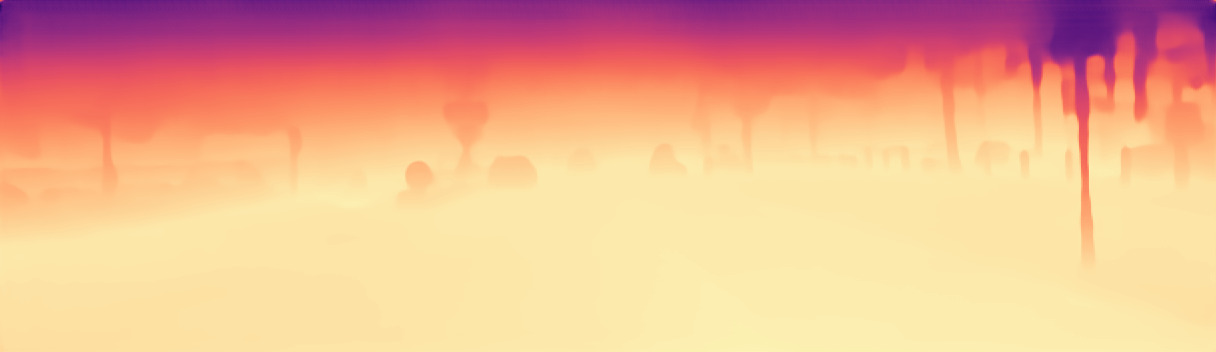}
	\end{minipage}
    \begin{minipage}{0.24\linewidth}
		\includegraphics[width=1\textwidth]{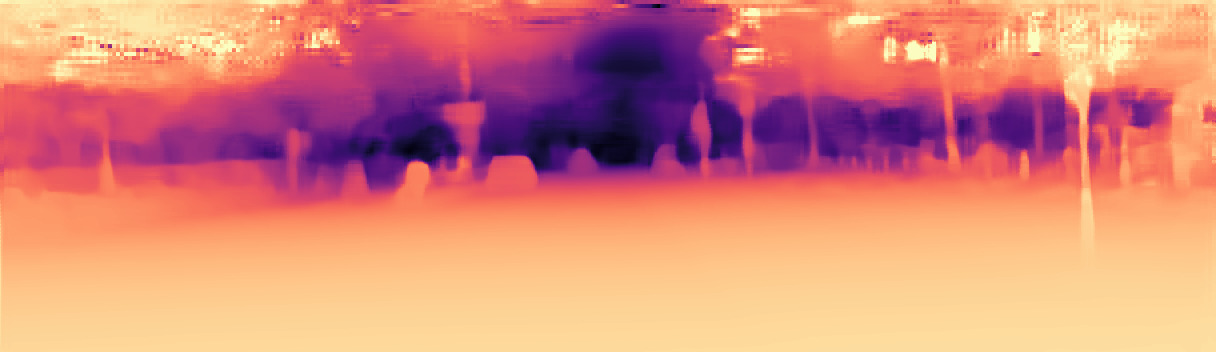}
		\includegraphics[width=1\textwidth]{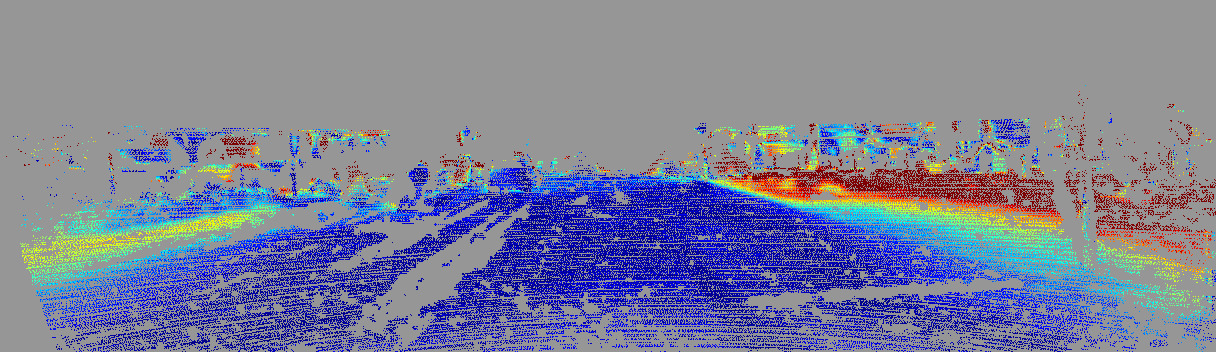}
	\end{minipage}
	\begin{minipage}{0.24\linewidth}
		\includegraphics[width=1\textwidth]{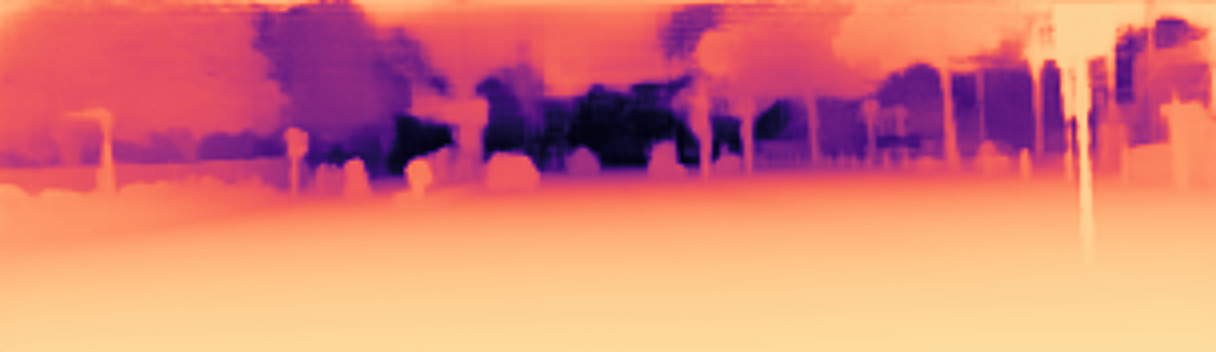}
		\includegraphics[width=1\textwidth]{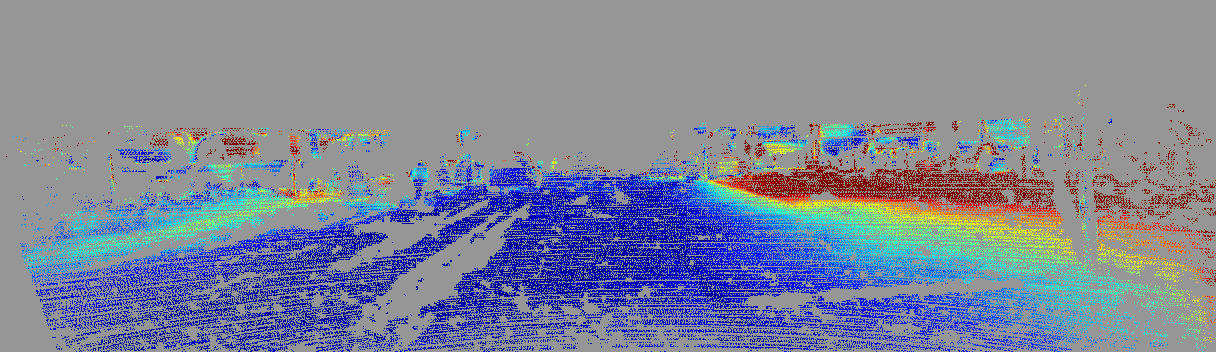}
	\end{minipage}
	\begin{minipage}{0.24\linewidth}
		\includegraphics[width=1\textwidth]{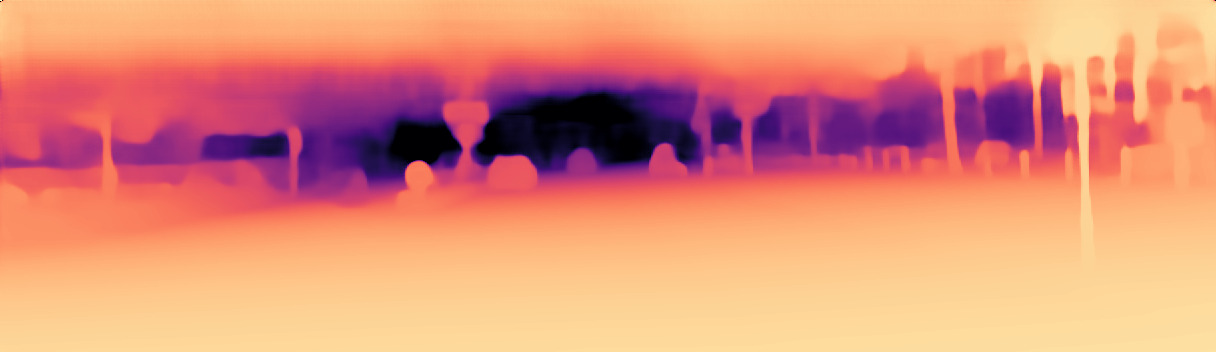}
		\includegraphics[width=1\textwidth]{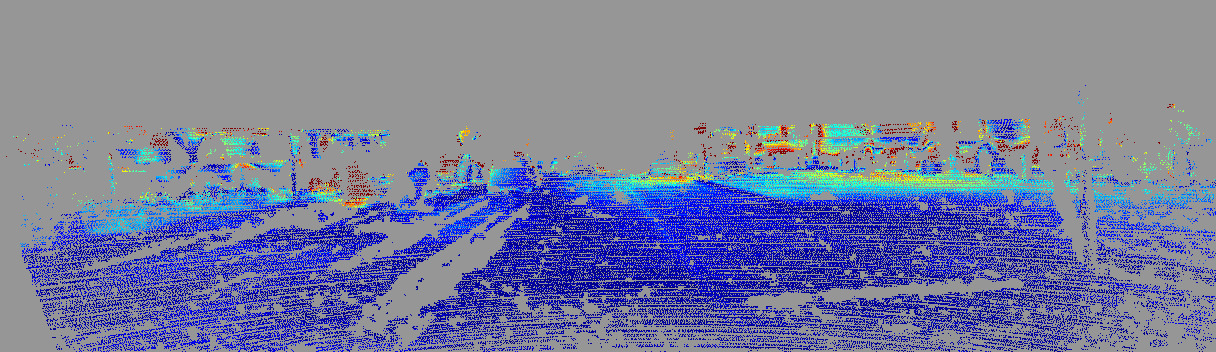}
	\end{minipage}

	\begin{minipage}{0.24\linewidth}
		\centering
		Input Image
	\end{minipage}
	\begin{minipage}{0.24\linewidth}
		\centering
        MaGNet~\cite{bae2022multi}
		% BTS~\cite{lee2019big}
	\end{minipage}
	\begin{minipage}{0.24\linewidth}
		\centering
		NeW CRFs~\cite{yuan2022newcrfs}
	\end{minipage}
	\begin{minipage}{0.24\linewidth}
		\centering
		Ours
	\end{minipage}

	\caption{Qualitative results on the Eigen split of KITTI dataset. For each sample, the first column shows the target image and the predicted $\gamma$ map by our model. The rest columns each show a model's predicted depth map and corresponding error map. Blue represents smaller errors, while red represents larger errors.}	\label{fig:result_pics}
\end{figure*}

\subsection{Comparisons with the state-of-the-art Depth}
\label{sec:comparisons}

We compare our method with state-of-the-art methods including Eigen et al.~\cite{eigen2014depth}, DORN~\cite{fu2018deep}, BTS~\cite{lee2019big}, DPT~\cite{Ranftl_2021_ICCV}, Adabin~\cite{bhat2021adabins}, NeW CRFs~\cite{yuan2022newcrfs}, PixelFormer~\cite{agarwal2023attention}, VA-DepthNet~\cite{liu2022va}, iDisc~\cite{piccinelli2023idisc}, RPANet~\cite{yuan2021monocular}, MaGNet~\cite{bae2022multi}, and DeepV2D~\cite{teeddeepv2d}.

\noindent
\textbf{Depth results on KITTI dataset}
In Tab.~\ref{tab:kitti}, we report the results of depth estimation on KITTI. As our focus is on developing models for autonomous driving scenarios that operate in real-time, we evaluate both single-frame and multi-frame methods. Moreover, to provide a more comprehensive comparison, we include the results of our model in offline mode, which utilizes future frames. The details of how we extend our model to multi-frame can be found in the appendix. The results demonstrate that our model outperforms the previous state-of-the-art model by a significant margin across all evaluation metrics, regardless of whether future frames are used. Notably, our online mode runs in real-time and achieves comparable speeds to single-frame methods while being significantly faster than multi-frame methods.
It is worth noting that although obtaining the estimated plane (EP) during autonomous driving scenes is relatively straightforward, we also report the results using the mean plane (MP), which does not require any plane estimation. With some loss of accuracy by the estimated plane, the improvement is still considerable.
Fig.~\ref{fig:result_pics} illustrates the qualitative results. As shown in the error map, we highlight the advantages of our method in estimating obstacles. By leveraging the ground's priors, our approach significantly enhances the accuracy of vertical and static objects.

\noindent
\textbf{Depth results on Waymo Open Dataset}
As for Waymo Open Dataset, we reproduce the RPANet following \cite{yuan2021monocular} and train BTS, DPT and NeW CRFs with official open-sourced code for comparison\footnote{\url{http://github.com/cleinc/bts},\url{http://github.com/aliyun/NeWCRFs}, \url{https://github.com/isl-org/DPT}}. Tab.~\ref{tab:waymo} presents the results for both the height~$< 1m$ and full range cases. Compared to the KITTI results, our proposed method still shows remarkable improvement in Abs Rel, although the improvement gap in Sq Rel has narrowed. This indicates that our model has higher accuracy overall, but there are more pixels with larger errors across all methods. This is partly due to the inclusion of more challenging data in WOD, such as rainy or nighttime scenes, and slow-driving scenarios that violate the motion assumption in Eqn.~\ref{eq:res}. Despite these challenges, our approach still exhibits superior performance in the height~$< 1m$ setting, thanks to the use of the plane prior. Furthermore, even without height restrictions, our method still performs well compared to RPANet, which produces larger errors in this setting.

\subsection{Ablation Study}
\label{sec:ablation}

\begin{table}[tb]
    \setlength{\tabcolsep}{1.6mm}{
	\begin{center}  
	\footnotesize
	\begin{tabular}{lccccc}  
	\hline  
	Target  & Frame & Warp & Pretrain & Abs Rel $\downarrow$ & RMSE $\downarrow$   \\  \hline
	Flow   & 2 & N & ImageNet & 0.160 & 5.661 	\\
	Depth  & 1 & N & ImageNet & 0.059 & 2.059 	\\
	\hline
	Gamma($\gamma$)  & 1 & N & -        & 0.055 & 2.271 	\\
	Gamma($\gamma$)  & 2 & N & -        & 0.054 & 2.231 	\\
	Gamma($\gamma$)  & 2 & Y & -        & 0.054 & 2.191 	\\
	\hline
	Gamma($\gamma$)  & 1 & N & ImageNet & 0.047 & 1.985 	\\
	Gamma($\gamma$)  & 2 & N & ImageNet & 0.046 & 1.944 	\\
	Gamma($\gamma$)  & 2 & Y & ImageNet & 0.045 & 1.927 	\\
	\hline
	Gamma($\gamma$)  & 1 & N & Flow     & 0.044 & 1.965   \\
	Gamma($\gamma$)  & 2 & N & Flow     & 0.039 & 1.755   \\
	Gamma($\gamma$)  & 2 & Y & Flow     & \textbf{0.035} & \textbf{1.460} 	\\
	\hline
	\end{tabular}
	\end{center}
    }
	\caption{\label{tab:flow} Ablation studies for flow pre-training. The target flow means we use the flow prediction and compute $\gamma$ by Eqn.~\ref{eq:res2gamma}. The target depth indicates that the model predicts depth directly. Frame indicates to number of consecutive frames we use. Warp denotes whether to warp with planar homography. All results are under condition height~$<1m$.}
\end{table}

\noindent
\textbf{Effectiveness of flow pre-training}
To better understand the benefits flow pre-training brings,  we conducted an ablation study where we removed the single frame branch, Planar Position Embedding, and random data augmentation to eliminate the impact of single frame information on the results. We also add a condition height~$<1m$ to highlight the influence caused by the prior of plane.
In Tab.~\ref{tab:flow}, we first show the pure geometry methods. We compute $\gamma$ by Eqn.~\ref{eq:res2gamma} with flow prediction from GMFlow. This method does not work because dynamic objects do not meet the hypothesis, and some pixels near the epipole are too sensitive to the flow's precision. Furthermore, the comparison between depth and $\gamma$ shows the superiority of $\gamma$ prediction. As in Eqn.~\ref{eq:gamma2Z}, $\gamma$ is independent of intrinsic $\mathcal{K}$, which improves its generalization. 
Above all, we conduct experiments in different pre-training, from scratch, ImageNet, and optical flow. The similar performance between one or two frames on scratch and ImageNet pre-train indicates that if the model is not properly pre-trained, even if it is designed to learn from multiple frames, it still prefers to learn from one frame rather than consecutive frames. 
On the contrary, the flow pre-training improves the results significantly and extends the gap between one or two frames.
Finally, we examined the influence of the planar prior by showing the results of the model without warping the image, which only used epipolar geometry. The results were still worse than our proposed method, indicating that the planar prior played a critical role in achieving superior results.

\begin{figure}[tb]
    \centering
    \begin{minipage}{0.49\linewidth}
    	\centering
    	\includegraphics[width=1\textwidth]{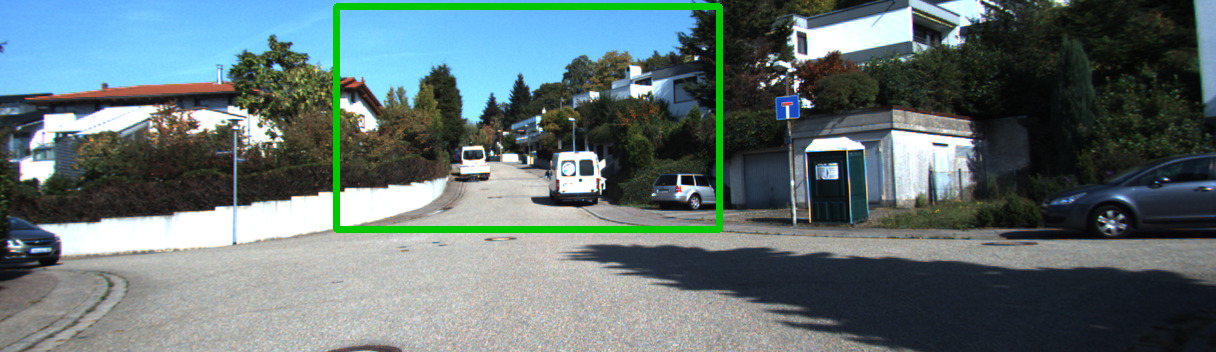}
    	\includegraphics[width=1\textwidth]{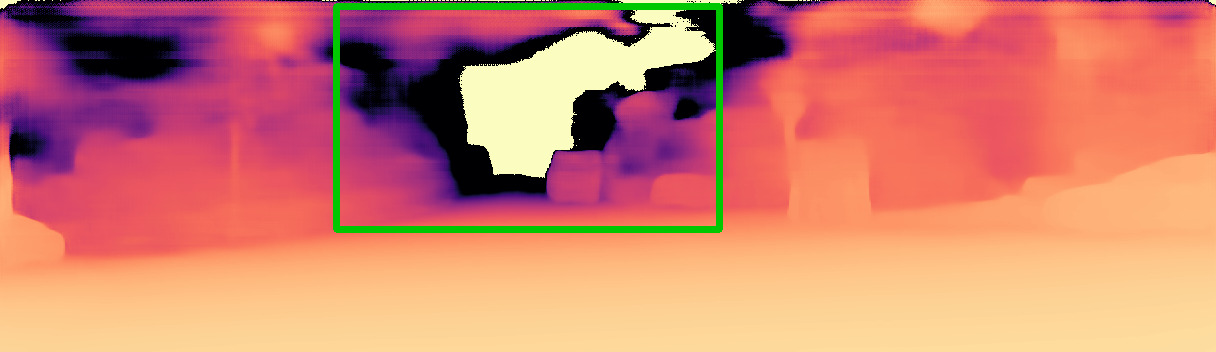}
    	\includegraphics[width=1\textwidth]{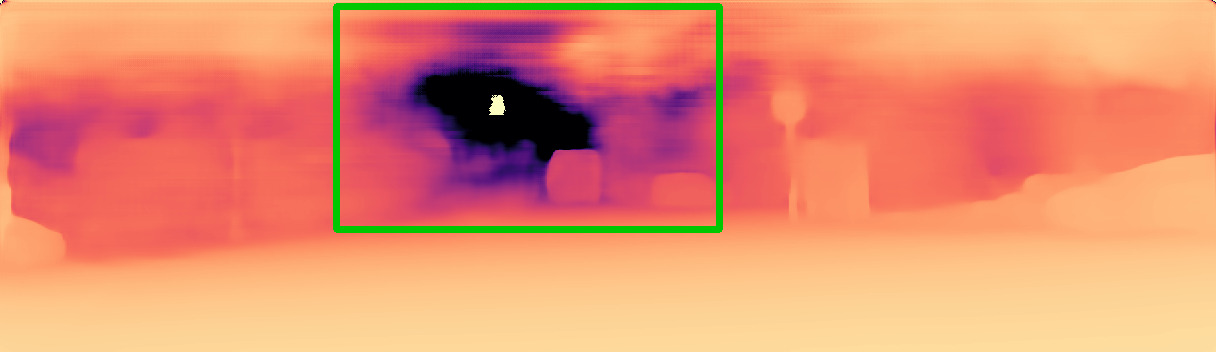}
    	\includegraphics[width=1\textwidth]{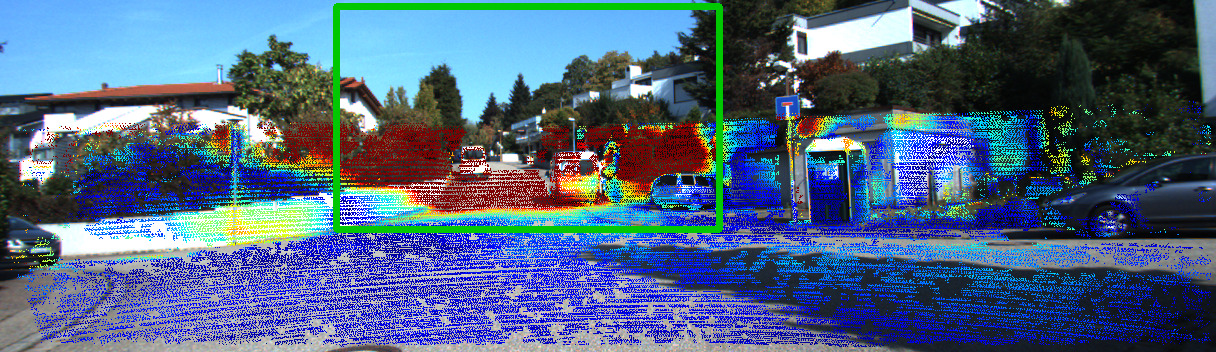}
    	\includegraphics[width=1\textwidth]{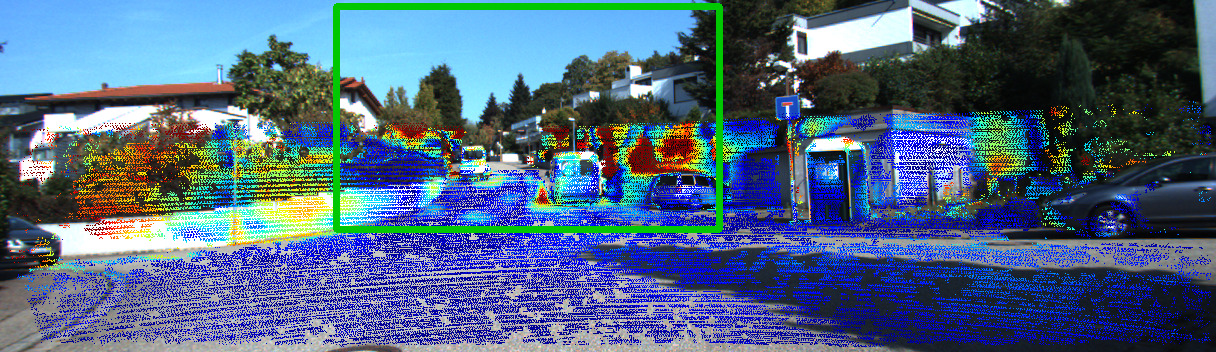}
    \end{minipage}
    \begin{minipage}{0.49\linewidth}
    	\centering
    	\includegraphics[width=1\textwidth]{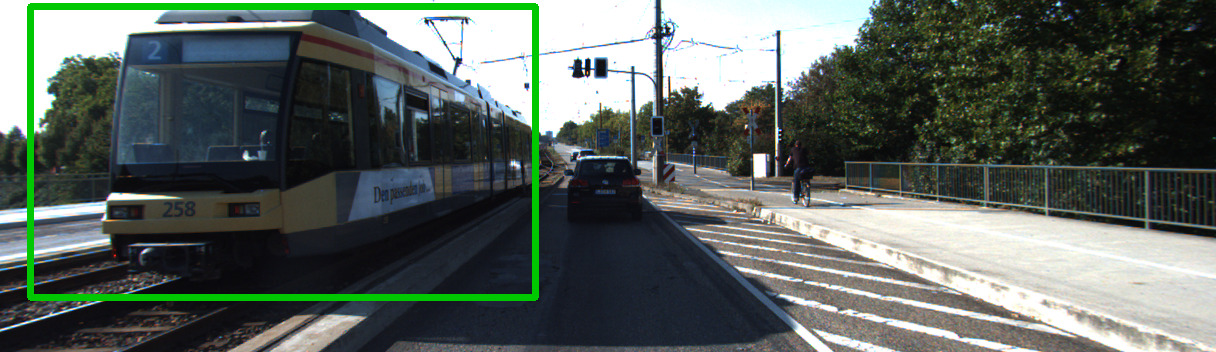}
    	\includegraphics[width=1\textwidth]{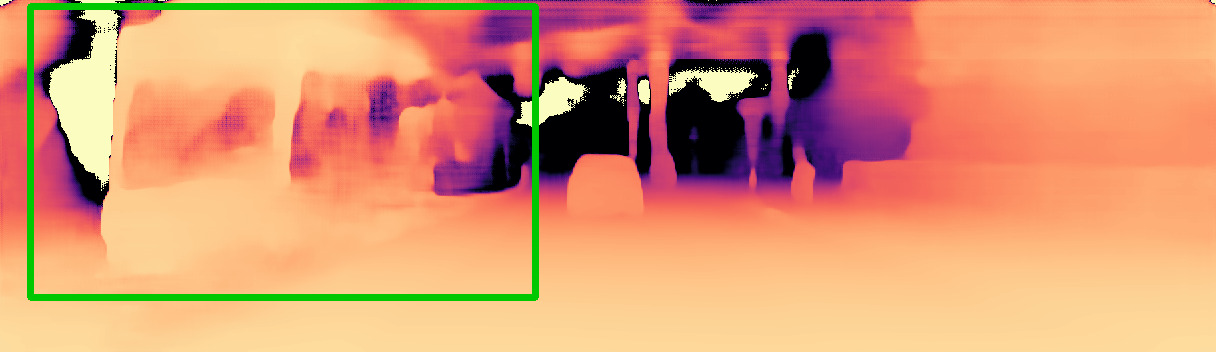}
    	\includegraphics[width=1\textwidth]{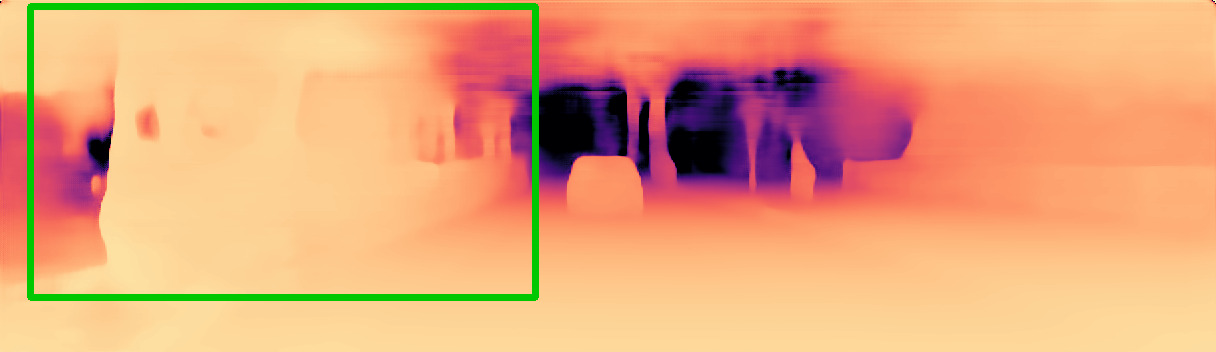}
    	\includegraphics[width=1\textwidth]{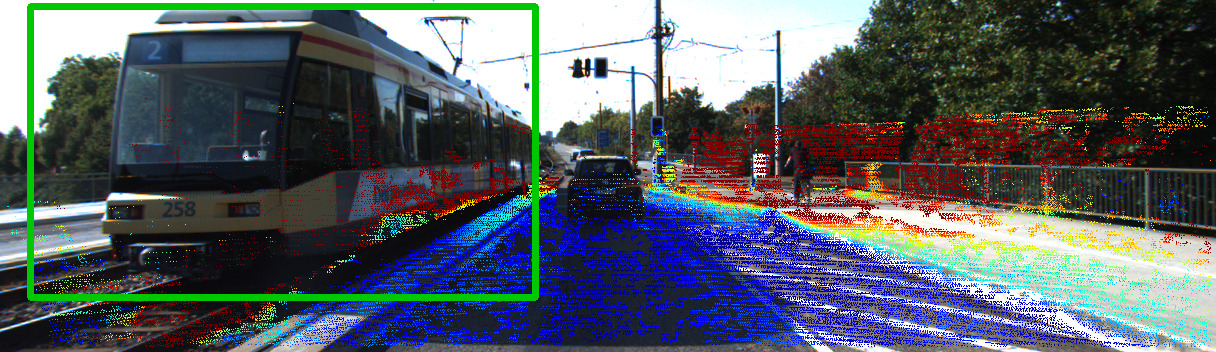}
    	\includegraphics[width=1\textwidth]{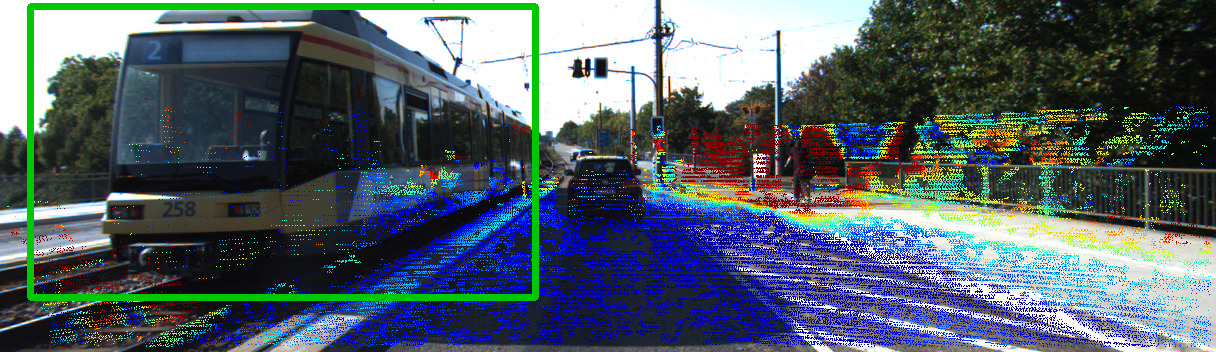}
    \end{minipage}
    \newline
    
    \caption{Row 1: Original images. The first sample contains an uphill road. The second sample contains a fast-moving train. Row 2: Predicted depth map without PPE. Row 3: Predicted depth map with PPE. Row 4: Error map without PPE. Row 5: Error map with PPE.}
    \label{fig:ppe}
\end{figure}

\begin{table}[tb]
	\begin{center}
	\begin{tabular}{lcccccc}  
	\hline  
	Method & Abs Rel $\downarrow$ & Sq Rel $\downarrow$ & RMSE $\downarrow$ & $\delta_1$ $\uparrow$    \\  \hline
	w/o FP & 0.064 & 0.294 & 2.898 & 0.938	\\
	w/o PPE & 0.045 & 0.169 & 2.222 & 0.970	\\
	w/o SFB & 0.041 & 0.137 & 1.936 & 0.976 \\
	w/o DL & \textbf{0.037} & 0.117 & 1.878 & 0.982	\\
	PPNet & \textbf{0.037} & \textbf{0.109} & \textbf{1.815} & \textbf{0.983} \\
	\hline
	\end{tabular}
	\end{center}
	\caption{\label{tab:ablation} Ablation studies for components. Flow pretrain is shown as FP. SFB means single frame branch. DL is the auxiliary depth loss(Eqn.~\ref{eq:lossd}). PPE is proposed Planar Position Embedding.}
\end{table}

\begin{figure}[tb]
    \centering
    \begin{minipage}{0.49\linewidth}
    	\centering
    	\includegraphics[width=1\textwidth]{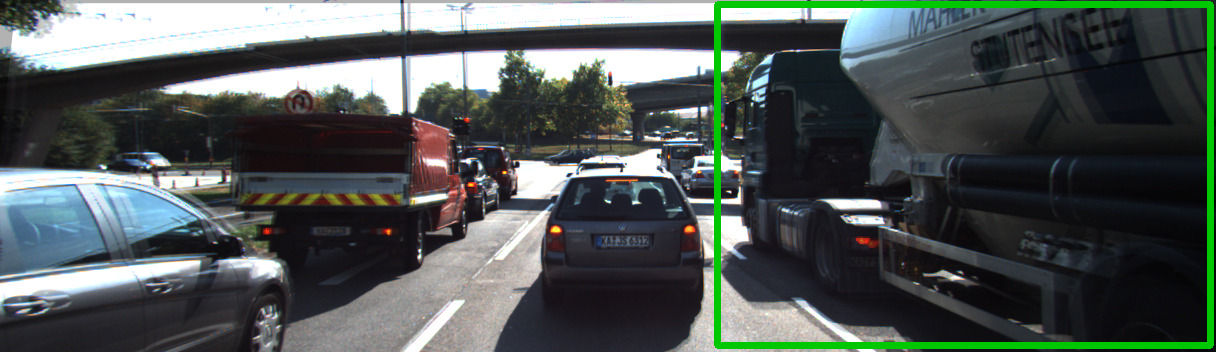}
    	\includegraphics[width=1\textwidth]{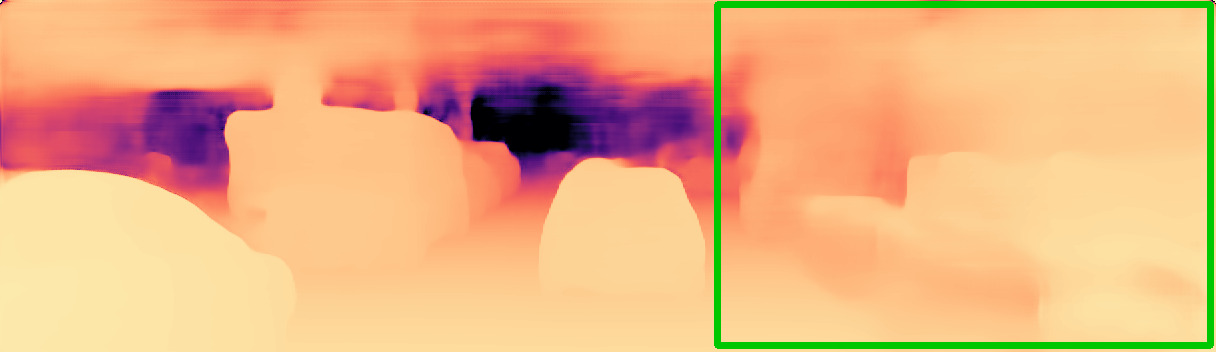}
    	\includegraphics[width=1\textwidth]{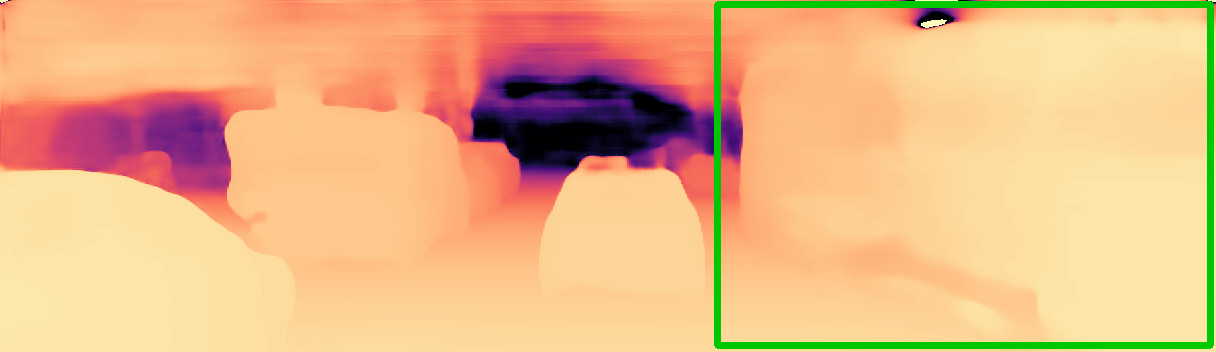}
    	\includegraphics[width=1\textwidth]{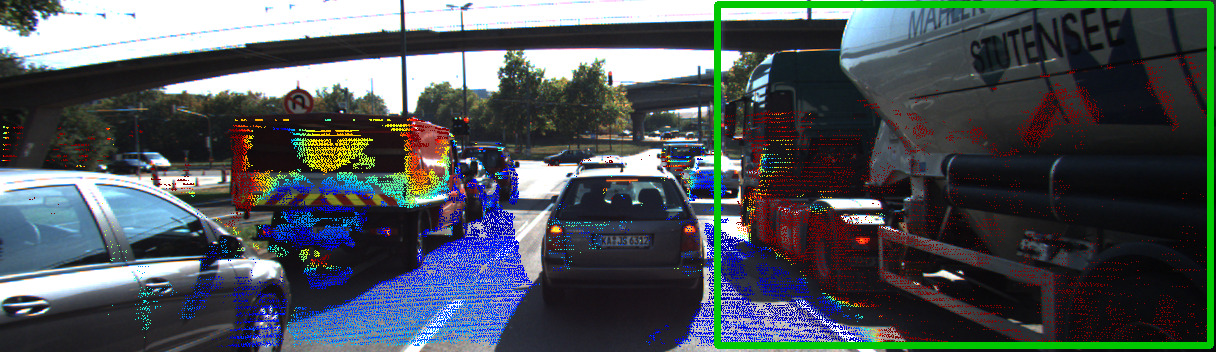}
    	\includegraphics[width=1\textwidth]{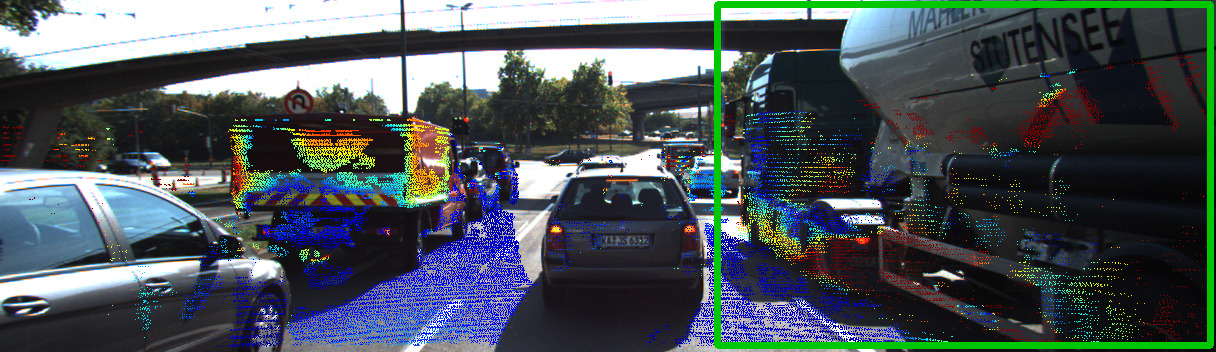}
    \end{minipage}
    \begin{minipage}{0.49\linewidth}
    	\centering
    	\includegraphics[width=1\textwidth]{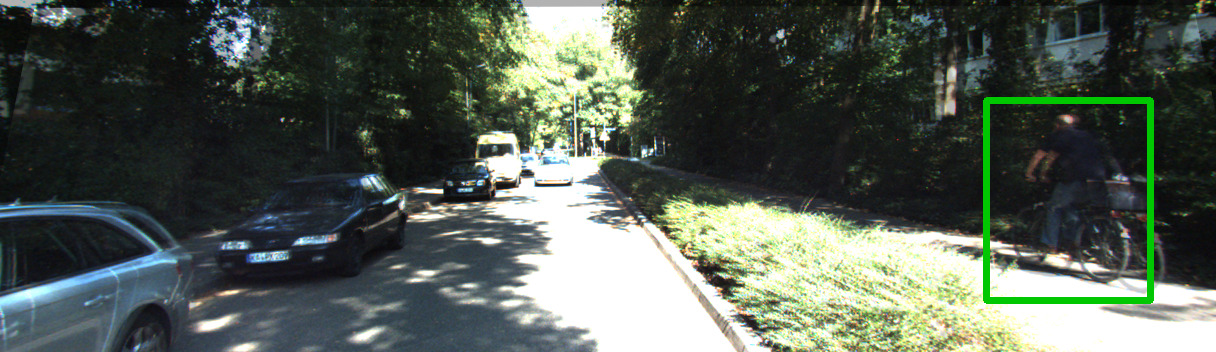}
    	\includegraphics[width=1\textwidth]{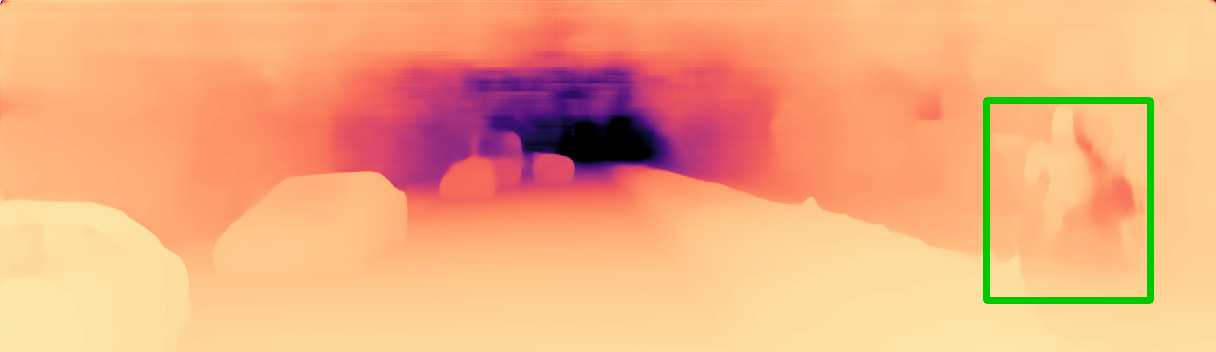}
    	\includegraphics[width=1\textwidth]{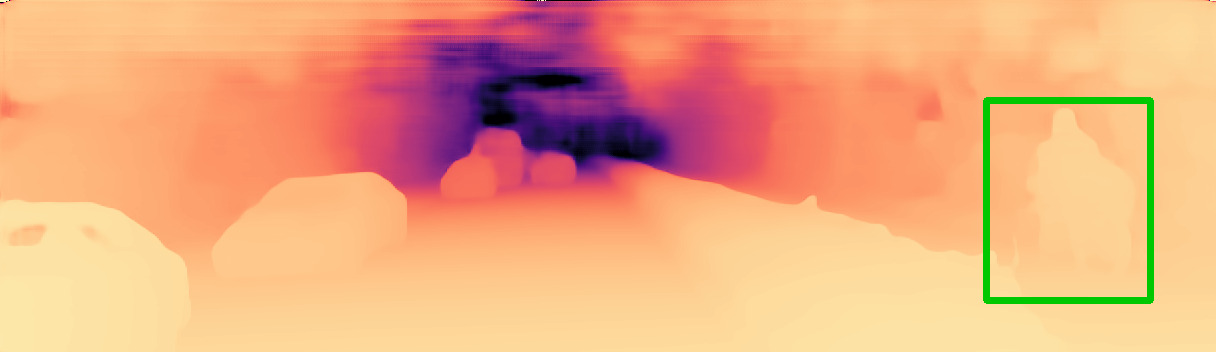}
    	\includegraphics[width=1\textwidth]{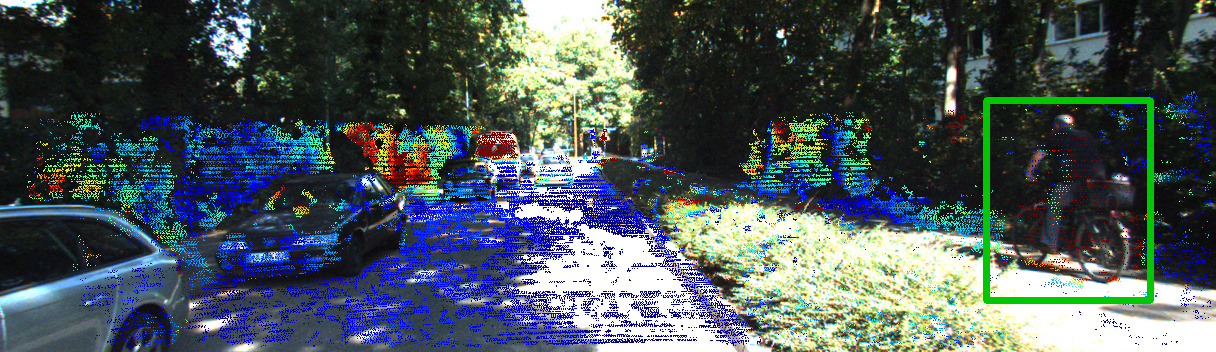}
    	\includegraphics[width=1\textwidth]{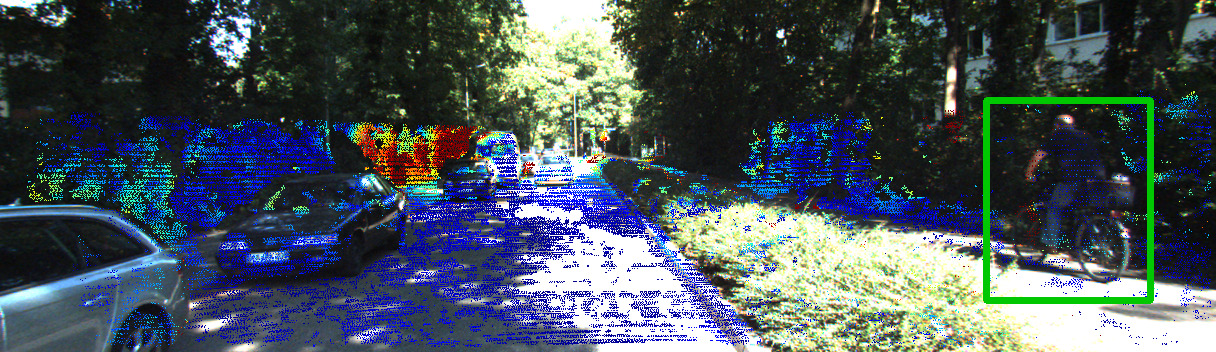}
    \end{minipage}
    \newline
    
    \caption{Row 1: Homography-aligned image pairs. Both samples have small ego motion and contain moving objects. Row 2: Predicted depth map without SFB. Row 3: Predicted depth map with SFB. Row 4: Error map without SFB. Row 5: Error map with SFB.}
    \label{fig:sfb}
\end{figure}

\noindent
\textbf{Effect of each component.}
In Tab.~\ref{tab:ablation}, we analyze the effectiveness of each component in our proposed model by removing them individually. Firstly, we verify the importance of flow pre-training, which significantly affects all evaluation metrics when removed.
The Planar Position Embedding (PPE) contains the positional information of each pixel on the ground plane, which helps to mitigate errors caused by uneven road surfaces or moving objects. The influence of PPE can be observed more intuitively in Fig.~\ref{fig:ppe}. While the idea of embedding seems straightforward, selecting the appropriate origin point is nontrivial. We conducted meticulous experiments to determine the optimal origin point.
As shown in Tab.~\ref{tab:ppe}, the choice based on Eqn.~\ref{eq:nkp} yielded the best results.
As shown in Fig.~\ref{fig:sfb}, the single frame branch improves the performance of dynamic objects and static frames.
Lastly, adding additional depth supervision can make the network learn the depth information directly, leading to the final performance.

\begin{table}[tb]
    \setlength{\tabcolsep}{2mm}{
	\begin{center}  
	\small
	\begin{tabular}{lcccccc}  
	\hline  
	Method & Abs Rel $\downarrow$ & Sq Rel $\downarrow$ & RMSE $\downarrow$ & $\delta_1$ $\uparrow$    \\  \hline
	
	- & 0.045 & 0.169 & 2.222 & 0.970	\\
	$\mathbf{p}$ & 0.043 & 0.175 & 2.108 & 0.975	\\
	$\mathcal{K}$ & 0.044 & 0.164 & 2.121 & 0.973	\\
	$\vec{\mathbf{N}}^T$ & 0.044 & 0.194 & 2.202 & 0.973 \\
	$\mathcal{K}^{-1}\mathbf{p}$ & 0.043 & 0.163 & 2.134 & 0.974	\\
	$\vec{\mathbf{N}}^T$ \& $\mathbf{p}$ & 0.044 & 0.167 & 2.107 & 0.973	\\
	$\vec{\mathbf{N}}^T(\mathcal{K}^{-1}\mathbf{p})$ & \textbf{0.037} & \textbf{0.109} & \textbf{1.815} & \textbf{0.983} \\
	\hline
	\end{tabular}
	\end{center}  
	}
	\caption{\label{tab:ppe}Ablation studies for Planar Position Embedding.}
\end{table}

\section{Conclusions and Future Work}
\label{sec:conclusion}
This paper introduces the Planar Parallax Network,a simple but effective depth estimation framework based on planar parallax geometry. By analyzing the effectiveness of geometric information, our method incorporates flow pre-training to ensure the network starts from a well-tuned initialization based on geometric prior. We address the limitations of the planar parallax pipeline through single-frame estimation and Planar Position Embedding. Our comprehensive experiments on the KITTI and Waymo Open Datasets demonstrate that PPNet significantly outperforms previous state-of-the-art methods.

As a potential area for future work, we are considering exploring low-cost flow supervision, as many unsupervised flow methods can be jointly trained in our framework. Additionally, we are interested in integrating our method into real autonomous driving perception systems.

{
    \small
    \bibliographystyle{ieeenat_fullname}
    \bibliography{ref}

\begin{thebibliography}{62}
\providecommand{\natexlab}[1]{#1}
\providecommand{\url}[1]{\texttt{#1}}
\expandafter\ifx\csname urlstyle\endcsname\relax
  \providecommand{\doi}[1]{doi: #1}\else
  \providecommand{\doi}{doi: \begingroup \urlstyle{rm}\Url}\fi

\bibitem[Agarwal and Arora(2023)]{agarwal2023attention}
Ashutosh Agarwal and Chetan Arora.
\newblock Attention attention everywhere: Monocular depth prediction with skip
  attention.
\newblock In \emph{WACV}, 2023.

\bibitem[Aoki et~al.(2019)Aoki, Goforth, Srivatsan, and
  Lucey]{aoki2019pointnetlk}
Yasuhiro Aoki, Hunter Goforth, Rangaprasad~Arun Srivatsan, and Simon Lucey.
\newblock {PointNetLK}: Robust \& efficient point cloud registration using
  {PointNet}.
\newblock In \emph{CVPR}, 2019.

\bibitem[Bae et~al.(2022)Bae, Budvytis, and Cipolla]{bae2022multi}
Gwangbin Bae, Ignas Budvytis, and Roberto Cipolla.
\newblock Multi-view depth estimation by fusing single-view depth probability
  with multi-view geometry.
\newblock In \emph{CVPR}, 2022.

\bibitem[Baehring et~al.(2005)Baehring, Simon, Niehsen, and
  Stiller]{baehring2005detection}
Dietrich Baehring, Stephan Simon, Wolfgang Niehsen, and Christoph Stiller.
\newblock Detection of close cut-in and overtaking vehicles for driver
  assistance based on planar parallax.
\newblock In \emph{IV}, 2005.

\bibitem[Bhat et~al.(2021)Bhat, Alhashim, and Wonka]{bhat2021adabins}
Shariq~Farooq Bhat, Ibraheem Alhashim, and Peter Wonka.
\newblock {AdaBins}: Depth estimation using adaptive bins.
\newblock In \emph{CVPR}, 2021.

\bibitem[Bian et~al.(2019)Bian, Li, Wang, Zhan, Shen, Cheng, and
  Reid]{bian2019}
Jiawang Bian, Zhichao Li, Naiyan Wang, Huangying Zhan, Chunhua Shen, Ming-Ming
  Cheng, and Ian Reid.
\newblock Unsupervised scale-consistent depth and ego-motion learning from
  monocular video.
\newblock \emph{NeurIPS}, 2019.

\bibitem[Bian et~al.(2021)Bian, Zhan, Wang, Li, Zhang, Shen, Cheng, and
  Reid]{bian2021}
Jia-Wang Bian, Huangying Zhan, Naiyan Wang, Zhichao Li, Le Zhang, Chunhua Shen,
  Ming-Ming Cheng, and Ian Reid.
\newblock Unsupervised scale-consistent depth learning from video.
\newblock \emph{IJCV}, 2021.

\bibitem[Butler et~al.(2012)Butler, Wulff, Stanley, and
  Black]{butler2012naturalistic}
Daniel~J Butler, Jonas Wulff, Garrett~B Stanley, and Michael~J Black.
\newblock A naturalistic open source movie for optical flow evaluation.
\newblock In \emph{ECCV}, 2012.

\bibitem[Chen and Medioni(1992)]{chen1992object}
Yang Chen and G{\'e}rard Medioni.
\newblock Object modelling by registration of multiple range images.
\newblock \emph{IVC}, 1992.

\bibitem[Cross et~al.(1999)Cross, Fitzgibbon, and Zisserman]{cross1999parallax}
Geoffrey Cross, Andrew~W Fitzgibbon, and Andrew Zisserman.
\newblock Parallax geometry of smooth surfaces in multiple views.
\newblock In \emph{ICCV}, 1999.

\bibitem[Donati et~al.(2020)Donati, Sharma, and Ovsjanikov]{donati2020deep}
Nicolas Donati, Abhishek Sharma, and Maks Ovsjanikov.
\newblock Deep geometric functional maps: Robust feature learning for shape
  correspondence.
\newblock In \emph{CVPR}, 2020.

\bibitem[Dosovitskiy et~al.(2015)Dosovitskiy, Fischer, Ilg, Hausser, Hazirbas,
  Golkov, Van Der~Smagt, Cremers, and Brox]{dosovitskiy2015flownet}
Alexey Dosovitskiy, Philipp Fischer, Eddy Ilg, Philip Hausser, Caner Hazirbas,
  Vladimir Golkov, Patrick Van Der~Smagt, Daniel Cremers, and Thomas Brox.
\newblock {FlowNet}: Learning optical flow with convolutional networks.
\newblock In \emph{ICCV}, 2015.

\bibitem[Eigen et~al.(2014)Eigen, Puhrsch, and Fergus]{eigen2014depth}
David Eigen, Christian Puhrsch, and Rob Fergus.
\newblock Depth map prediction from a single image using a multi-scale deep
  network.
\newblock \emph{NeurIPS}, 2014.

\bibitem[Fu et~al.(2018)Fu, Gong, Wang, Batmanghelich, and Tao]{fu2018deep}
Huan Fu, Mingming Gong, Chaohui Wang, Kayhan Batmanghelich, and Dacheng Tao.
\newblock Deep ordinal regression network for monocular depth estimation.
\newblock In \emph{CVPR}, 2018.

\bibitem[Geiger et~al.(2012)Geiger, Lenz, and Urtasun]{geiger2012we}
Andreas Geiger, Philip Lenz, and Raquel Urtasun.
\newblock Are we ready for autonomous driving? the {KITTI} vision benchmark
  suite.
\newblock In \emph{CVPR}, 2012.

\bibitem[Gu et~al.(2020)Gu, Fan, Zhu, Dai, Tan, and Tan]{gu2020cascade}
Xiaodong Gu, Zhiwen Fan, Siyu Zhu, Zuozhuo Dai, Feitong Tan, and Ping Tan.
\newblock Cascade cost volume for high-resolution multi-view stereo and stereo
  matching.
\newblock In \emph{CVPR}, 2020.

\bibitem[He et~al.(2016)He, Zhang, Ren, and Sun]{he2016deep}
Kaiming He, Xiangyu Zhang, Shaoqing Ren, and Jian Sun.
\newblock Deep residual learning for image recognition.
\newblock In \emph{CVPR}, 2016.

\bibitem[Ilg et~al.(2017)Ilg, Mayer, Saikia, Keuper, Dosovitskiy, and
  Brox]{flownet2}
Eddy Ilg, Nikolaus Mayer, Tonmoy Saikia, Margret Keuper, Alexey Dosovitskiy,
  and Thomas Brox.
\newblock {FlowNet} 2.0: Evolution of optical flow estimation with deep
  networks.
\newblock In \emph{CVPR}, 2017.

\bibitem[Irani and Anandan(1996)]{irani1996parallax}
Michal Irani and Prabu Anandan.
\newblock Parallax geometry of pairs of points for {3D} scene analysis.
\newblock In \emph{ECCV}, 1996.

\bibitem[Irani et~al.(2002)Irani, Anandan, and Cohen]{irani2002direct}
Michal Irani, P Anandan, and Meir Cohen.
\newblock Direct recovery of planar-parallax from multiple frames.
\newblock \emph{PAMI}, 2002.

\bibitem[Janai et~al.(2018)Janai, Guney, Ranjan, Black, and
  Geiger]{janai2018unsupervised}
Joel Janai, Fatma Guney, Anurag Ranjan, Michael Black, and Andreas Geiger.
\newblock Unsupervised learning of multi-frame optical flow with occlusions.
\newblock In \emph{ECCV}, 2018.

\bibitem[Jung and Hong(2021)]{jung2021quantitative}
Kyunghwa Jung and Jaesung Hong.
\newblock Quantitative assessment method of image stitching performance based
  on estimation of planar parallax.
\newblock \emph{IA}, 2021.

\bibitem[Kopf et~al.(2021)Kopf, Rong, and Huang]{kopf2021robust}
Johannes Kopf, Xuejian Rong, and Jia-Bin Huang.
\newblock Robust consistent video depth estimation.
\newblock In \emph{CVPR}, 2021.

\bibitem[Lee et~al.(2019)Lee, Han, Ko, and Suh]{lee2019big}
Jin~Han Lee, Myung-Kyu Han, Dong~Wook Ko, and Il~Hong Suh.
\newblock From big to small: Multi-scale local planar guidance for monocular
  depth estimation.
\newblock \emph{arXiv}, 2019.

\bibitem[Li and Lee(2021)]{li2021deepi2p}
Jiaxin Li and Gim~Hee Lee.
\newblock {DeepI2P}: Image-to-point cloud registration via deep classification.
\newblock In \emph{CVPR}, 2021.

\bibitem[Li et~al.(2023)Li, Gong, Yin, Chen, Zhu, Wang, Chen, Sun, and
  Zhang]{li2023learning}
Rui Li, Dong Gong, Wei Yin, Hao Chen, Yu Zhu, Kaixuan Wang, Xiaozhi Chen,
  Jinqiu Sun, and Yanning Zhang.
\newblock Learning to fuse monocular and multi-view cues for multi-frame depth
  estimation in dynamic scenes.
\newblock In \emph{CVPR}, 2023.

\bibitem[Liu et~al.(2022)Liu, Kumar, Gu, Timofte, and Van~Gool]{liu2022va}
Ce Liu, Suryansh Kumar, Shuhang Gu, Radu Timofte, and Luc Van~Gool.
\newblock Va-depthnet: A variational approach to single image depth prediction.
\newblock In \emph{ICLR}, 2022.

\bibitem[Liu et~al.(2020)Liu, Zhang, He, Liu, Wang, Tai, Luo, Wang, Li, and
  Huang]{liu2020learning}
Liang Liu, Jiangning Zhang, Ruifei He, Yong Liu, Yabiao Wang, Ying Tai, Donghao
  Luo, Chengjie Wang, Jilin Li, and Feiyue Huang.
\newblock Learning by analogy: Reliable supervision from transformations for
  unsupervised optical flow estimation.
\newblock In \emph{CVPR}, 2020.

\bibitem[Liu et~al.(2021)Liu, Lin, Cao, Hu, Wei, Zhang, Lin, and
  Guo]{liu2021Swin}
Ze Liu, Yutong Lin, Yue Cao, Han Hu, Yixuan Wei, Zheng Zhang, Stephen Lin, and
  Baining Guo.
\newblock {Swin Transformer}: Hierarchical vision transformer using shifted
  windows.
\newblock In \emph{ICCV}, 2021.

\bibitem[Long et~al.(2021)Long, Liu, Li, Theobalt, and Wang]{long2021multi}
Xiaoxiao Long, Lingjie Liu, Wei Li, Christian Theobalt, and Wenping Wang.
\newblock Multi-view depth estimation using epipolar spatio-temporal networks.
\newblock In \emph{CVPR}, 2021.

\bibitem[Loshchilov and Hutter(2017)]{loshchilov2017decoupled}
Ilya Loshchilov and Frank Hutter.
\newblock Decoupled weight decay regularization.
\newblock \emph{arXiv}, 2017.

\bibitem[Lourakis and Orphanoudakis(1999)]{lourakis1999using}
Manolis~IA Lourakis and Stelios~C Orphanoudakis.
\newblock Using planar parallax to estimate the time-to-contact.
\newblock In \emph{CVPR}, 1999.

\bibitem[Luo et~al.(2020)Luo, Huang, Szeliski, Matzen, and
  Kopf]{luo2020consistent}
Xuan Luo, Jia-Bin Huang, Richard Szeliski, Kevin Matzen, and Johannes Kopf.
\newblock Consistent video depth estimation.
\newblock \emph{ToG}, 2020.

\bibitem[Mallot et~al.(1991)Mallot, B{\"u}lthoff, Little, and Bohrer]{ipm}
Hanspeter~A Mallot, Heinrich~H B{\"u}lthoff, JJ Little, and Stefan Bohrer.
\newblock Inverse perspective mapping simplifies optical flow computation and
  obstacle detection.
\newblock \emph{Biological cybernetics}, 1991.

\bibitem[Mayer et~al.(2016)Mayer, Ilg, Hausser, Fischer, Cremers, Dosovitskiy,
  and Brox]{mayer2016large}
Nikolaus Mayer, Eddy Ilg, Philip Hausser, Philipp Fischer, Daniel Cremers,
  Alexey Dosovitskiy, and Thomas Brox.
\newblock A large dataset to train convolutional networks for disparity,
  optical flow, and scene flow estimation.
\newblock In \emph{CVPR}, 2016.

\bibitem[Meister et~al.(2018)Meister, Hur, and Roth]{meister2018unflow}
Simon Meister, Junhwa Hur, and Stefan Roth.
\newblock {UnFlow}: Unsupervised learning of optical flow with a bidirectional
  census loss.
\newblock In \emph{AAAI}, 2018.

\bibitem[Menze and Geiger(2015)]{menze2015object}
Moritz Menze and Andreas Geiger.
\newblock Object scene flow for autonomous vehicles.
\newblock In \emph{CVPR}, 2015.

\bibitem[Paszke et~al.(2019)Paszke, Gross, Massa, Lerer, Bradbury, Chanan,
  Killeen, Lin, Gimelshein, Antiga, et~al.]{paszke2019pytorch}
Adam Paszke, Sam Gross, Francisco Massa, Adam Lerer, James Bradbury, Gregory
  Chanan, Trevor Killeen, Zeming Lin, Natalia Gimelshein, Luca Antiga, et~al.
\newblock Pytorch: An imperative style, high-performance deep learning library.
\newblock \emph{NeurIPS}, 2019.

\bibitem[Piccinelli et~al.(2023)Piccinelli, Sakaridis, and
  Yu]{piccinelli2023idisc}
Luigi Piccinelli, Christos Sakaridis, and Fisher Yu.
\newblock idisc: Internal discretization for monocular depth estimation.
\newblock In \emph{CVPR}, 2023.

\bibitem[Ranftl et~al.(2021)Ranftl, Bochkovskiy, and Koltun]{Ranftl_2021_ICCV}
Ren\'e Ranftl, Alexey Bochkovskiy, and Vladlen Koltun.
\newblock Vision transformers for dense prediction.
\newblock In \emph{ICCV}, 2021.

\bibitem[Sawhney(1994{\natexlab{a}})]{sawhney19943d}
Harpreet~S Sawhney.
\newblock {3D} geometry from planar parallax.
\newblock In \emph{CVPR}, 1994{\natexlab{a}}.

\bibitem[Sawhney(1994{\natexlab{b}})]{sawhney1994motion}
Harpreet~S Sawhney.
\newblock Motion video analysis using planar parallax.
\newblock In \emph{Storage and Retrieval for Image and Video Databases II},
  1994{\natexlab{b}}.

\bibitem[Sawhney(1994{\natexlab{c}})]{sawhney1994simplifying}
Harpreet~S Sawhney.
\newblock Simplifying motion and structure analysis using planar parallax and
  image warping.
\newblock In \emph{ICPR}, 1994{\natexlab{c}}.

\bibitem[Shashua and Navab(1994)]{shashua1994relative}
Shashua and Navab.
\newblock Relative affine structure: theory and application to {3D}
  reconstruction from perspective views.
\newblock In \emph{CVPR}, 1994.

\bibitem[Sun et~al.(2018)Sun, Yang, Liu, and Kautz]{pwcnet}
Deqing Sun, Xiaodong Yang, Ming-Yu Liu, and Jan Kautz.
\newblock {PWC-Net}: {CNN}s for optical flow using pyramid, warping, and cost
  volume.
\newblock In \emph{CVPR}, 2018.

\bibitem[Sun et~al.(2020)Sun, Kretzschmar, Dotiwalla, Chouard, Patnaik, Tsui,
  Guo, Zhou, Chai, Caine, et~al.]{sun2020scalability}
Pei Sun, Henrik Kretzschmar, Xerxes Dotiwalla, Aurelien Chouard, Vijaysai
  Patnaik, Paul Tsui, James Guo, Yin Zhou, Yuning Chai, Benjamin Caine, et~al.
\newblock Scalability in perception for autonomous driving: Waymo open dataset.
\newblock In \emph{CVPR}, 2020.

\bibitem[Teed and Deng(2020{\natexlab{a}})]{raft}
Zachary Teed and Jia Deng.
\newblock {RAFT}: Recurrent all-pairs field transforms for optical flow.
\newblock In \emph{ECCV}, 2020{\natexlab{a}}.

\bibitem[Teed and Deng(2020{\natexlab{b}})]{teeddeepv2d}
Zachary Teed and Jia Deng.
\newblock Deepv2d: Video to depth with differentiable structure from motion.
\newblock In \emph{ICLR}, 2020{\natexlab{b}}.

\bibitem[Vaish et~al.(2004)Vaish, Wilburn, Joshi, and Levoy]{vaish2004using}
Vaibhav Vaish, Bennett Wilburn, Neel Joshi, and Marc Levoy.
\newblock Using plane+parallax for calibrating dense camera arrays.
\newblock In \emph{CVPR}, 2004.

\bibitem[Wang et~al.(2022)Wang, Pang, and Lin]{wang2022dfm}
Tai Wang, Jiangmiao Pang, and Dahua Lin.
\newblock Monocular {3D} object detection with depth from motion.
\newblock In \emph{ECCV}, 2022.

\bibitem[Wang et~al.(2018)Wang, Yang, Yang, Zhao, Wang, and
  Xu]{wang2018occlusion}
Yang Wang, Yi Yang, Zhenheng Yang, Liang Zhao, Peng Wang, and Wei Xu.
\newblock Occlusion aware unsupervised learning of optical flow.
\newblock In \emph{CVPR}, 2018.

\bibitem[Xing et~al.(2022)Xing, Cao, Biber, Zhou, and Burschka]{xing2022joint}
Hao Xing, Yifan Cao, Maximilian Biber, Mingchuan Zhou, and Darius Burschka.
\newblock Joint prediction of monocular depth and structure using planar and
  parallax geometry.
\newblock \emph{PR}, 2022.

\bibitem[Xu et~al.(2022)Xu, Zhang, Cai, Rezatofighi, and Tao]{xu2022gmflow}
Haofei Xu, Jing Zhang, Jianfei Cai, Hamid Rezatofighi, and Dacheng Tao.
\newblock {GMFlow}: Learning optical flow via global matching.
\newblock In \emph{CVPR}, 2022.

\bibitem[Yang et~al.(2023)Yang, Ma, Ji, and Ren]{yang2023gedepth}
Xiaodong Yang, Zhuang Ma, Zhiyu Ji, and Zhe Ren.
\newblock Gedepth: Ground embedding for monocular depth estimation.
\newblock In \emph{ICCV}, 2023.

\bibitem[Yin et~al.(2019)Yin, Liu, Shen, and Yan]{yin2019enforcing}
Wei Yin, Yifan Liu, Chunhua Shen, and Youliang Yan.
\newblock Enforcing geometric constraints of virtual normal for depth
  prediction.
\newblock In \emph{ICCV}, 2019.

\bibitem[Yin et~al.(2021)Yin, Liu, and Shen]{yin2021virtual}
Wei Yin, Yifan Liu, and Chunhua Shen.
\newblock Virtual normal: Enforcing geometric constraints for accurate and
  robust depth prediction.
\newblock \emph{PAMI}, 2021.

\bibitem[Yuan et~al.(2007)Yuan, Medioni, Kang, and Cohen]{yuan2007detecting}
Chang Yuan, Gerard Medioni, Jinman Kang, and Isaac Cohen.
\newblock Detecting motion regions in the presence of a strong parallax from a
  moving camera by multiview geometric constraints.
\newblock \emph{PAMI}, 2007.

\bibitem[Yuan et~al.(2021)Yuan, Chen, Sui, Xie, Zhang, Li, and
  Zhang]{yuan2021monocular}
Haobo Yuan, Teng Chen, Wei Sui, Jiafeng Xie, Lefei Zhang, Yuan Li, and Qian
  Zhang.
\newblock Monocular road planar parallax estimation.
\newblock \emph{arXiv}, 2021.

\bibitem[Yuan et~al.(2022)Yuan, Gu, Dai, Zhu, and Tan]{yuan2022newcrfs}
Weihao Yuan, Xiaodong Gu, Zuozhuo Dai, Siyu Zhu, and Ping Tan.
\newblock {NeWCRFs}: Neural window fully-connected {CRFs} for monocular depth
  estimation.
\newblock In \emph{CVPR}, 2022.

\bibitem[Zhao et~al.(2020)Zhao, Sheng, Dong, Chang, Xu,
  et~al.]{zhao2020maskflownet}
Shengyu Zhao, Yilun Sheng, Yue Dong, Eric~I Chang, Yan Xu, et~al.
\newblock {MaskFlowNet}: Asymmetric feature matching with learnable occlusion
  mask.
\newblock In \emph{CVPR}, 2020.

\bibitem[Zhong et~al.(2019)Zhong, Ji, Wang, Dai, and Li]{zhong2019unsupervised}
Yiran Zhong, Pan Ji, Jianyuan Wang, Yuchao Dai, and Hongdong Li.
\newblock Unsupervised deep epipolar flow for stationary or dynamic scenes.
\newblock In \emph{CVPR}, 2019.

\bibitem[Zhu and Liu(2023)]{zhu2023lighteddepth}
Shengjie Zhu and Xiaoming Liu.
\newblock Lighteddepth: Video depth estimation in light of limited inference
  view angles.
\newblock In \emph{CVPR}, 2023.

\end{thebibliography}
}


\newpage
\appendix
%%%%%%%%% BODY TEXT

\section{Planar Parallax Geometry}

We provide a complete derivation process in this section. We use capital letters to represent 3D points, lowercase letters for 2D points, bold font for vectors, and matrices in calligraphy.

The ratio of height to depth $\gamma$ is defined as: 
\begin{equation}
	\gamma = \frac{h}{z},
\end{equation}
where $h$ and $z$ is the height and depth of a pixel.

In paper's Fig.~\ref{fig:pp}, we present the geometry visually. Define $\mathbf{P}_s = (x', y', z')^T$ and $\mathbf{P}_t =(x, y, z)^T$ as the coordinates of a point $\mathbf{P}$ in source view and target view, separately. Let $\mathcal{R}$ and $\mathbf{T}=(t_x, t_y, t_z)^T$ denote the rotation matrix and translation vector between the two camera views. The transformation from $\mathbf{P}_s$ to $\mathbf{P}_t$ can be written as:
\begin{equation}  \label{appendix_eq:ps2pt}
\mathbf{P}_t = \mathcal{R}\mathbf{P}_s + \mathbf{T}.
\end{equation}

The height above the reference plane $\pi$ of the point $\mathbf{P}$ can be expressed as:
\begin{equation}  \label{appendix_eq:height}
	h = h_c - \vec{\mathbf{N}}^T\mathbf{P},
\end{equation}
where $\vec{\mathbf{N}}^T$ is the normal of the plane $\pi$ and $h_c$ is the height of the camera. Eqn.~\ref{appendix_eq:height} can be transformed into:
\begin{equation}
	\frac{h + \vec{\mathbf{N}}^T\mathbf{P}}{h_c} = 1.
\end{equation}

By multiply $\mathbf{T}$ by $1$ in Eqn.~\ref{appendix_eq:ps2pt}, we can obtain
\begin{equation}
  \begin{split}
	\mathbf{P}_t &= \mathcal{R}\mathbf{P}_s + \mathbf{T}\frac{h + \vec{\mathbf{N}}^T\mathbf{P}_s}{h_c}    \\
	             &= (\mathcal{R} + \frac{\mathbf{T}\vec{\mathbf{N}}^T}{h_c})\mathbf{P}_s + \frac{h}{h_c}\mathbf{T}
  \end{split}
\end{equation}

Let $\mathbf{p}_s=\frac{1}{z'}\mathcal{K}\mathbf{P}_s$, $\mathbf{p}_t=\frac{1}{z}\mathcal{K}\mathbf{P}_t$ and $\mathbf{t}=\mathcal{K}\mathbf{T}$, where $\mathcal{K}$ is intrinsic matrix of the camera. Then we can obtain
\begin{equation}
	z\mathcal{K}^{-1}\mathbf{p}_t = (\mathcal{R} + \frac{\mathbf{T}\vec{\mathbf{N}}^T}{h_c})z'\mathcal{K}^{-1}\mathbf{p}_s + \frac{h}{h_c}\mathbf{T}
\end{equation}

By mutiply $\frac{1}{z'}K$ on both sides, we have:
\begin{equation}  \label{appendix_eq:zpt}
	\frac{z}{z'}\mathbf{p}_t = \mathcal{K}(\mathcal{R} + \frac{\mathbf{T}\vec{\mathbf{N}}^T}{h_c})\mathcal{K}^{-1}\mathbf{p}_s + \frac{h}{h_cz'}\mathbf{t}.
\end{equation}

With the homography matrix between the two images written as:
\begin{equation}
  \mathcal{H} = \mathcal{K}(\mathcal{R} + \frac{\mathbf{T}\vec{\mathbf{N}}^T}{h_c})\mathcal{K}^{-1},
\end{equation}
Eqn.~\ref{appendix_eq:zpt} can be reformulated as
\begin{equation}
	\frac{z}{z'}\mathbf{p}_t = \mathcal{H}\mathbf{p}_s + \frac{h}{h_cz'}\mathbf{t}.
\end{equation}

By considering the z-axis of both sides, we have:
\begin{equation}
	\frac{z}{z'} = \mathcal{H}_3\mathbf{p}_s + \frac{ht_z}{h_cz'},
\end{equation}
where $\mathcal{H}_3$ denote the third row of homography matrix $\mathcal{H}$

Note that, the z-axis of $\mathbf{p}_s$ and $\mathbf{p}_t$ is 1. Scaling both sides by their z-axis, we can obtain
\begin{equation}
  \begin{split}
	\mathbf{p}_t &= \frac{\mathcal{H}\mathbf{p}_s + \frac{h}{h_cz'}\mathbf{t}}{\mathcal{H}_3\mathbf{p}_s + \frac{ht_z}{h_cz'}}    \\
	    &= \frac{\mathcal{H}\mathbf{p}_s}{\mathcal{H}_3\mathbf{p}_s} - \frac{\mathcal{H}\mathbf{p}_s}{\mathcal{H}_3\mathbf{p}_s} + \frac{\mathcal{H}\mathbf{p}_s + \frac{h}{h_cz'}\mathbf{t}}{\mathcal{H}_3\mathbf{p}_s + \frac{ht_z}{h_cz'}}    \\
	    &= \frac{\mathcal{H}\mathbf{p}_s}{\mathcal{H}_3\mathbf{p}_s} - \frac{\frac{ht_z}{h_cz'}}{(\mathcal{H}_3\mathbf{p}_s + \frac{ht_z}{h_cz'})} \frac{\mathcal{H}\mathbf{p}_s}{\mathcal{H}_3\mathbf{p}_s} + \frac{\frac{h}{h_cz'}\mathbf{t}}{\mathcal{H}_3\mathbf{p}_s + \frac{ht_z}{h_cz'}}    \\
	    &= \frac{\mathcal{H}\mathbf{p}_s}{\mathcal{H}_3\mathbf{p}_s} - \frac{ht_z}{zh_c} \frac{\mathcal{H}\mathbf{p}_s}{\mathcal{H}_3\mathbf{p}_s} + \frac{h}{h_cz}\mathbf{t}.
  \end{split}
\end{equation}

With epipole $\mathbf{e}_t=\frac{1}{t_z}\mathbf{t}$, $\gamma=\frac{h}{z}$, $\mathbf{p}_s$ warped by homography $\mathbf{p}_w=\frac{\mathcal{H}\mathbf{p}_s}{\mathcal{H}_3\mathbf{p}_s}$, when $t_z = 0$, we have
\begin{equation}
	\mathbf{p}_t = \mathbf{p}_w + \frac{h}{h_cz}\mathbf{t}.
\end{equation}

When $t_z \neq 0$, we have
\begin{equation}
	\mathbf{p}_t = \mathbf{p}_w - \gamma\frac{t_z}{h_c} (\mathbf{p}_w - \mathbf{e}_t).
\end{equation}

Then we can obtain
\begin{equation}
	\mathbf{p}_w - \mathbf{p}_t = \gamma\frac{t_z}{h_c} (\mathbf{p}_w - \mathbf{e}_t),
\end{equation}
which can also be converted to
\begin{gather}
	\mathbf{p}_w - \mathbf{p}_t = \gamma\frac{t_z}{h_c} (\mathbf{p}_w - \mathbf{p}_t + \mathbf{p}_t - \mathbf{e}_t) \\
	(1 - \gamma\frac{t_z}{h_c})(\mathbf{p}_w - \mathbf{p}_t) = \gamma\frac{t_z}{h_c} (\mathbf{p}_t - \mathbf{e}_t)    \\
	\mathbf{p}_w - \mathbf{p}_t = \frac{\gamma\frac{t_z}{h_c}}{1 - \gamma\frac{t_z}{h_c}} (\mathbf{p}_t - \mathbf{e}_t)
\end{gather}
Now we get the relationship between $\mathbf{u}_{res}=\mathbf{p}_w - \mathbf{p}_t$ and $\gamma$.

Except for the relationship with the residual flow, $\gamma$ can also be used for 3D reconstruction. Since $\mathbf{P}_t$ can be calculated by an inverse projection 
\begin{equation}	\label{appendix_eq:p2P}
	\mathbf{P}_t = z\mathcal{K}^{-1}\mathbf{p}_t.
\end{equation}

By substituting it into Eqn.~\ref{appendix_eq:height}, we can obtain an important formula discussed in the proposed Planar Position Embedding.

\begin{equation} \label{appendix_eq:nkp}
	\vec{\mathbf{N}}^T(\mathcal{K}^{-1}\mathbf{p}_t) = \frac{h_c - h}{z}.
\end{equation}

Eqn.~\ref{appendix_eq:nkp} can be finally transformed into
\begin{equation}	\label{appendix_eq:gamma2Z}
	z = \frac{h_c}{\gamma + \vec{\mathbf{N}}^T(\mathcal{K}^{-1}\mathbf{p}_t)}.
\end{equation}

We could use it to convert predicted $\gamma$ to depth results given the plane and camera height above the plane.

\begin{table}[tb]
    \setlength{\tabcolsep}{1mm}{
	\begin{center}  
	\small
	\begin{tabular}{lcccccc}  
	\hline  
	$w_{d}$ & Abs Rel $\downarrow$ & Sq Rel $\downarrow$ & RMSE $\downarrow$ & RMSE log $\downarrow$ & $\delta_1$ $\uparrow$    \\  \hline
	
	$1$ & 0.040 & 0.118 & 1.881 & 0.066 & 0.980	\\
	$10^{-1}$ & 0.038 & 0.110 & 1.828 & 0.063 & 0.982	\\
	$10^{-2}$ & \textbf{0.037} & \textbf{0.109} & \textbf{1.815} & \textbf{0.062} & \textbf{0.983} \\
	$10^{-3}$ & \textbf{0.037} & 0.113 & 1.844 & 0.064 & 0.981 \\
	w/o DL & \textbf{0.037} & 0.117 & 1.878 & 0.064 & 0.982	\\
	\hline
	\end{tabular}
	\end{center}  
	}
	\caption{\label{appendix_tab:dl}Ablation studies for weight of depth loss.}
\end{table}

\begin{figure}[tb]
    \centering
    \begin{minipage}{0.49\linewidth}
    	\centering
    	\includegraphics[width=1\textwidth]{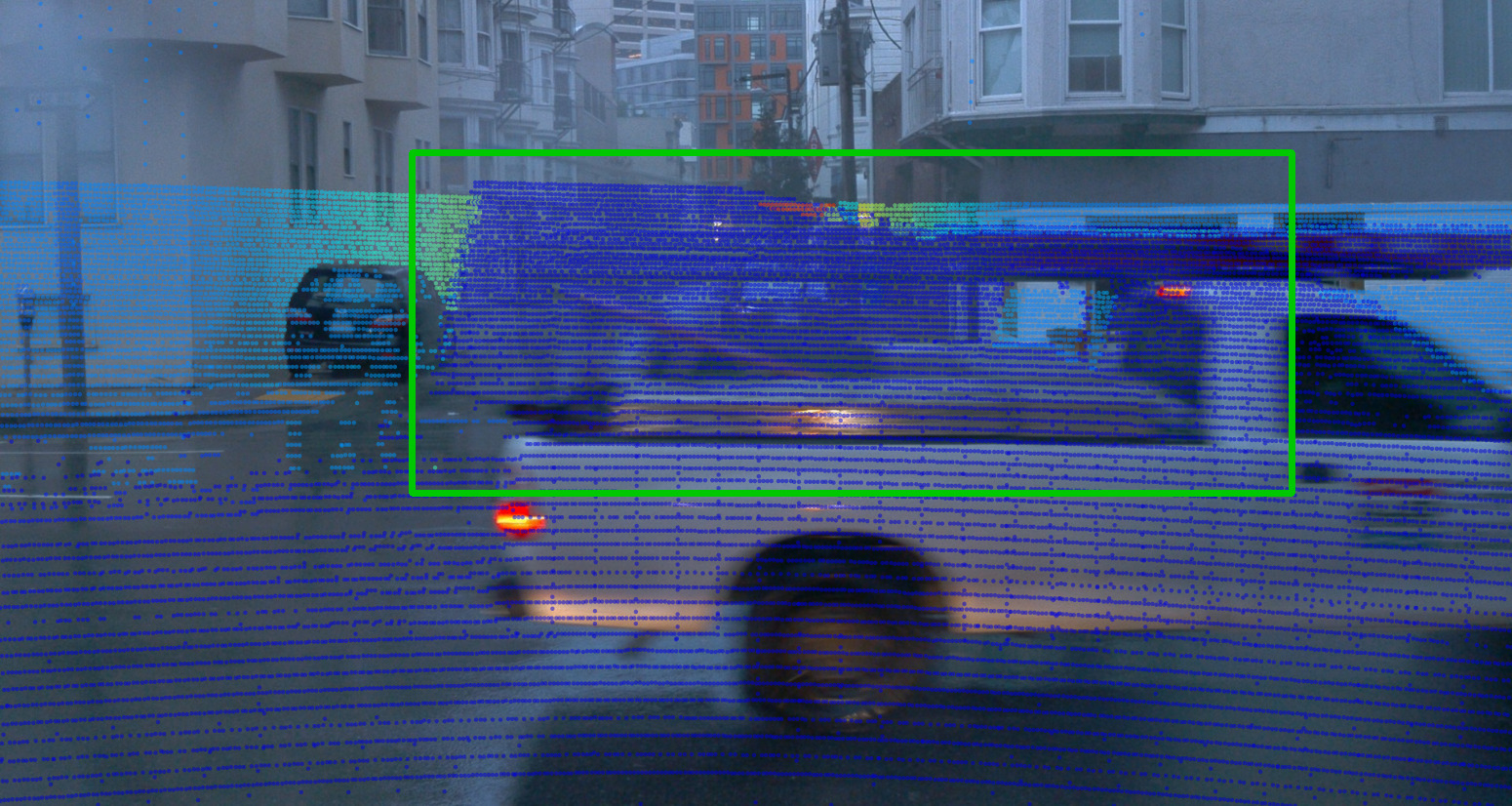}
    	\includegraphics[width=1\textwidth]{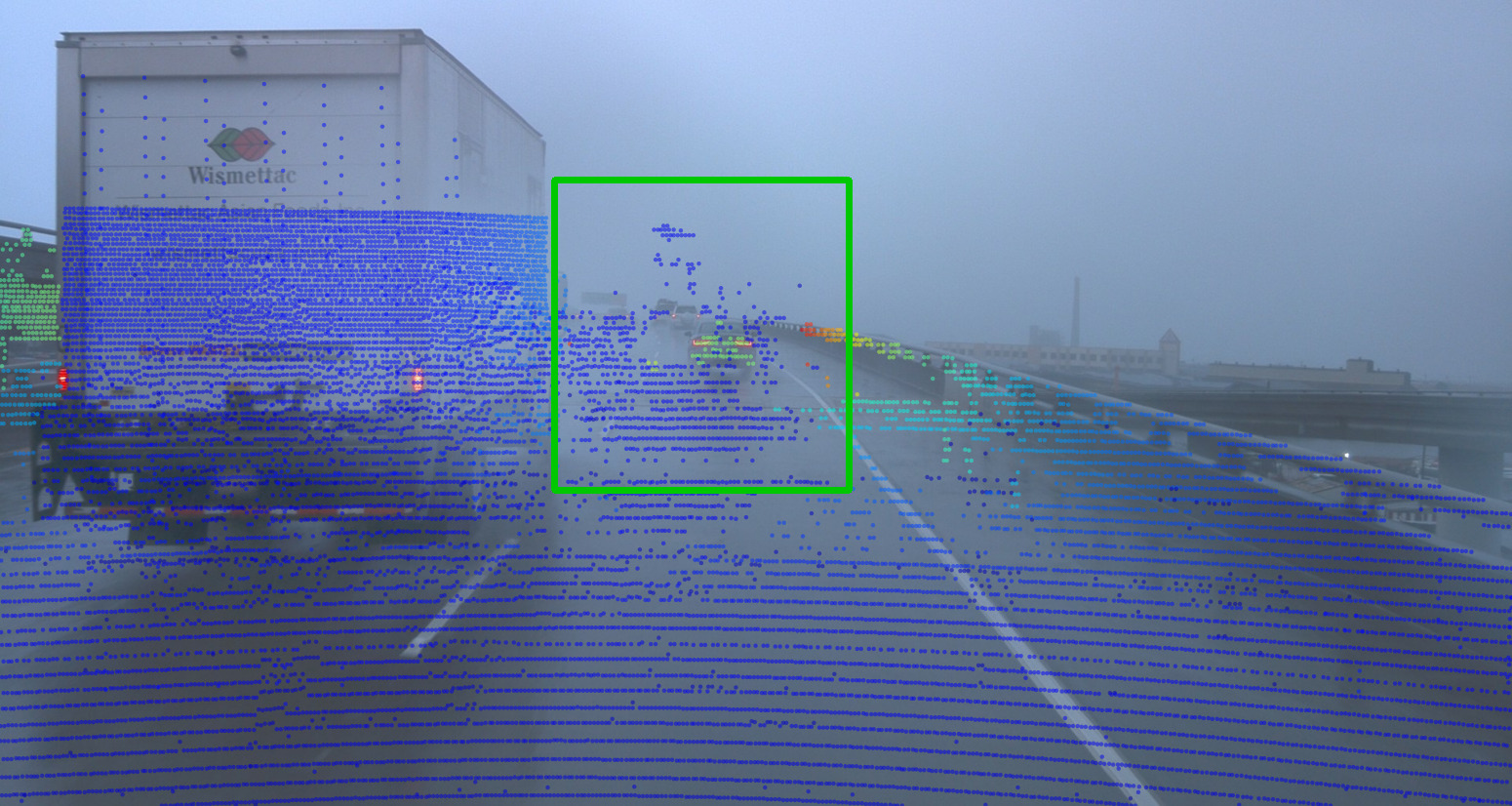}
    \end{minipage}
    \begin{minipage}{0.49\linewidth}
    	\centering
    	\includegraphics[width=1\textwidth]{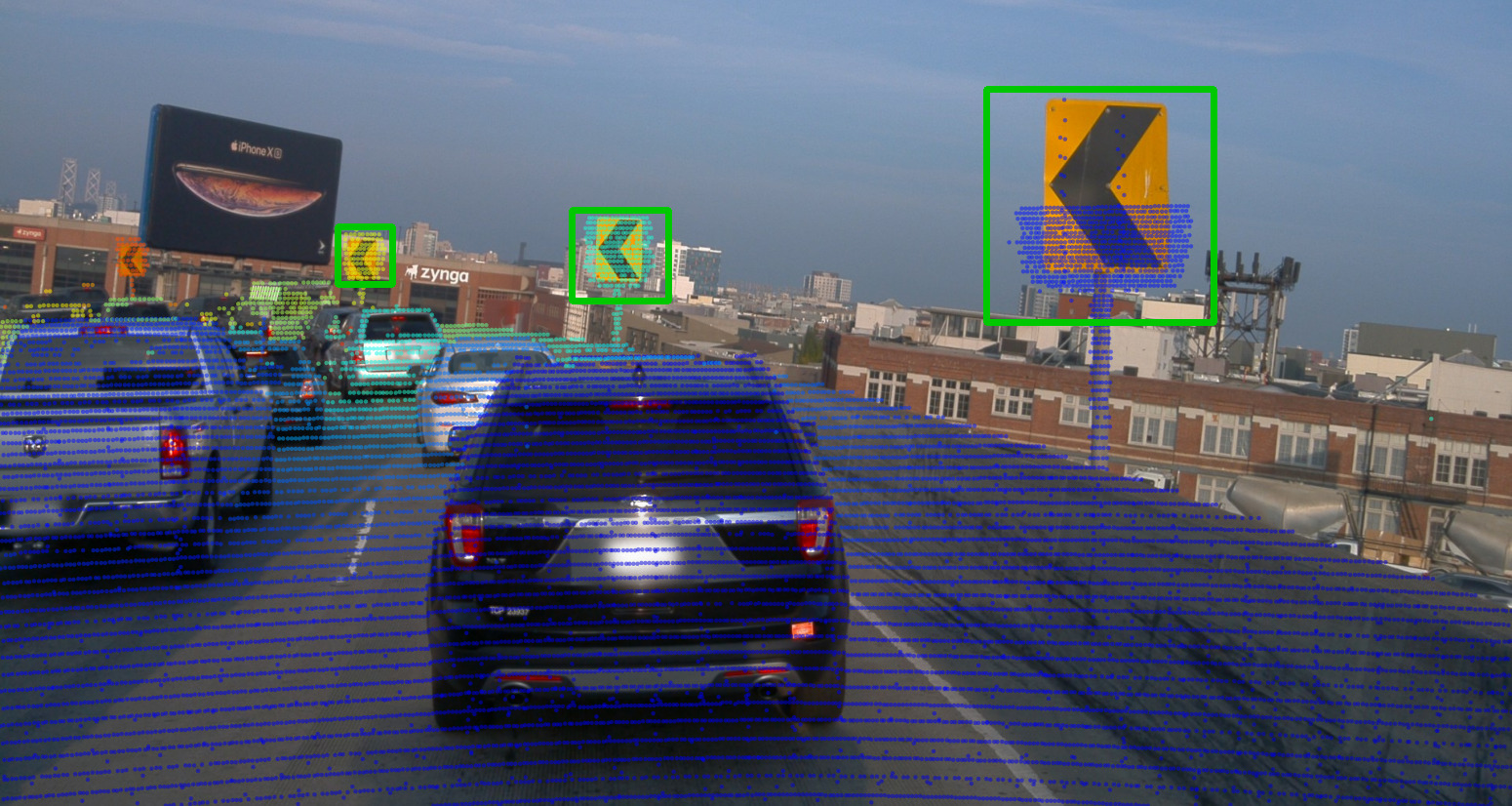}
    	\includegraphics[width=1\textwidth]{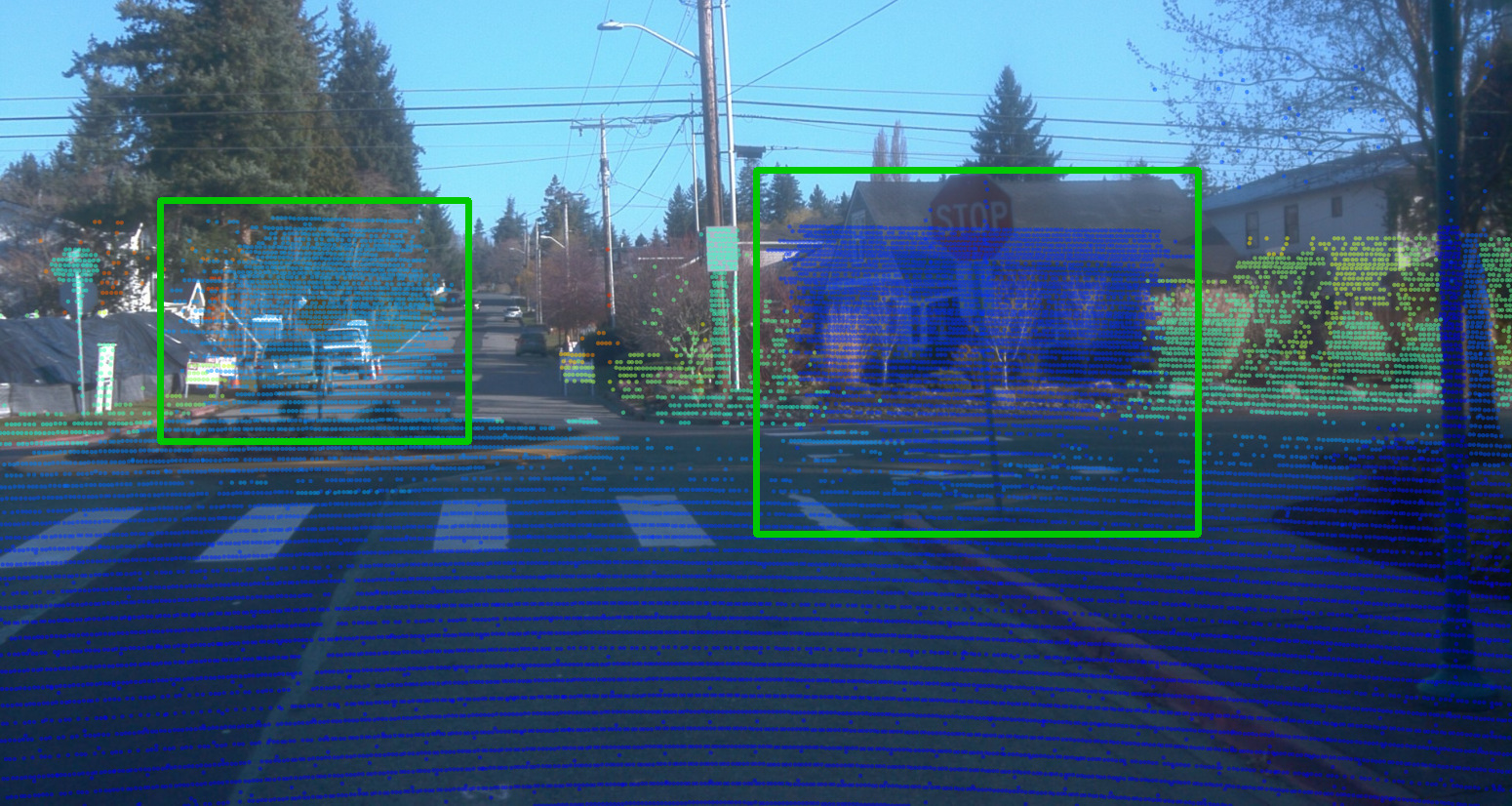}
    \end{minipage}
    \newline
    
    \caption{Mismatch examples of Waymo Open Dataset. Left Up: motion distortion. Right Up: noise of high reflectance. Left Bottom: rainy noise. Right Bottom: unknown noise.}
    \label{appendix_fig:waymo_mismatch}
\end{figure}

\section{Loss Weight}

Adding additional depth supervision can make the network learn the depth information directly. However, with $\gamma$ appearing at the denominator in Eqn.~\ref{appendix_eq:gamma2Z}, the predicted depth can be very unstable, and this can produce unexpected gradients. We select the loss weight by grid search from $[1, 10^{-1}, 10^{-2}, 10^{-3}]$ as shown in Tab.~\ref{appendix_tab:dl}.

\section{Mismatches in Waymo Open Dataset}

While experiments conducted on WOD~\cite{sun2020scalability} demonstrate the generalization of the methods, it is worth considering that the ground-truth depth obtained from LiDAR may contain some unexpected errors if not processed deliberately. For instance, as illustrated in Fig.~\ref{appendix_fig:waymo_mismatch}, objects with high reflectance or motion distortion will result in mismatches between the observations from the image and LiDAR.

\begin{figure}[tb]
  \centering
  \includegraphics[width=1\linewidth]{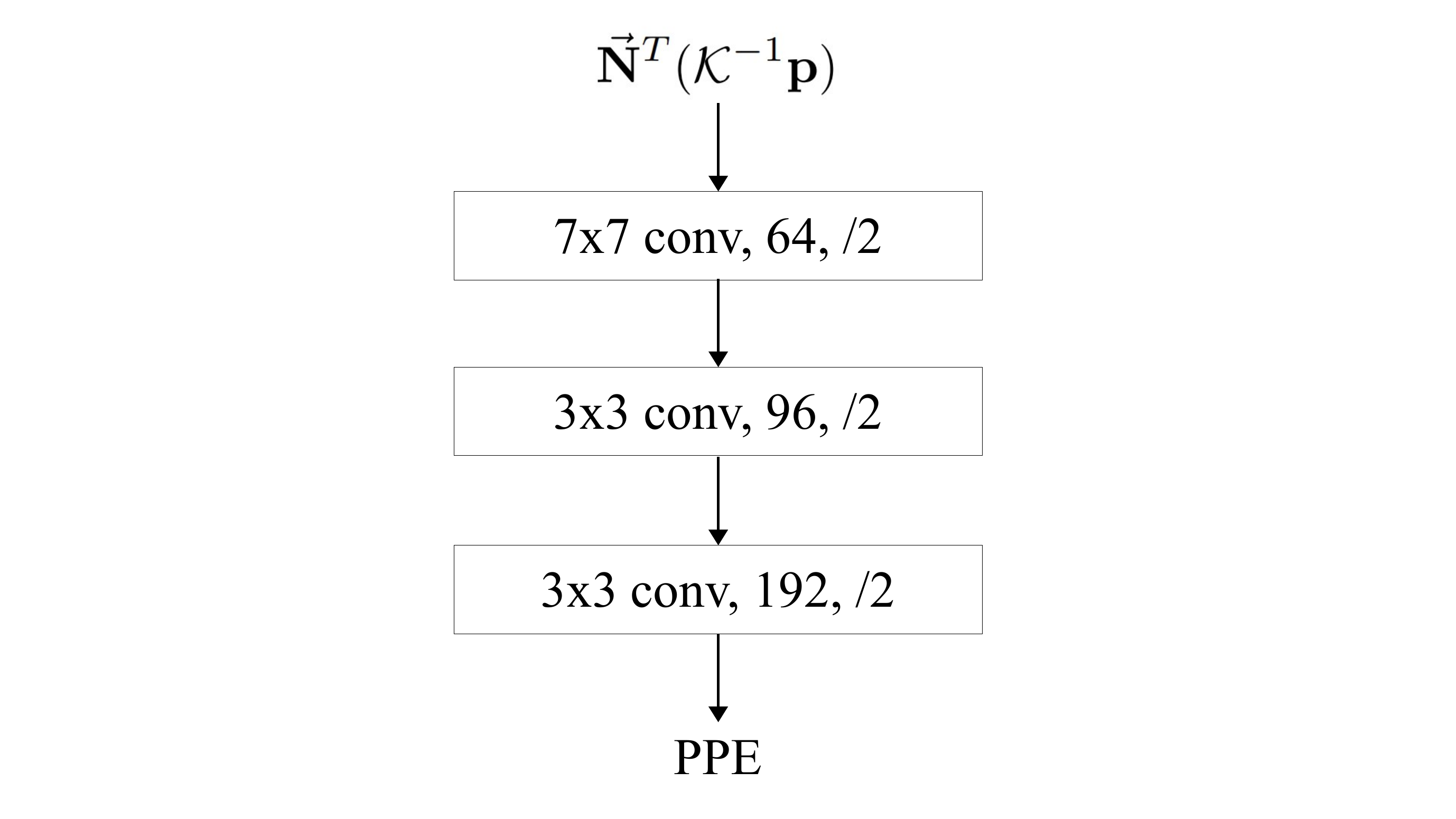}
  \caption{The structure of Planar Position Embedding.}
  \label{appendix_fig:ppe_structure}
\end{figure}

\section{More Implementation Details}

\noindent
\textbf{Planar Position Embedding.} Fig.~\ref{appendix_fig:ppe_structure} shows the detailed structure of the simple convolutional network in the proposed Planar Position Embedding. We adopt batch normalization (BN) right after each convolution and before activation, following~\cite{he2016deep}. The output is then introduced into the network after the fusion of single-frame and multi-frame features.

\noindent
\textbf{Refinement.} We perform the refinement following~\cite{flownet2}. We apply the `UpConvolution' to feature maps, and concatenate it with an upsampled coarser depth prediction. We repeat this three times, resulting in a predicted depth map with the same resolution as the input image.

\section{Multi-frame Extension}

The extended model takes three consecutive images $I_{t-1}$, $I_{t}$ and $I_{t+1}$ as input (which can also be replaced with $I_{t-2}$, $I_{t-1}$ and $I_{t}$). Except for $I_t$, all other two images are warped using road plane homography. We first extract features of the three images using the same Swin-Tiny backbone. The single frame branch remains the same since it is only related to $I_t$. Since the feature enhancement transformer only takes two features as input, we make the three images into two pairs ($I_{t-1}$, $I_{t}$) and ($I_{t+1}$, $I_{t}$). Both pairs are fed into the same feature enhancement transformer. The outputs of the two pairs are first concatenated and then fused together by a simple convolutional layer. The rest of the network remains the same as the two-frame network.

\section{More Qualitative Results}

\begin{figure*}[tb]
	\centering
	
	\begin{minipage}{0.24\linewidth}
		\includegraphics[width=1\textwidth]{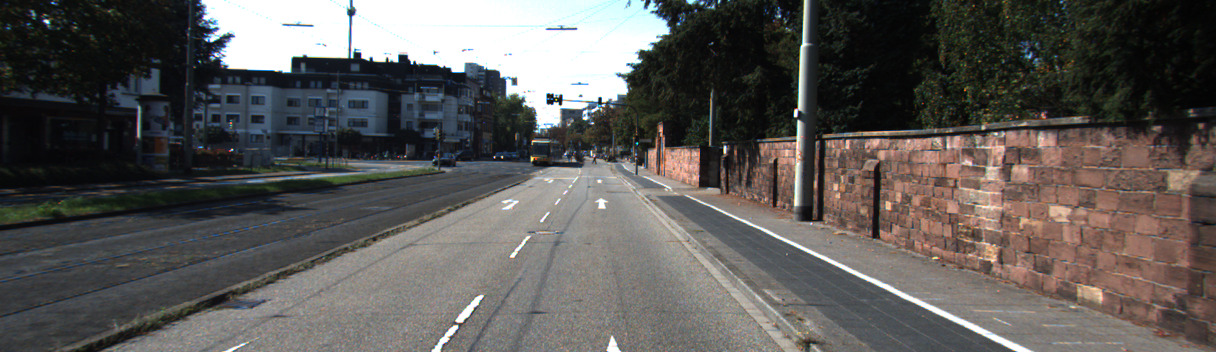}
		\includegraphics[width=1\textwidth]{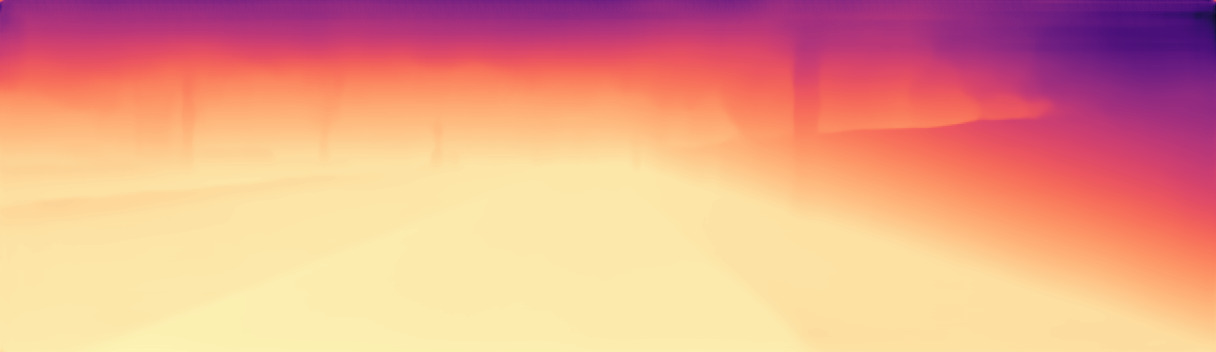}
	\end{minipage}
	\begin{minipage}{0.24\linewidth}
		\includegraphics[width=1\textwidth]{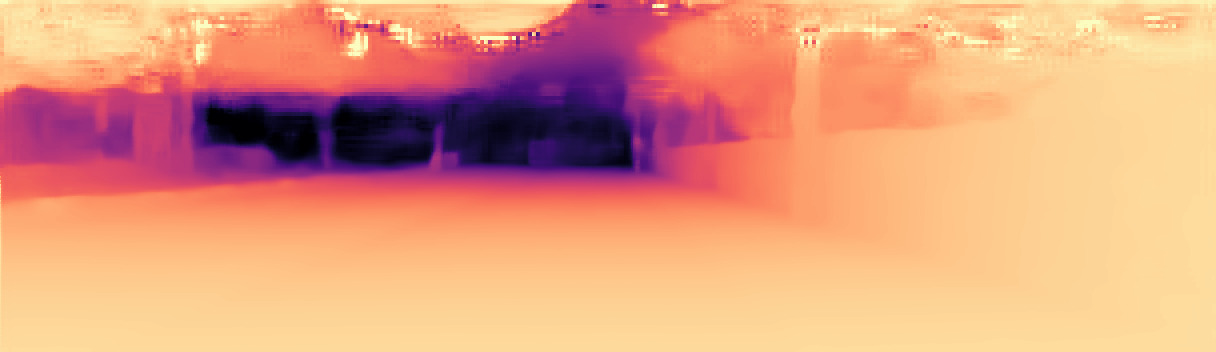}
		\includegraphics[width=1\textwidth]{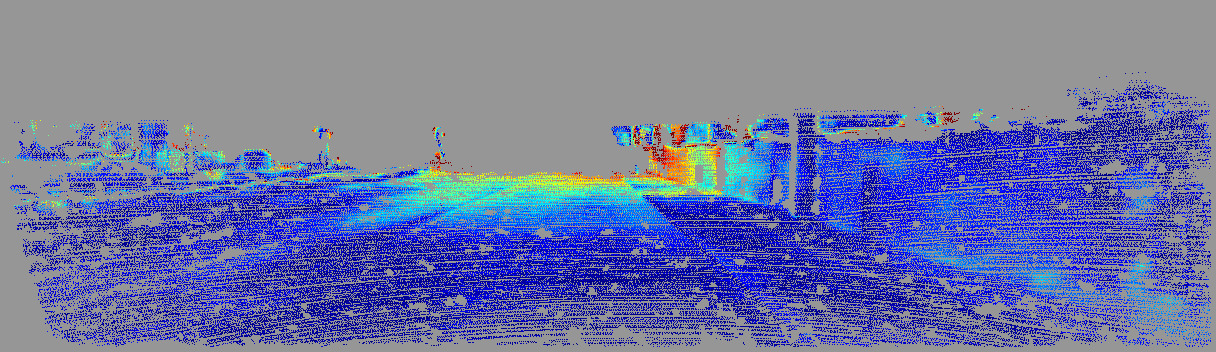}
	\end{minipage}
	\begin{minipage}{0.24\linewidth}
		\includegraphics[width=1\textwidth]{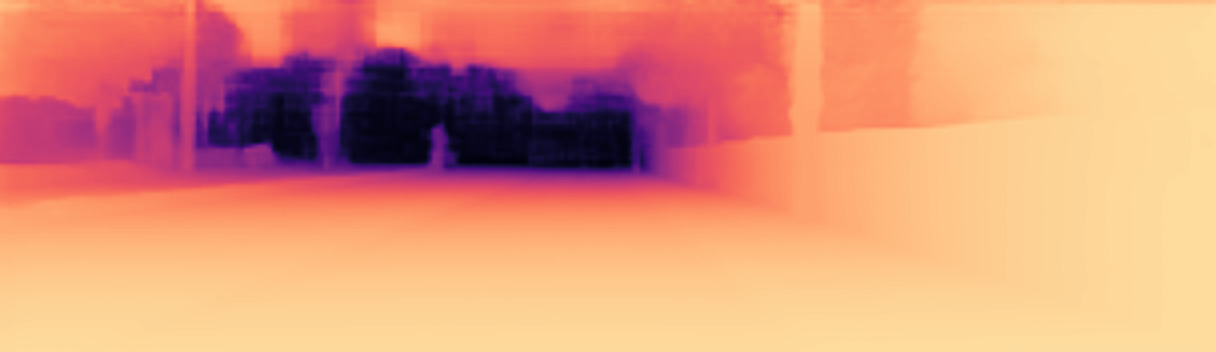}
		\includegraphics[width=1\textwidth]{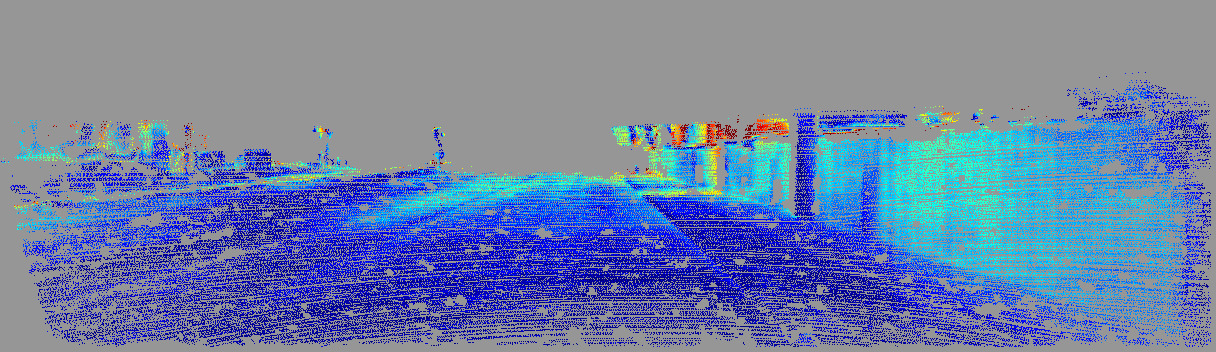}
	\end{minipage}
	\begin{minipage}{0.24\linewidth}
		\includegraphics[width=1\textwidth]{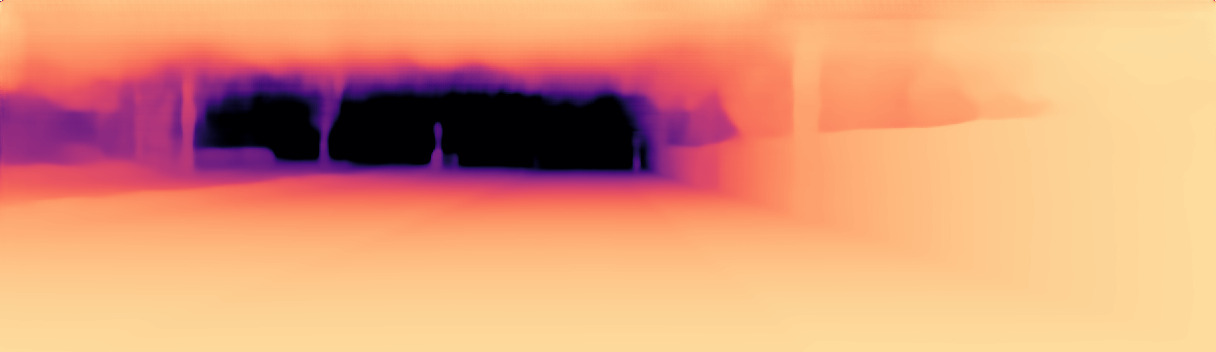}
		\includegraphics[width=1\textwidth]{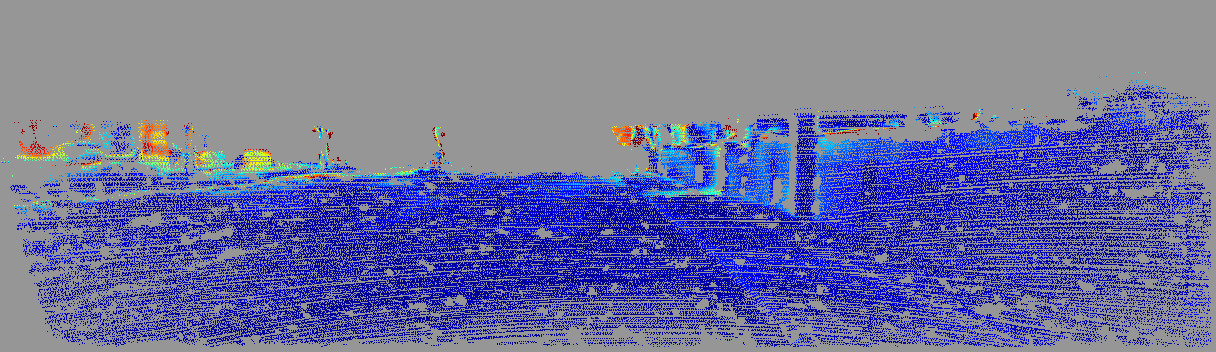}
	\end{minipage}
	
	\begin{minipage}{0.24\linewidth}
		\includegraphics[width=1\textwidth]{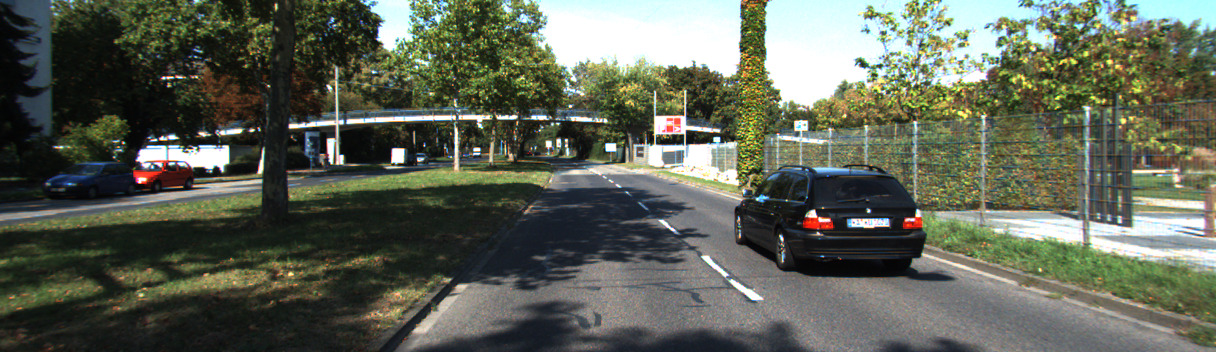}
		\includegraphics[width=1\textwidth]{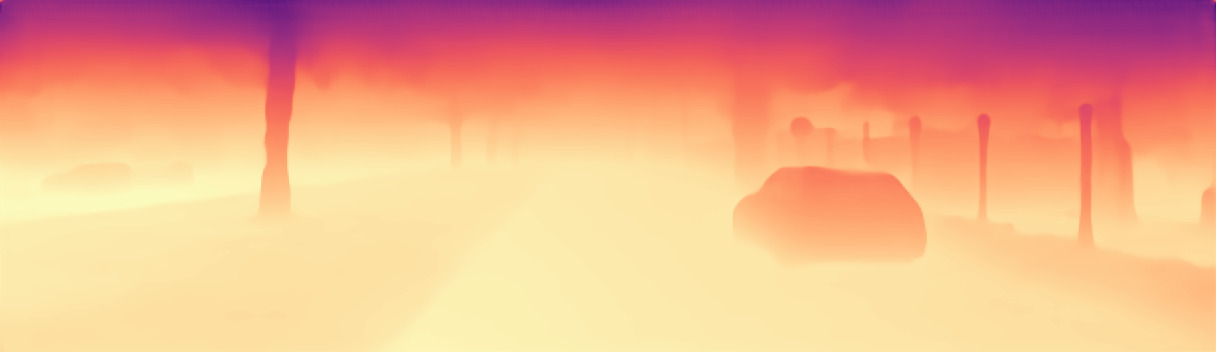}
	\end{minipage}
	\begin{minipage}{0.24\linewidth}
		\includegraphics[width=1\textwidth]{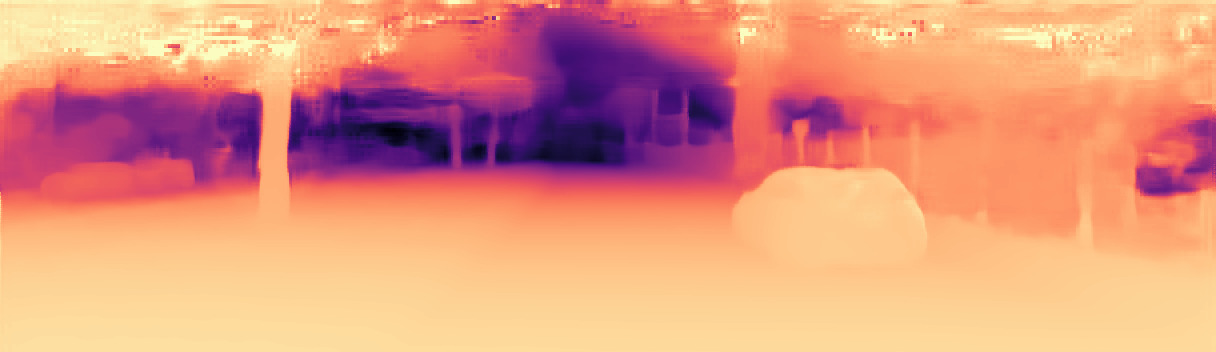}
		\includegraphics[width=1\textwidth]{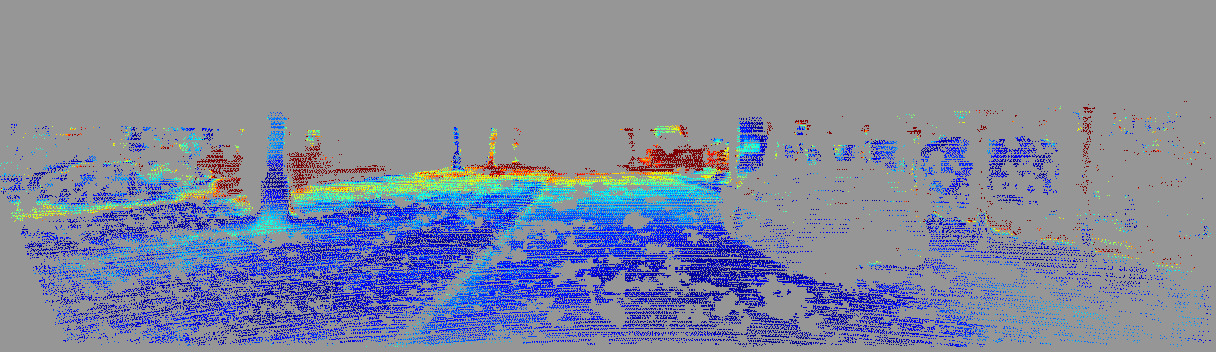}
	\end{minipage}
	\begin{minipage}{0.24\linewidth}
		\includegraphics[width=1\textwidth]{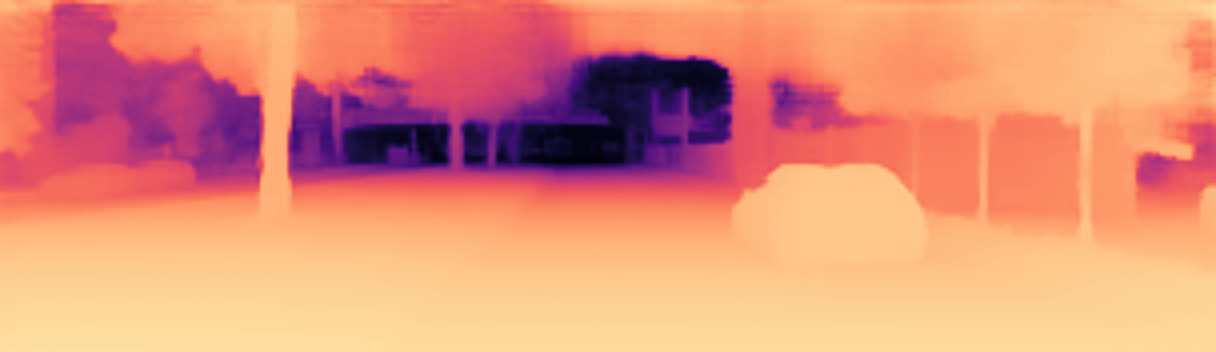}
		\includegraphics[width=1\textwidth]{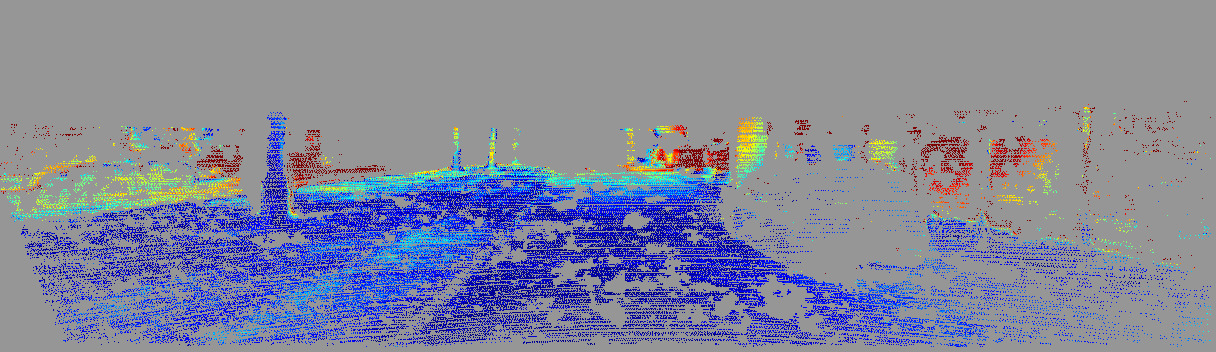}
	\end{minipage}
	\begin{minipage}{0.24\linewidth}
		\includegraphics[width=1\textwidth]{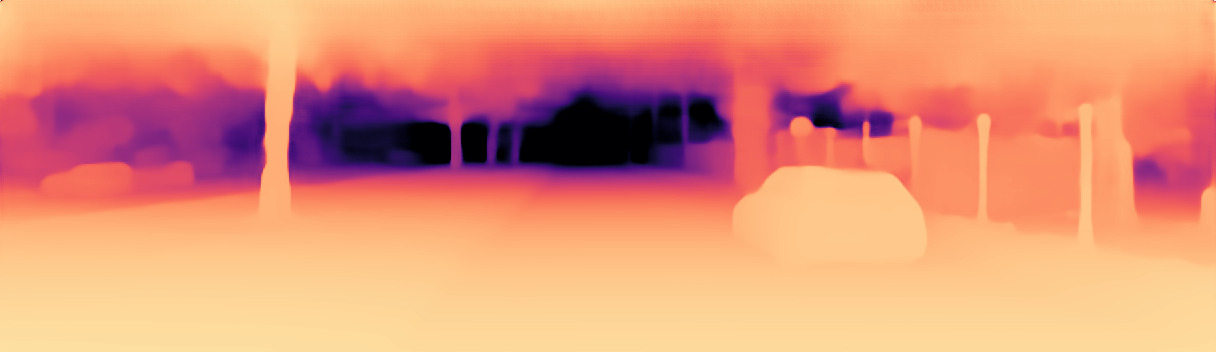}
		\includegraphics[width=1\textwidth]{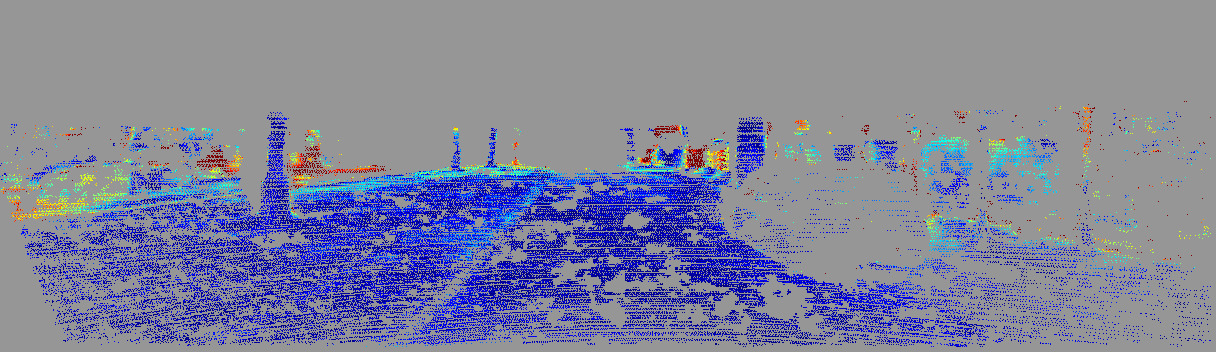}
	\end{minipage}
	
	\begin{minipage}{0.24\linewidth}
		\includegraphics[width=1\textwidth]{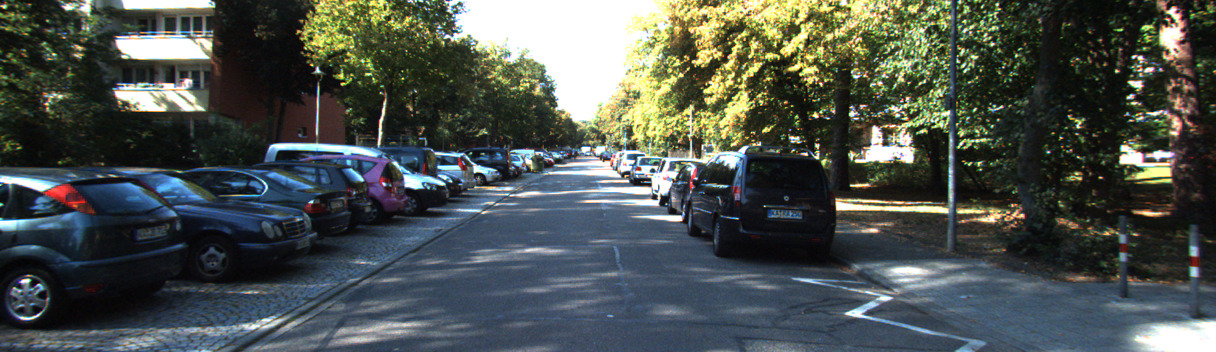}
		\includegraphics[width=1\textwidth]{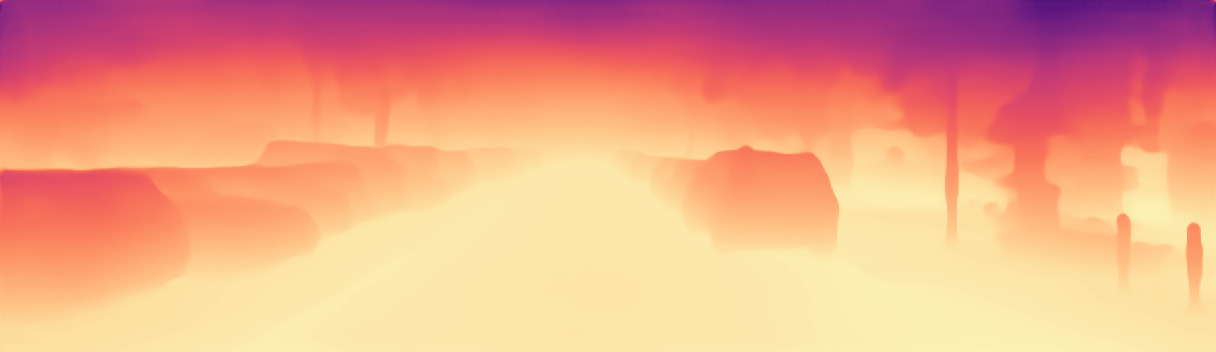}
	\end{minipage}
	\begin{minipage}{0.24\linewidth}
		\includegraphics[width=1\textwidth]{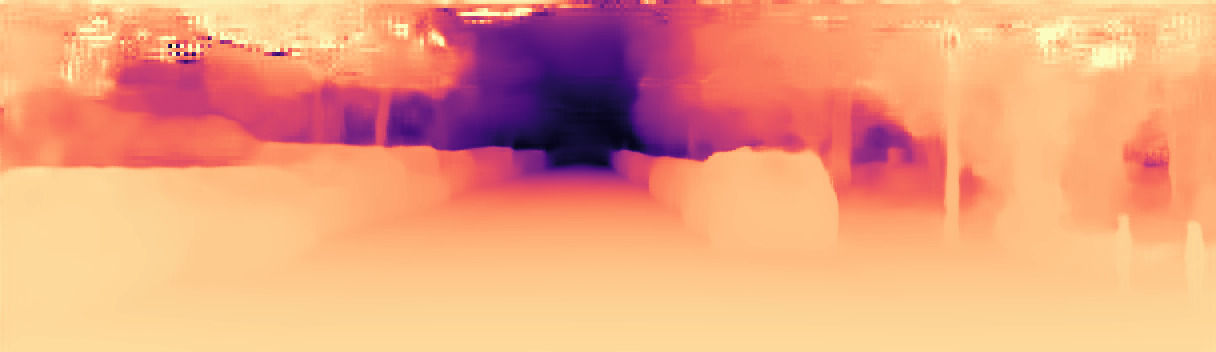}
		\includegraphics[width=1\textwidth]{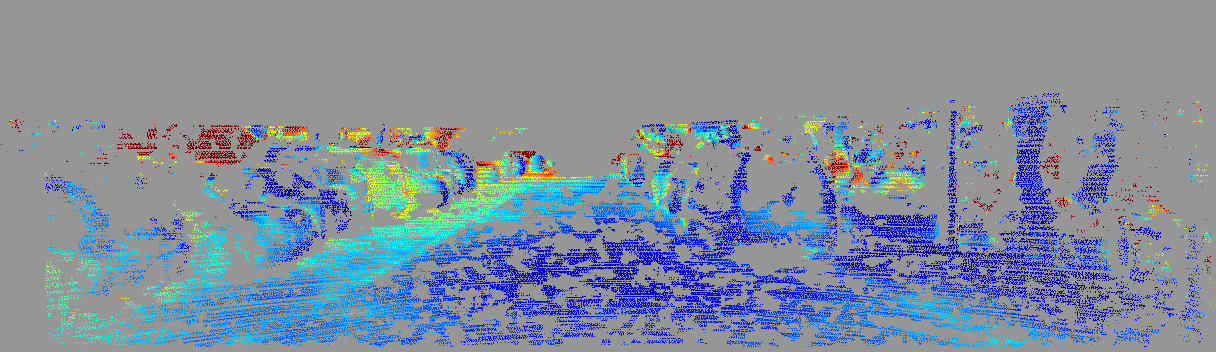}
	\end{minipage}
	\begin{minipage}{0.24\linewidth}
		\includegraphics[width=1\textwidth]{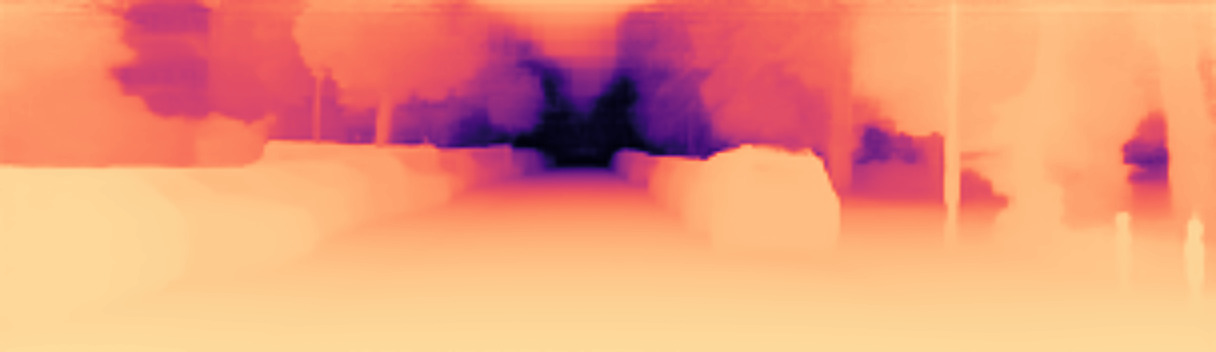}
		\includegraphics[width=1\textwidth]{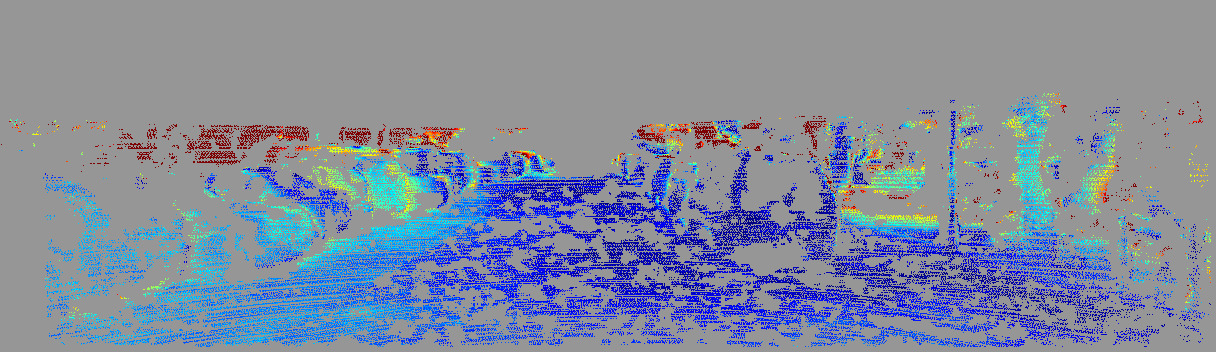}
	\end{minipage}
	\begin{minipage}{0.24\linewidth}
		\includegraphics[width=1\textwidth]{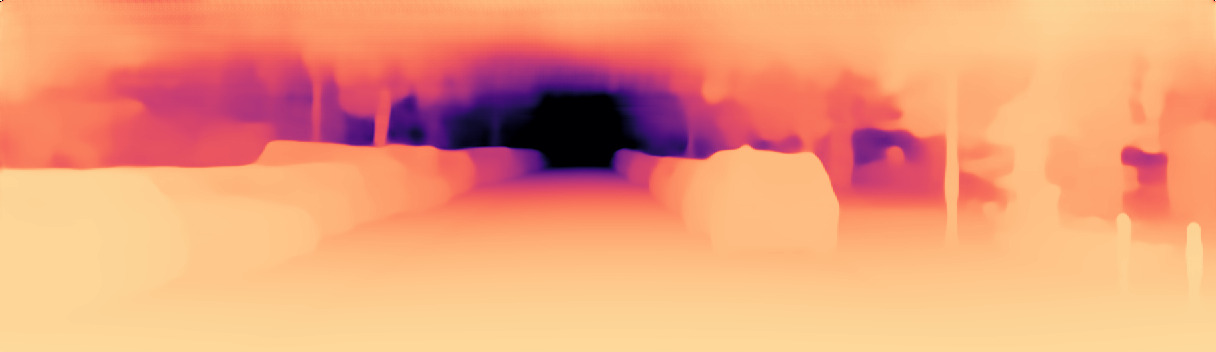}
		\includegraphics[width=1\textwidth]{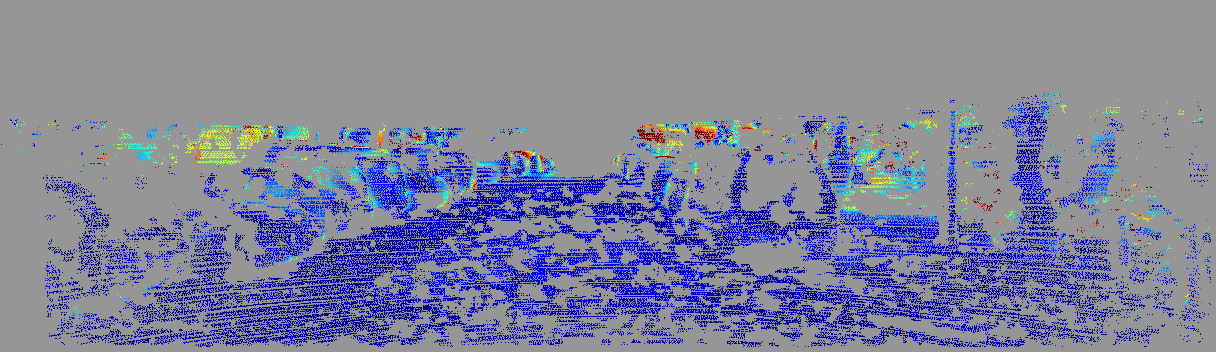}
	\end{minipage}
	
	\begin{minipage}{0.24\linewidth}
		\includegraphics[width=1\textwidth]{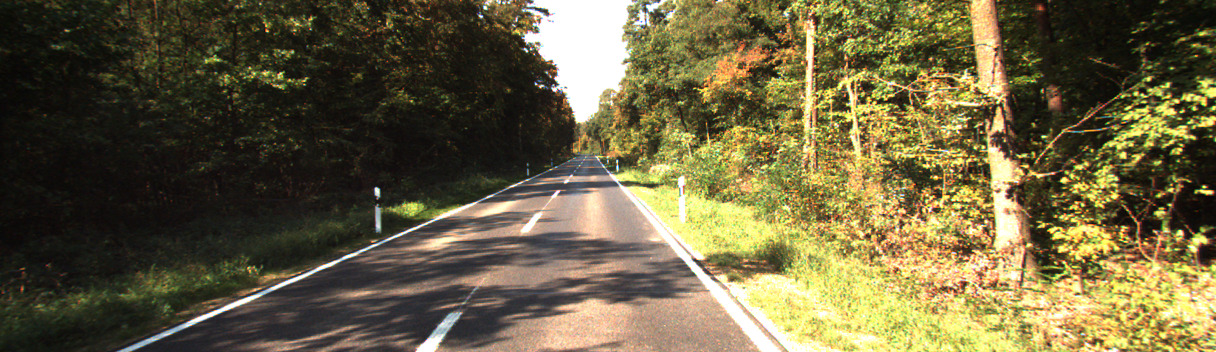}
		\includegraphics[width=1\textwidth]{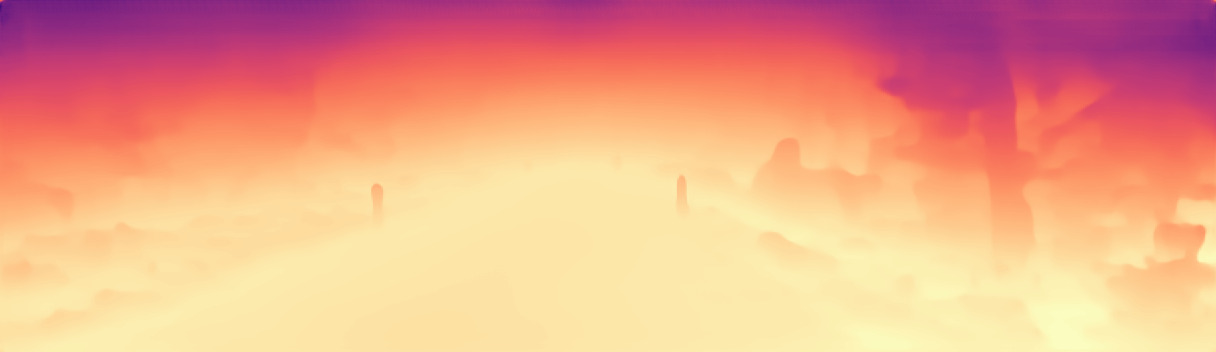}
	\end{minipage}
	\begin{minipage}{0.24\linewidth}
		\includegraphics[width=1\textwidth]{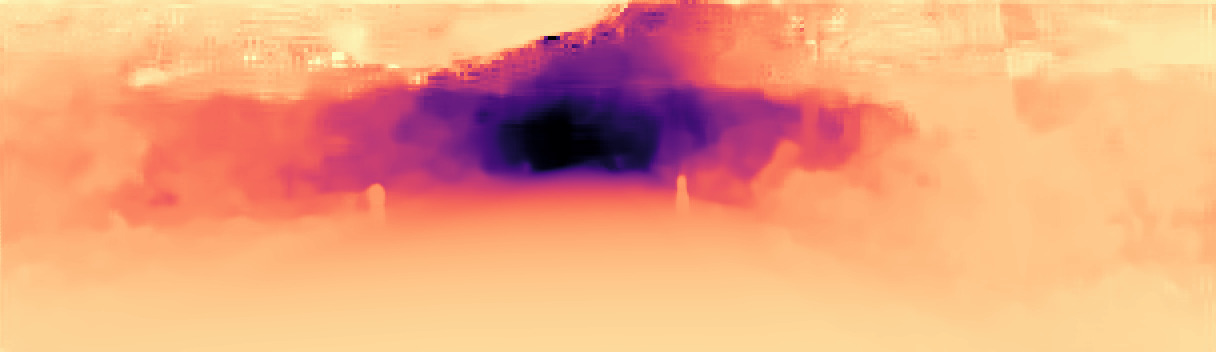}
		\includegraphics[width=1\textwidth]{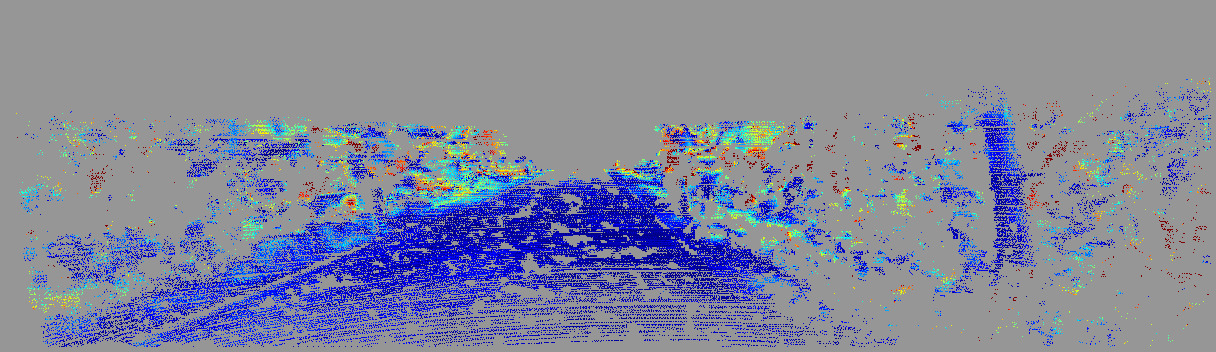}
	\end{minipage}
	\begin{minipage}{0.24\linewidth}
		\includegraphics[width=1\textwidth]{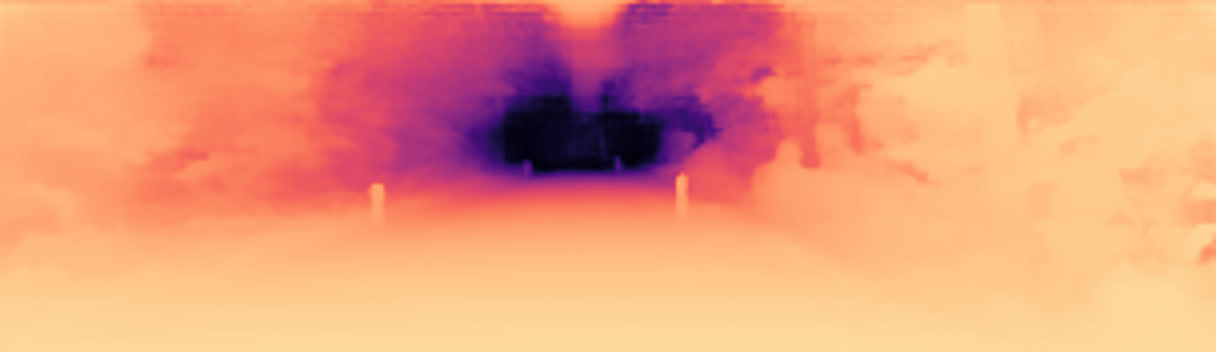}
		\includegraphics[width=1\textwidth]{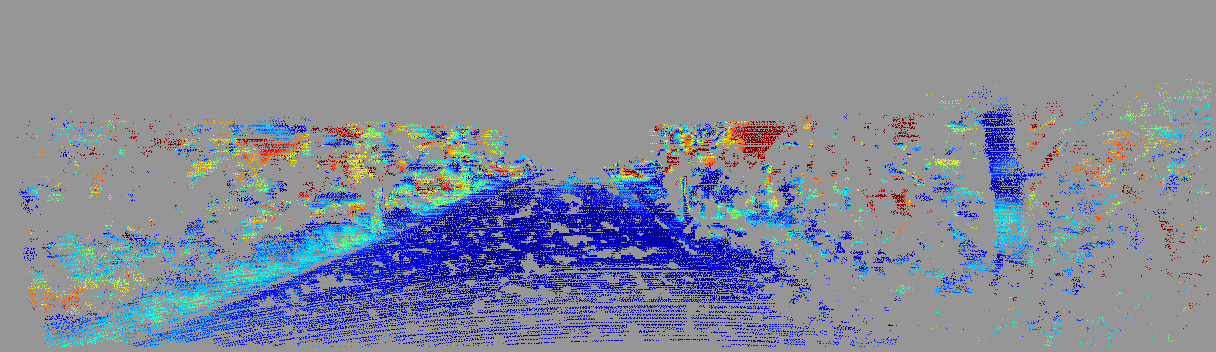}
	\end{minipage}
	\begin{minipage}{0.24\linewidth}
		\includegraphics[width=1\textwidth]{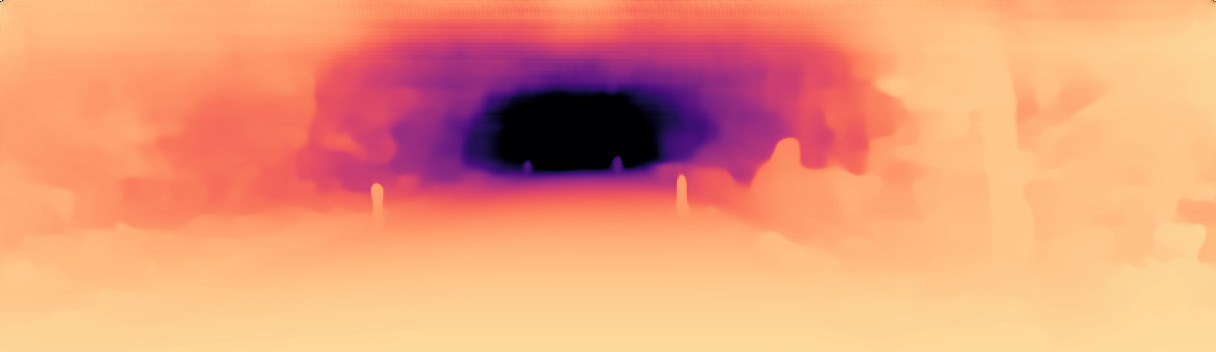}
		\includegraphics[width=1\textwidth]{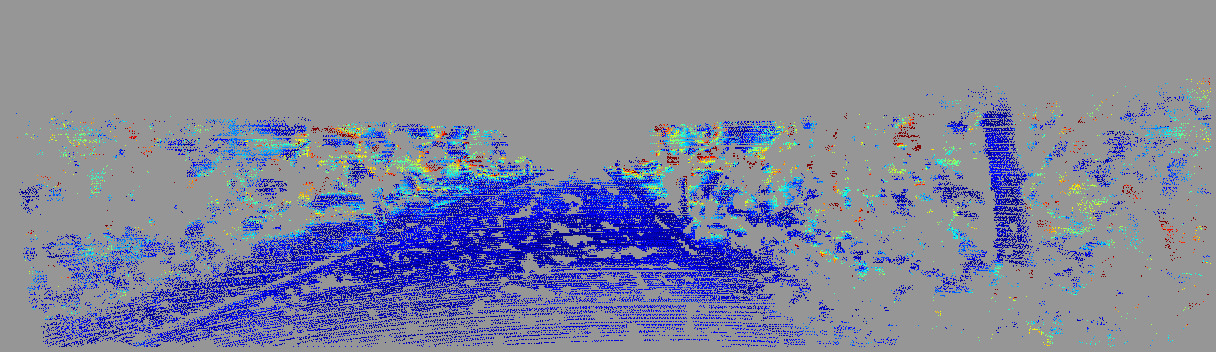}
	\end{minipage}
	
	\begin{minipage}{0.24\linewidth}
		\includegraphics[width=1\textwidth]{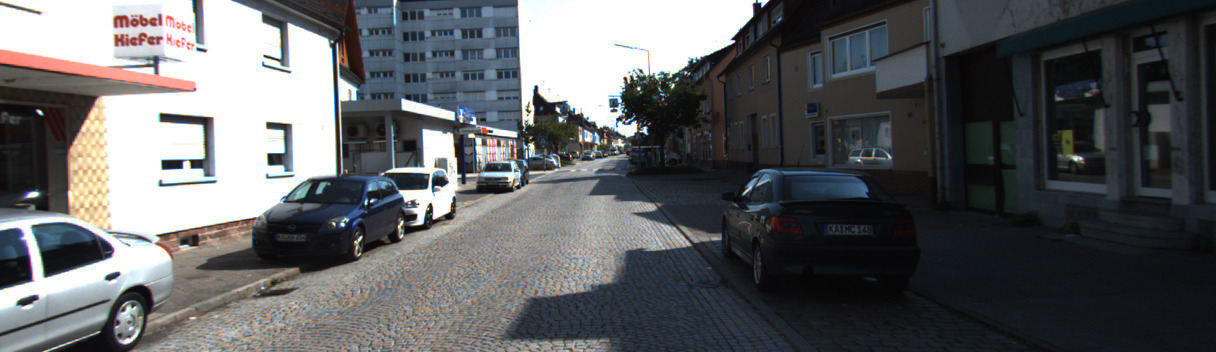}
		\includegraphics[width=1\textwidth]{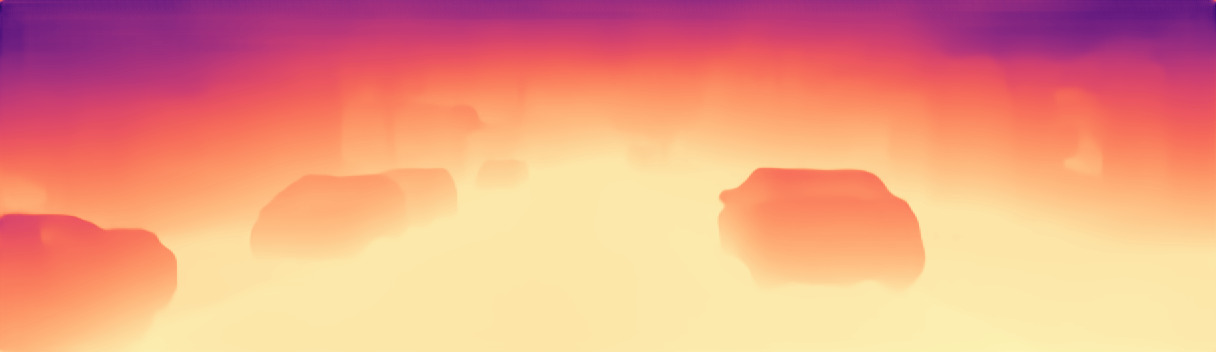}
	\end{minipage}
	\begin{minipage}{0.24\linewidth}
		\includegraphics[width=1\textwidth]{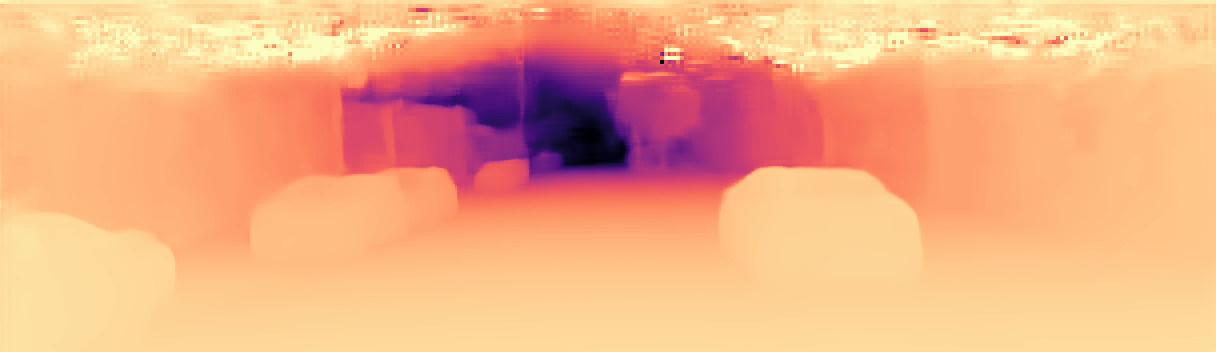}
		\includegraphics[width=1\textwidth]{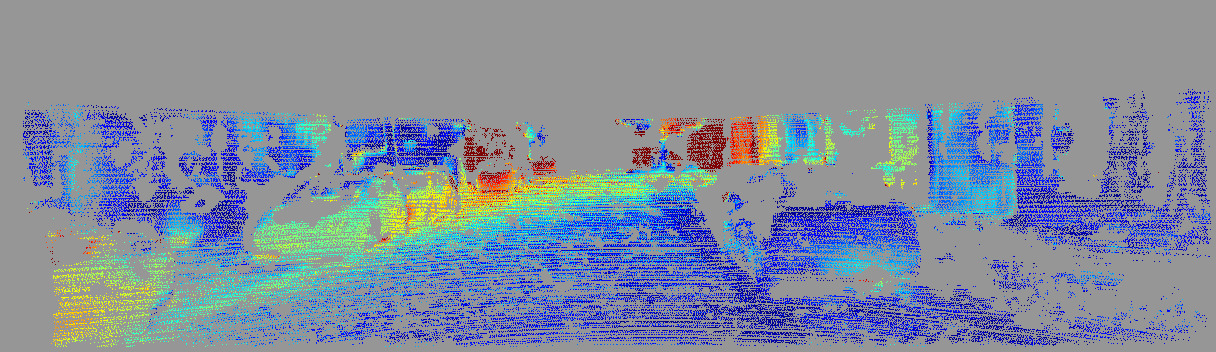}
	\end{minipage}
	\begin{minipage}{0.24\linewidth}
		\includegraphics[width=1\textwidth]{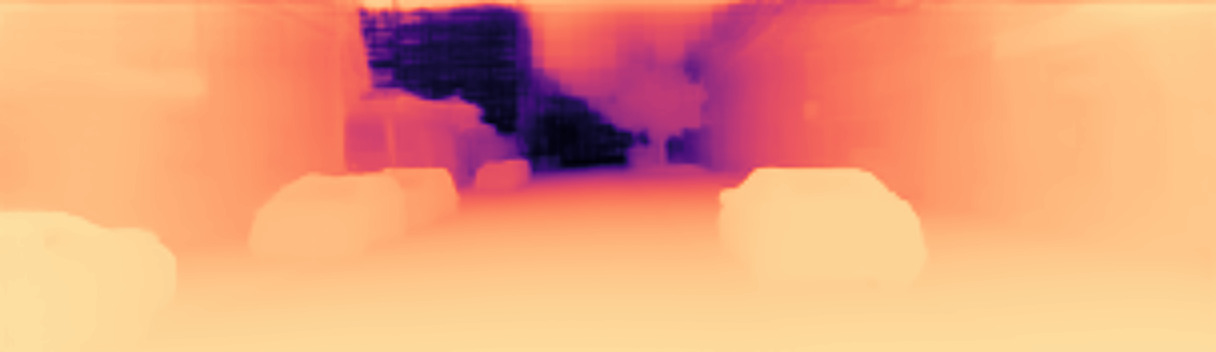}
		\includegraphics[width=1\textwidth]{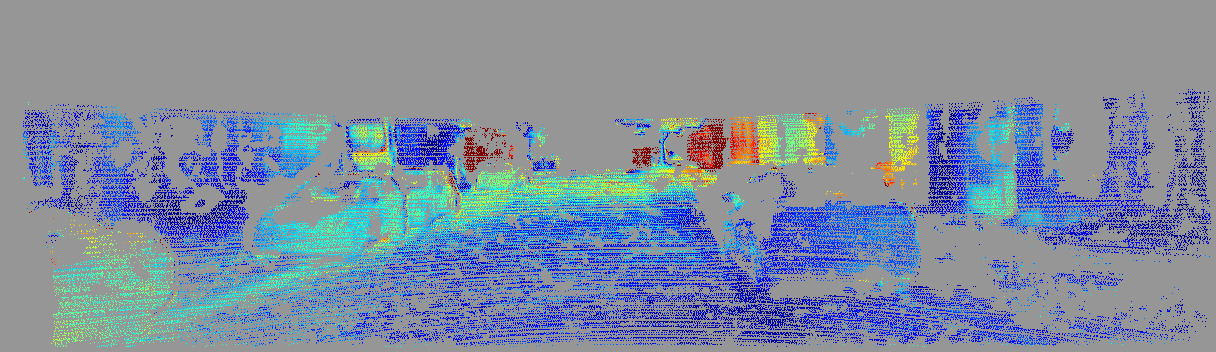}
	\end{minipage}
	\begin{minipage}{0.24\linewidth}
		\includegraphics[width=1\textwidth]{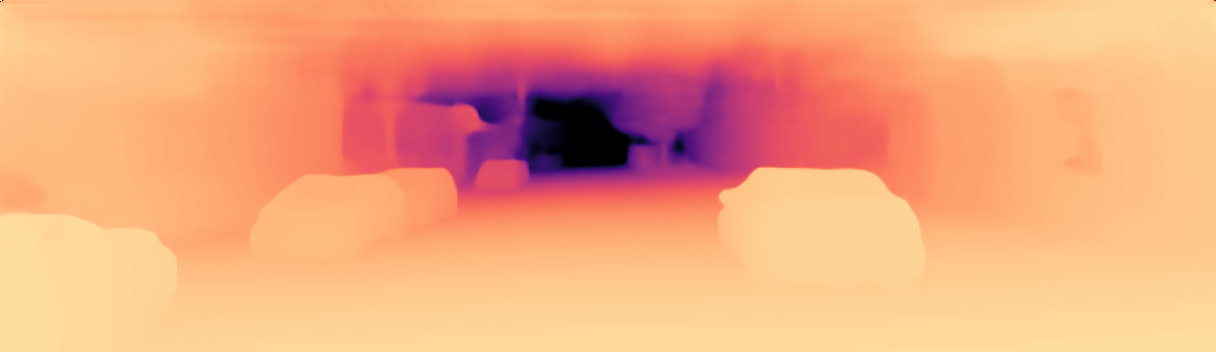}
		\includegraphics[width=1\textwidth]{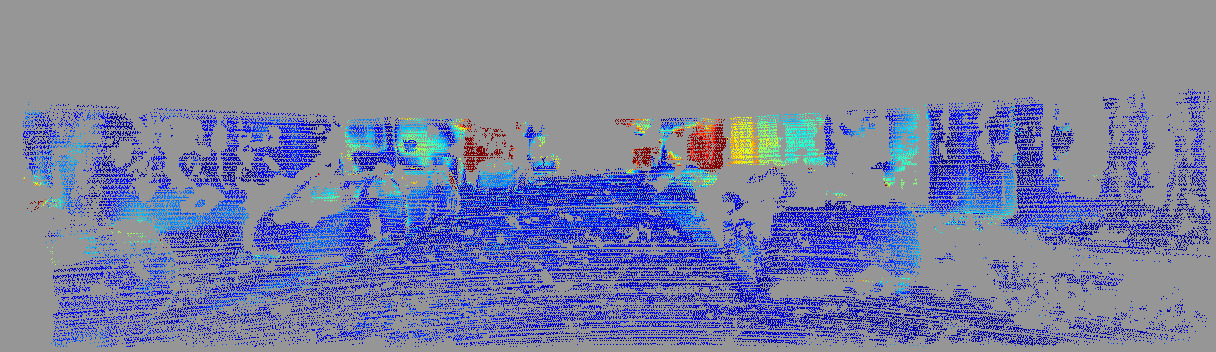}
	\end{minipage}
	
	\begin{minipage}{0.24\linewidth}
		\includegraphics[width=1\textwidth]{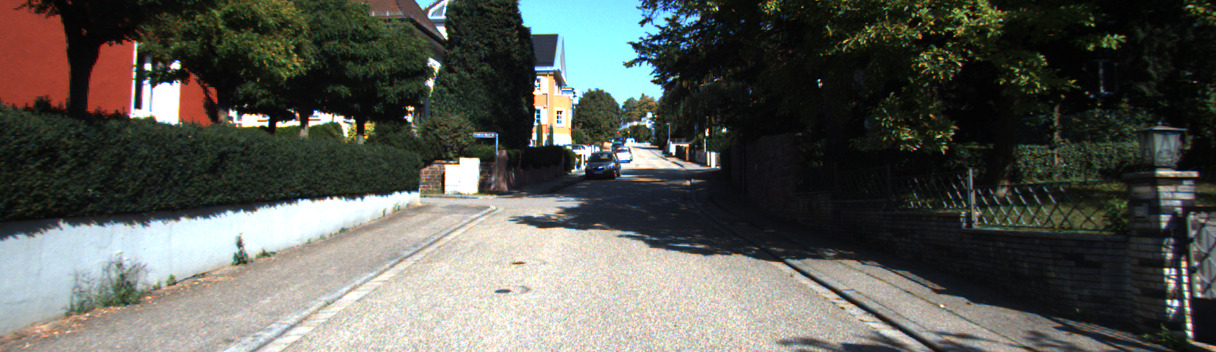}
		\includegraphics[width=1\textwidth]{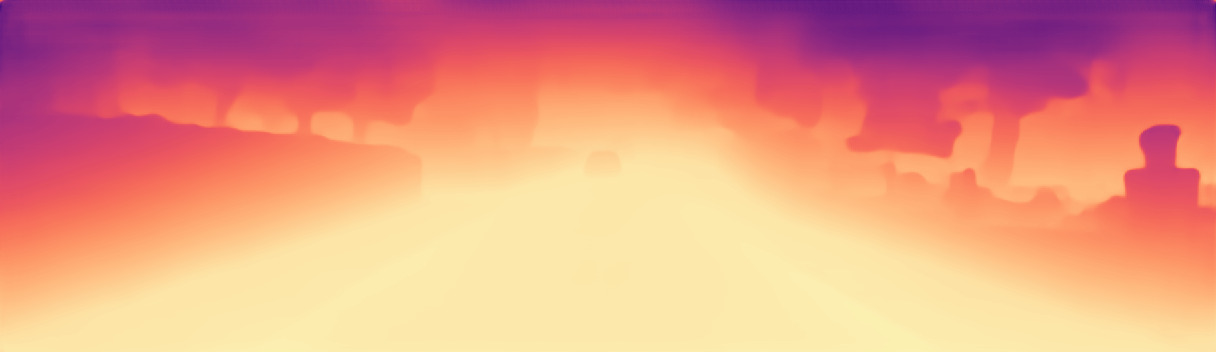}
	\end{minipage}
	\begin{minipage}{0.24\linewidth}
		\includegraphics[width=1\textwidth]{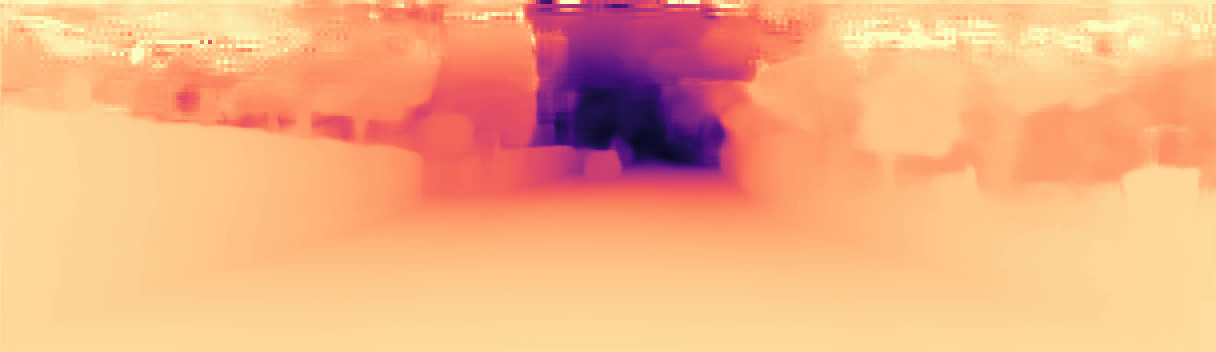}
		\includegraphics[width=1\textwidth]{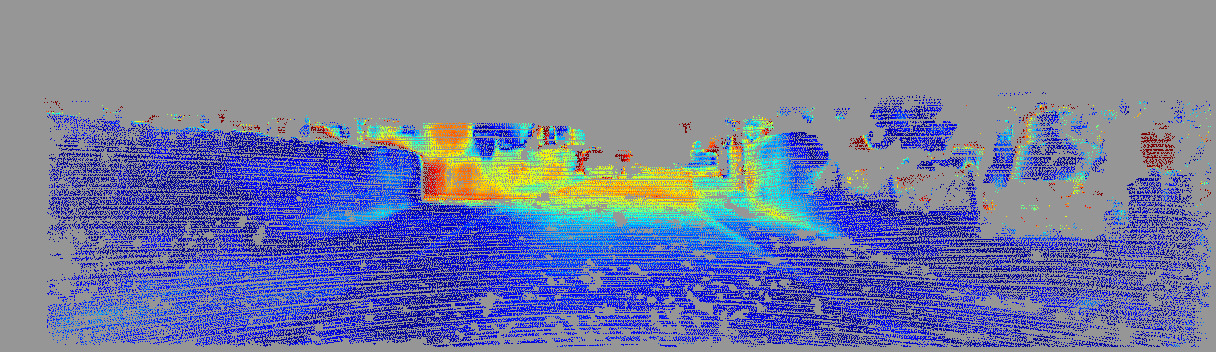}
	\end{minipage}
	\begin{minipage}{0.24\linewidth}
		\includegraphics[width=1\textwidth]{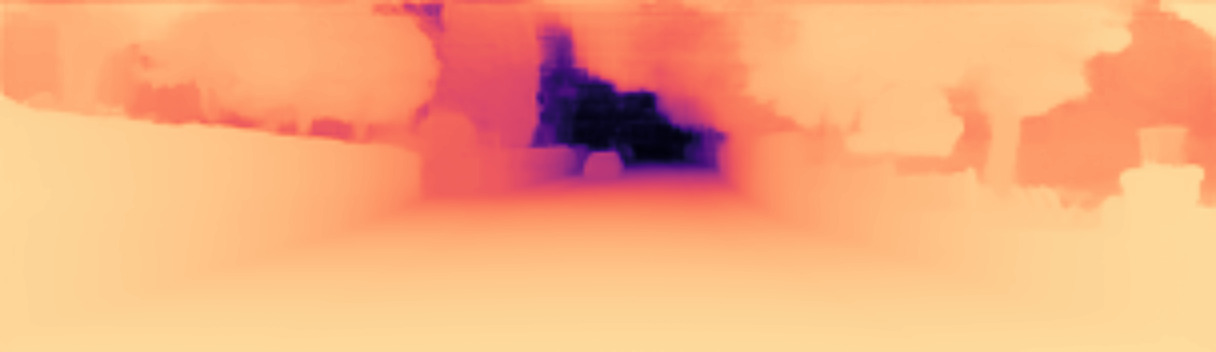}
		\includegraphics[width=1\textwidth]{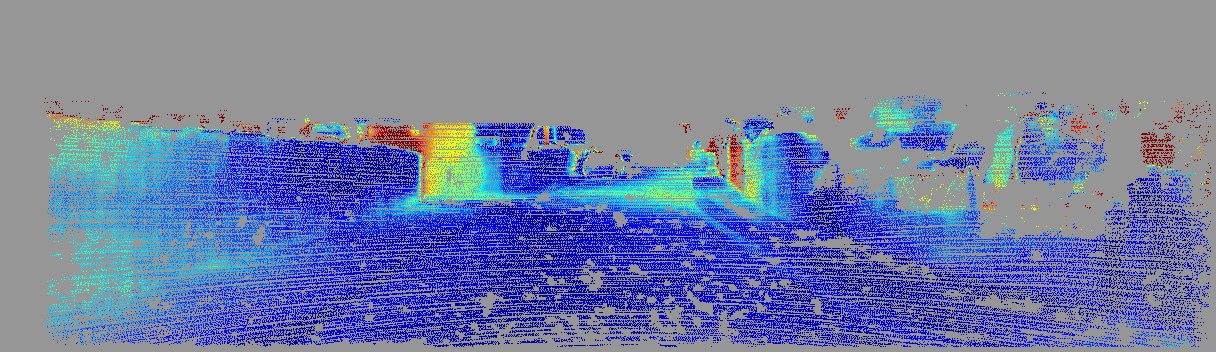}
	\end{minipage}
	\begin{minipage}{0.24\linewidth}
		\includegraphics[width=1\textwidth]{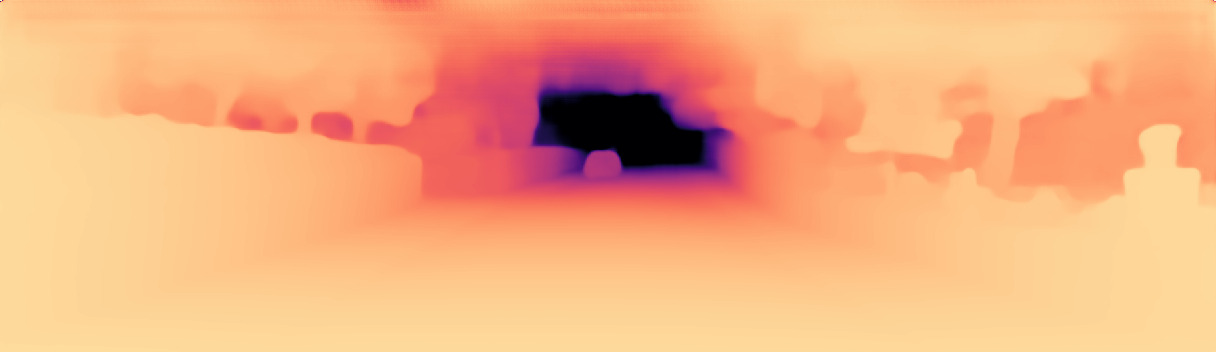}
		\includegraphics[width=1\textwidth]{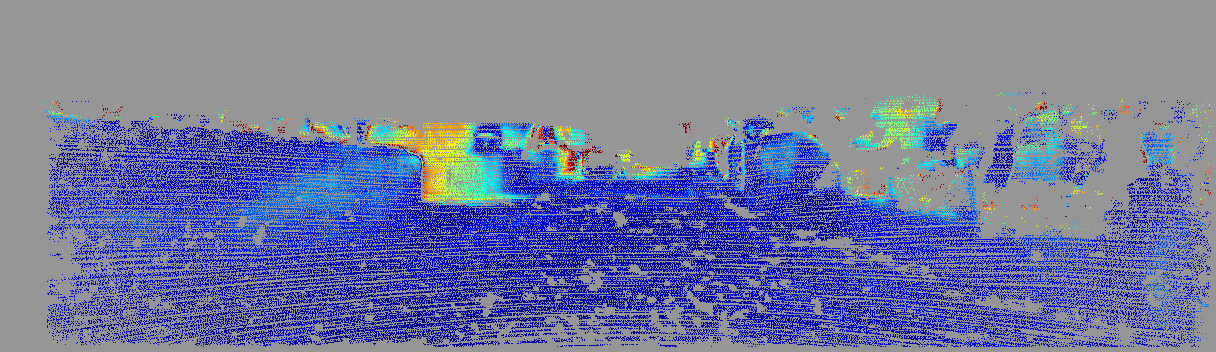}
	\end{minipage}
	
	\begin{minipage}{0.24\linewidth}
		\includegraphics[width=1\textwidth]{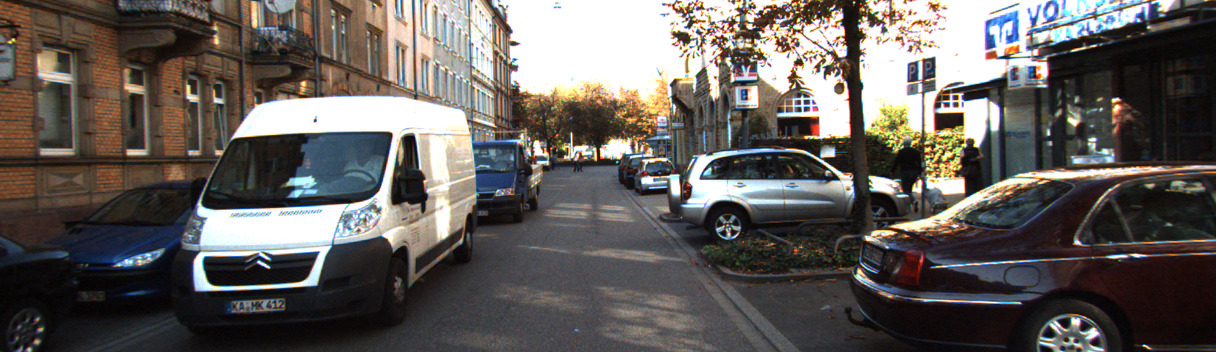}
		\includegraphics[width=1\textwidth]{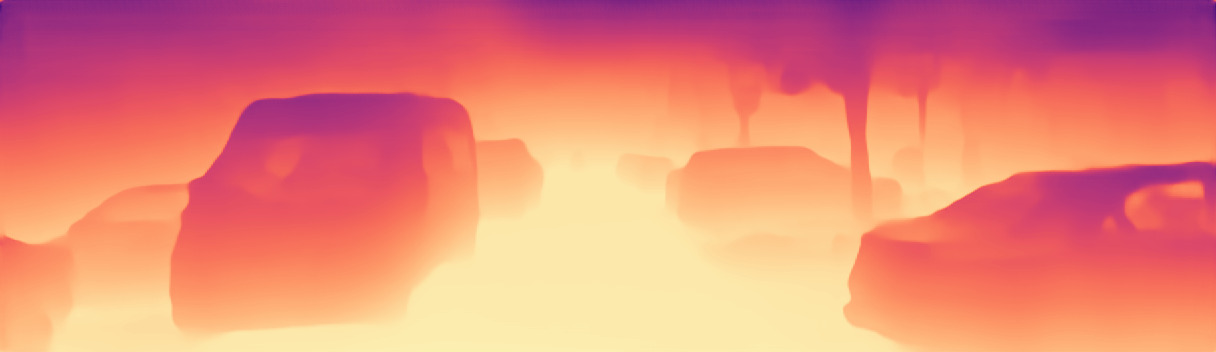}
	\end{minipage}
	\begin{minipage}{0.24\linewidth}
		\includegraphics[width=1\textwidth]{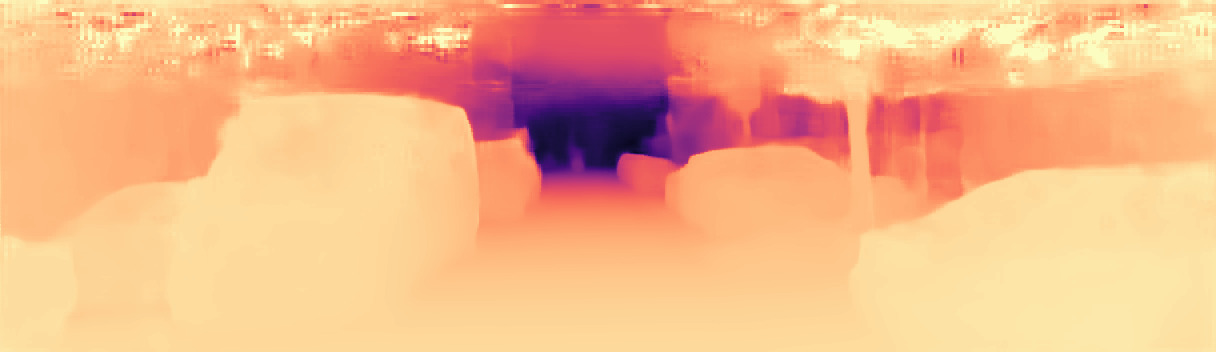}
		\includegraphics[width=1\textwidth]{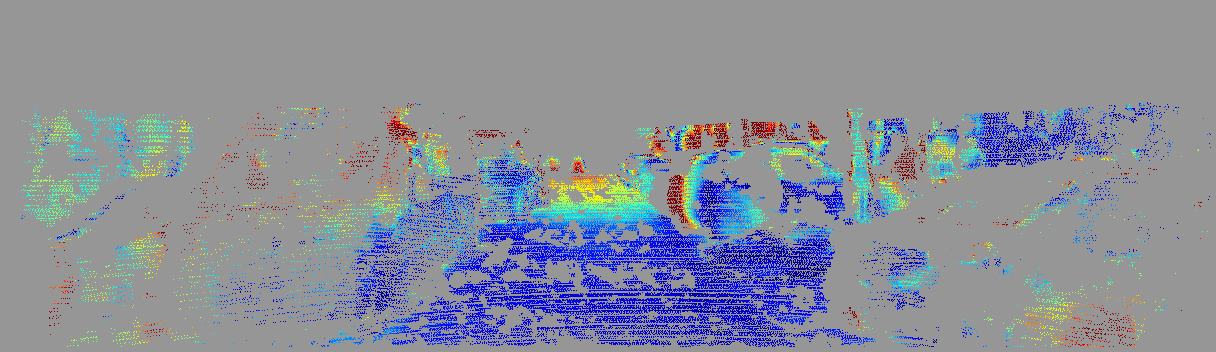}
	\end{minipage}
	\begin{minipage}{0.24\linewidth}
		\includegraphics[width=1\textwidth]{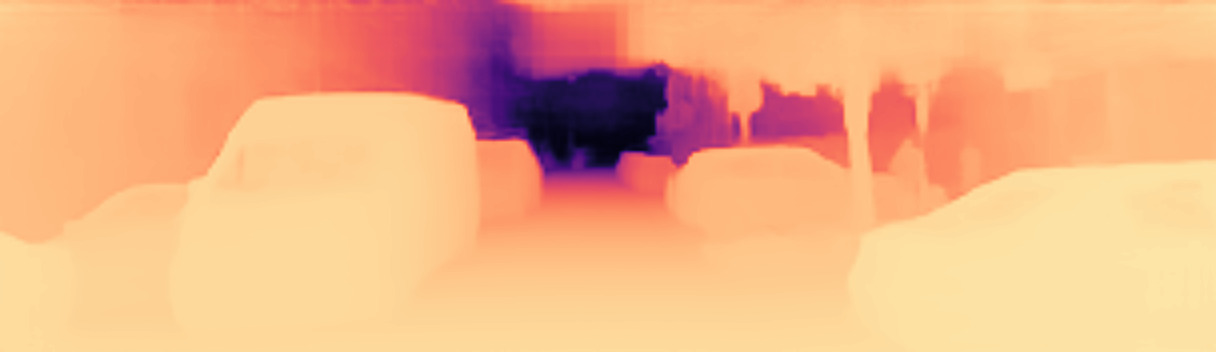}
		\includegraphics[width=1\textwidth]{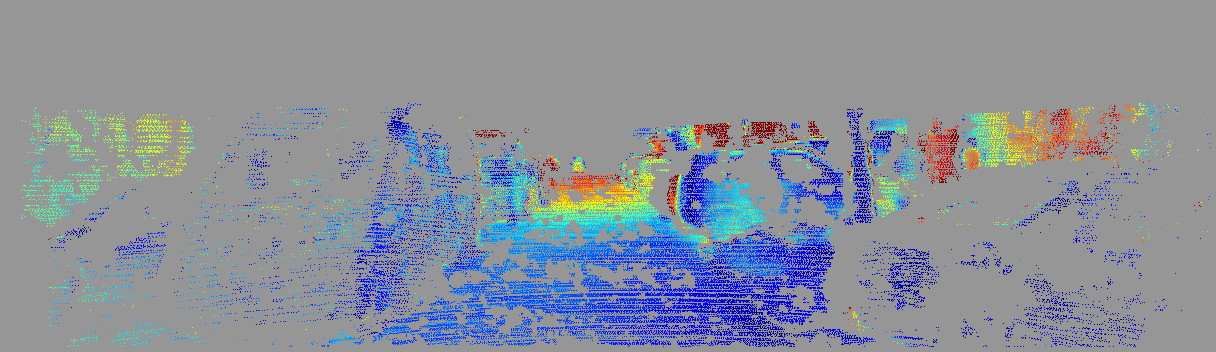}
	\end{minipage}
	\begin{minipage}{0.24\linewidth}
		\includegraphics[width=1\textwidth]{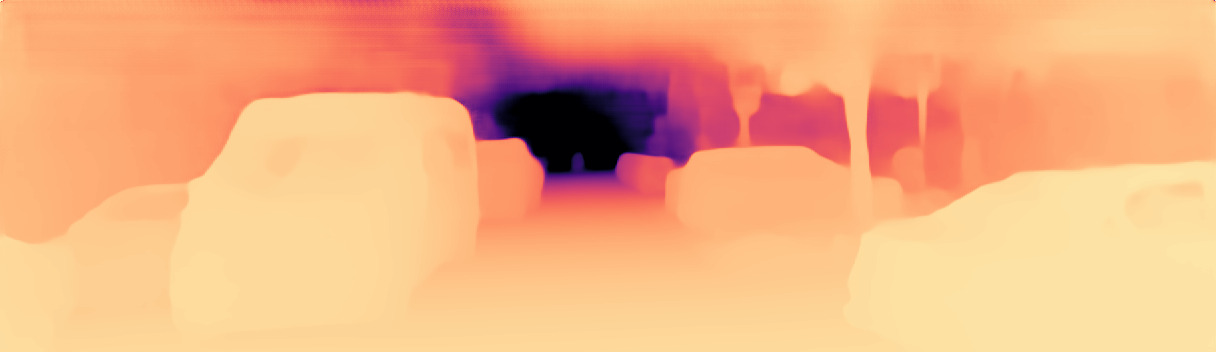}
		\includegraphics[width=1\textwidth]{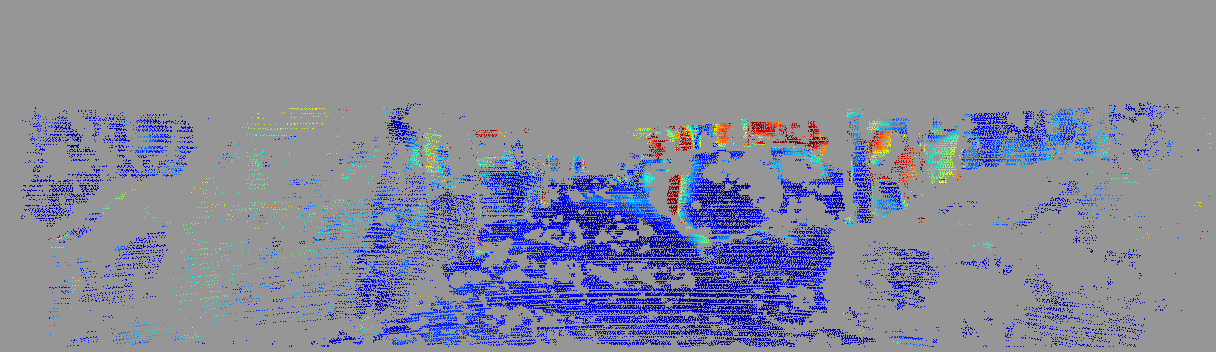}
	\end{minipage}
	
	\begin{minipage}{0.24\linewidth}
		\includegraphics[width=1\textwidth]{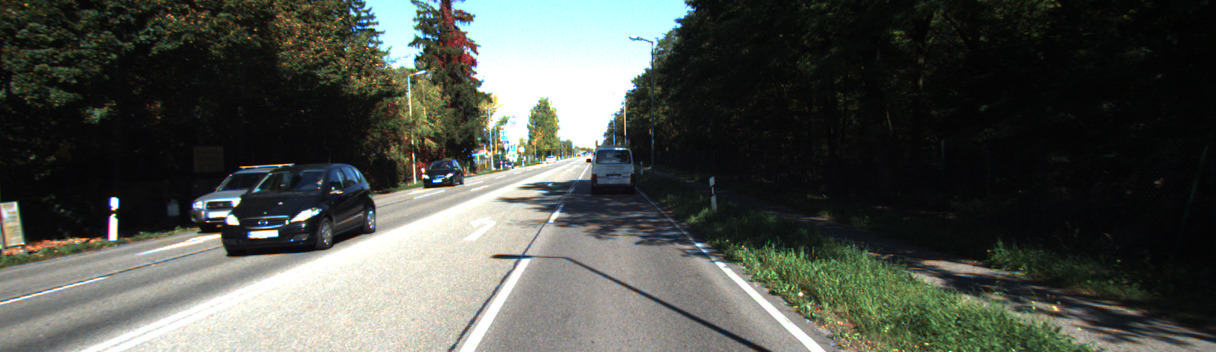}
		\includegraphics[width=1\textwidth]{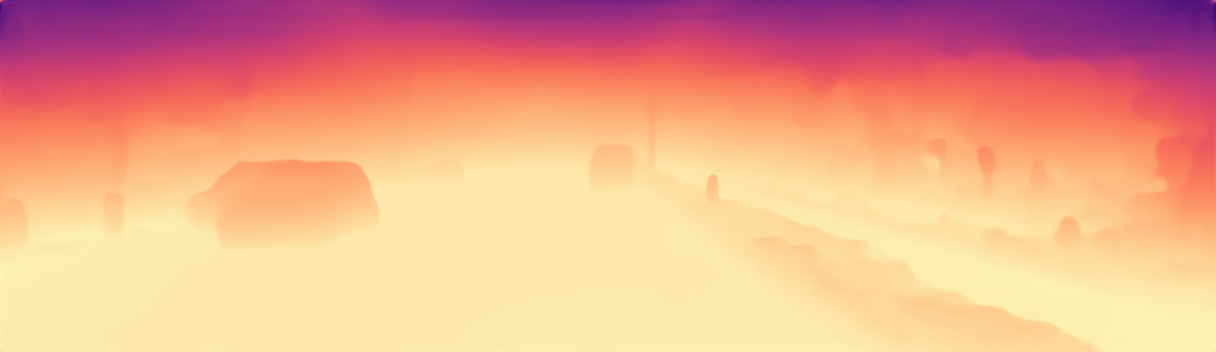}
	\end{minipage}
	\begin{minipage}{0.24\linewidth}
		\includegraphics[width=1\textwidth]{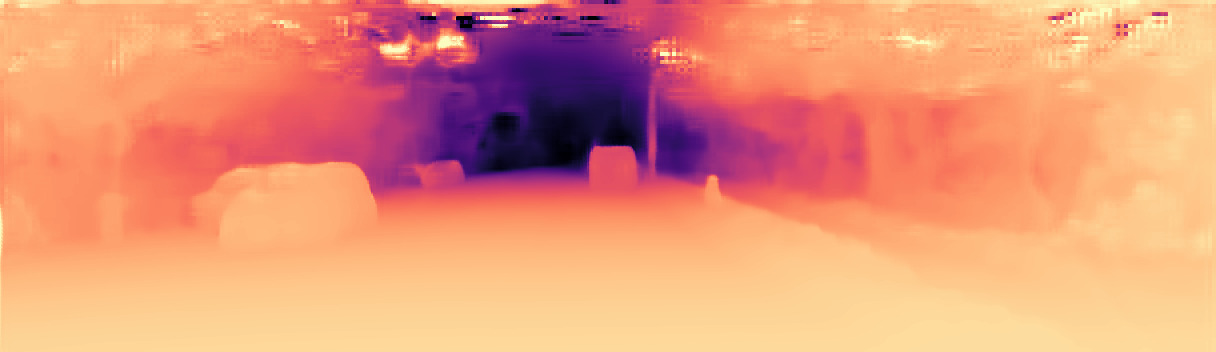}
		\includegraphics[width=1\textwidth]{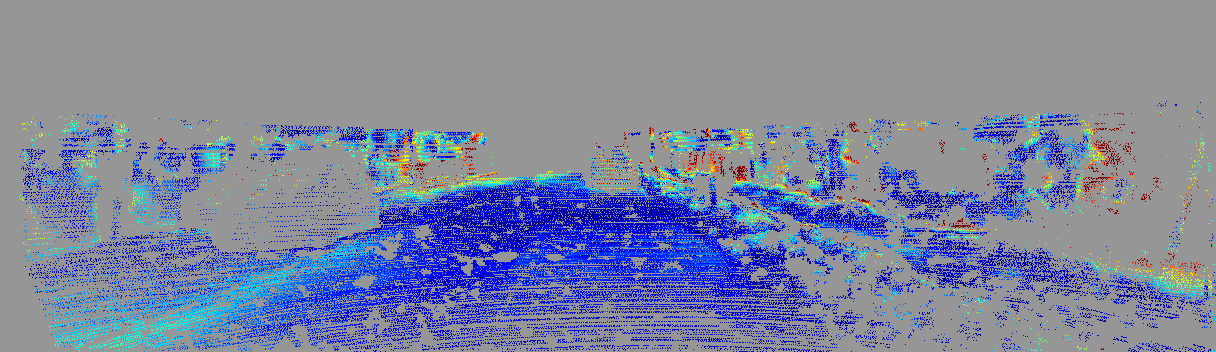}
	\end{minipage}
	\begin{minipage}{0.24\linewidth}
		\includegraphics[width=1\textwidth]{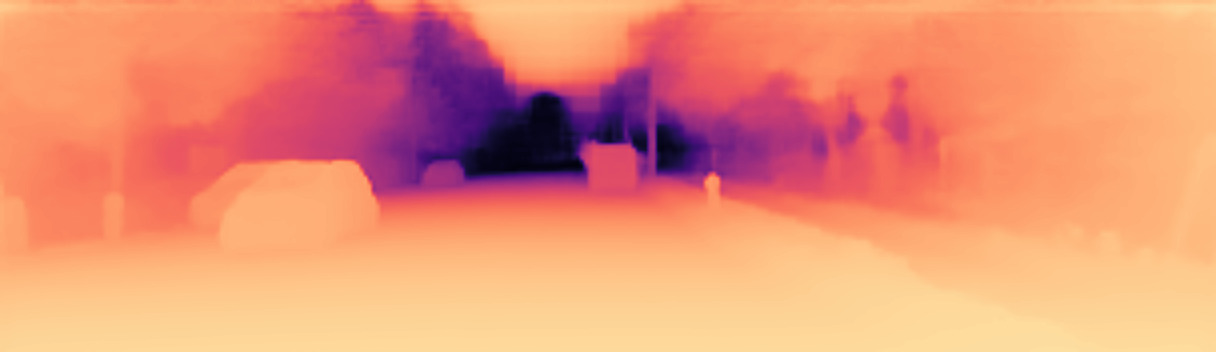}
		\includegraphics[width=1\textwidth]{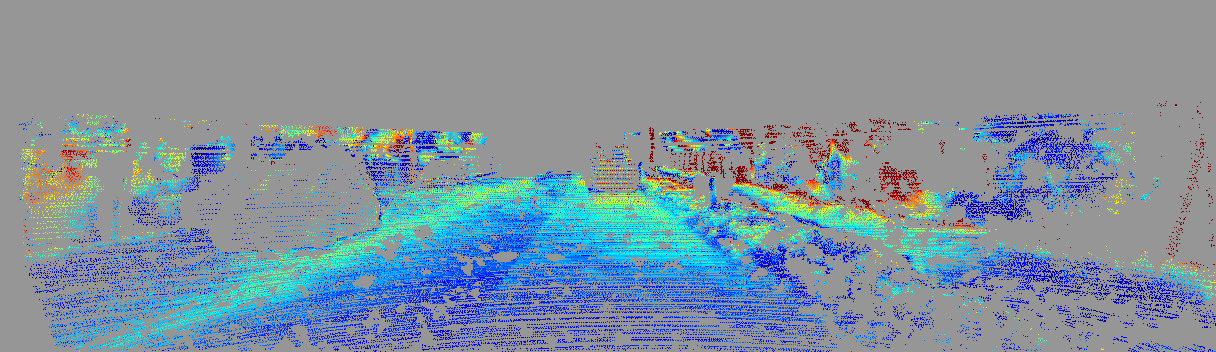}
	\end{minipage}
	\begin{minipage}{0.24\linewidth}
		\includegraphics[width=1\textwidth]{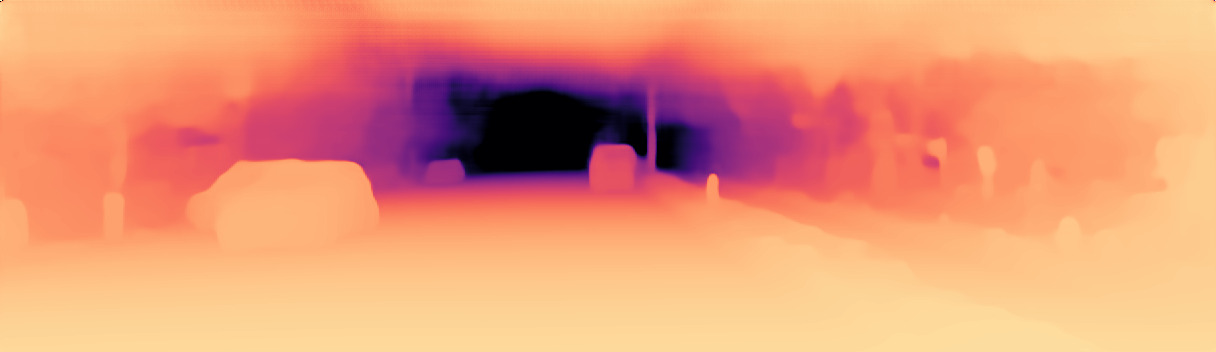}
		\includegraphics[width=1\textwidth]{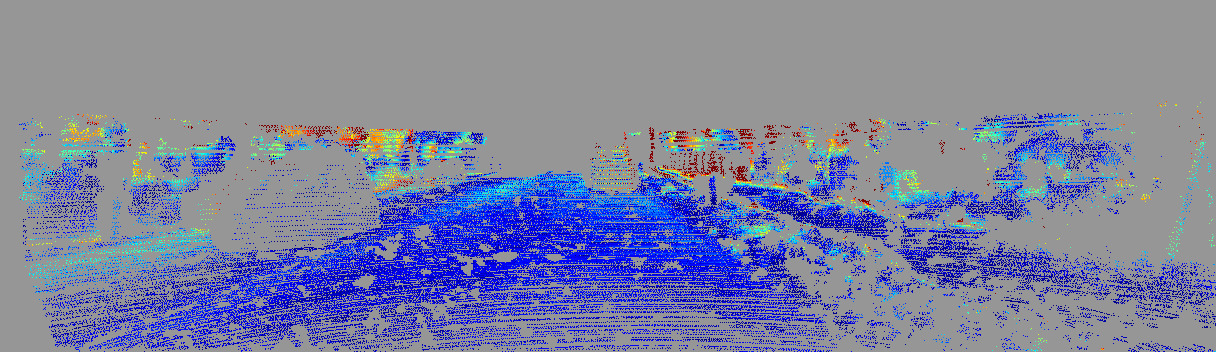}
	\end{minipage}

	\begin{minipage}{0.24\linewidth}
		\centering
		Input Image
	\end{minipage}
	\begin{minipage}{0.24\linewidth}
		\centering
        MaGNet~\cite{bae2022multi}
	\end{minipage}
	\begin{minipage}{0.24\linewidth}
		\centering
		NeW CRFs~\cite{yuan2022newcrfs}
	\end{minipage}
	\begin{minipage}{0.24\linewidth}
		\centering
		Ours
	\end{minipage}

	\caption{Qualitative results on the Eigen split of KITTI dataset. For each sample, the first column shows the target image and the predicted $\gamma$ map by our model. The rest columns each shows the predicted depth map and the corresponding error map for a model. Blue represents smaller error, while red represents larger error.}	\label{appendix_fig:more_result_pics}
\end{figure*}

Fig.~\ref{appendix_fig:more_result_pics} shows more qualitative results of MaGNet~\cite{bae2022multi}, NeW CRFs~\cite{yuan2022newcrfs}, and our method. As shown in the error maps, the result have improved significantly.

\begin{figure*}[tb]
    \centering
    \begin{minipage}{0.19\linewidth}
    	\centering
    	Input Image
    \end{minipage}
    \begin{minipage}{0.39\linewidth}
    	\centering
    	\includegraphics[width=1\textwidth]{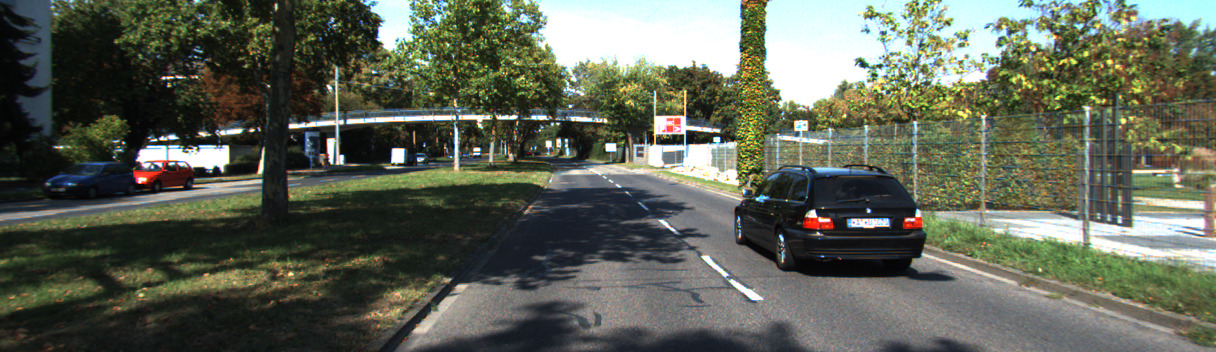}
    \end{minipage}
    \begin{minipage}{0.39\linewidth}
    	\centering
    	\includegraphics[width=1\textwidth]{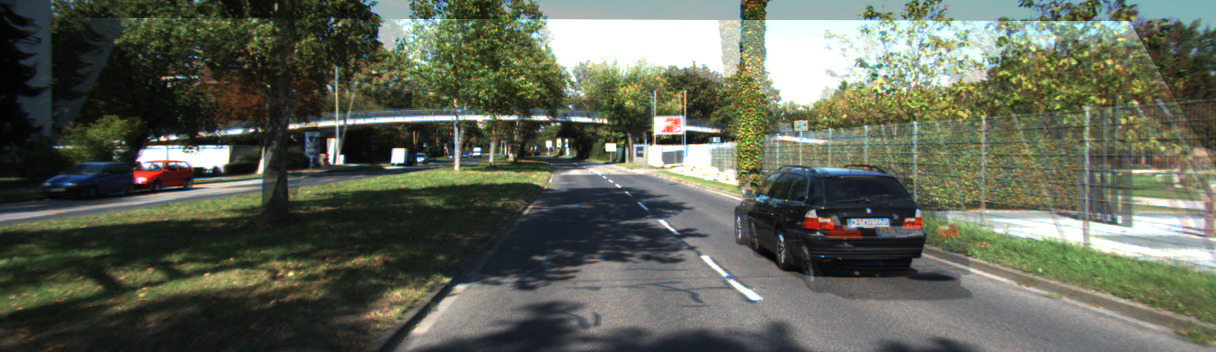}
    \end{minipage}
    
    \begin{minipage}{0.19\linewidth}
    	\centering
    	Baseline
    \end{minipage}
    \begin{minipage}{0.39\linewidth}
    	\centering
    	\includegraphics[width=1\textwidth]{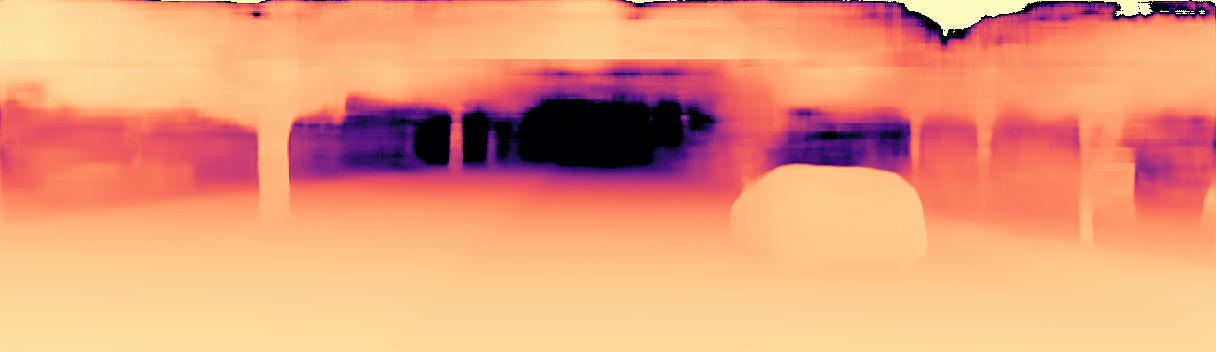}
    \end{minipage}
    \begin{minipage}{0.39\linewidth}
    	\centering
    	\includegraphics[width=1\textwidth]{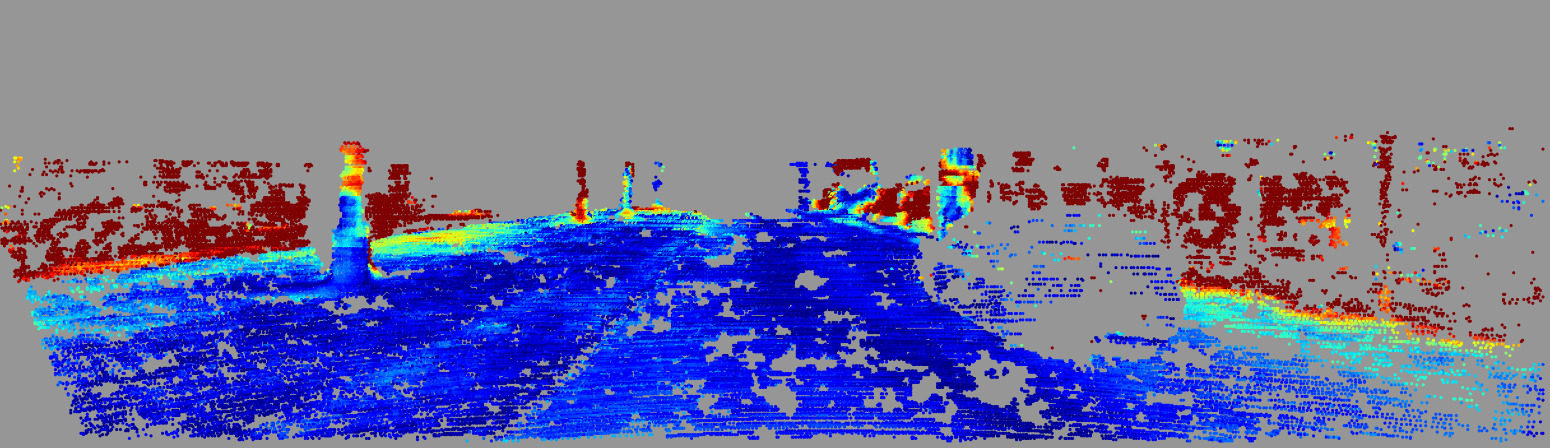}
    \end{minipage}
    
    \begin{minipage}{0.19\linewidth}
    	\centering
    	+FP
    \end{minipage}
    \begin{minipage}{0.39\linewidth}
    	\centering
    	\includegraphics[width=1\textwidth]{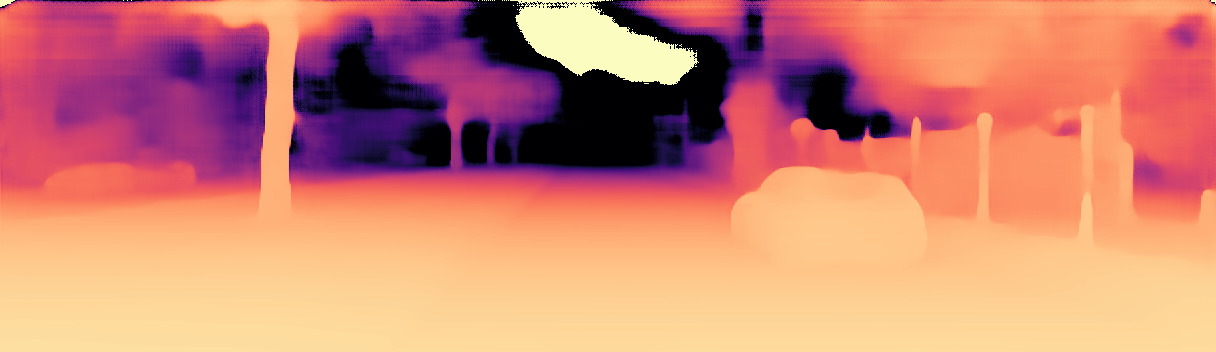}
    \end{minipage}
    \begin{minipage}{0.39\linewidth}
    	\centering
    	\includegraphics[width=1\textwidth]{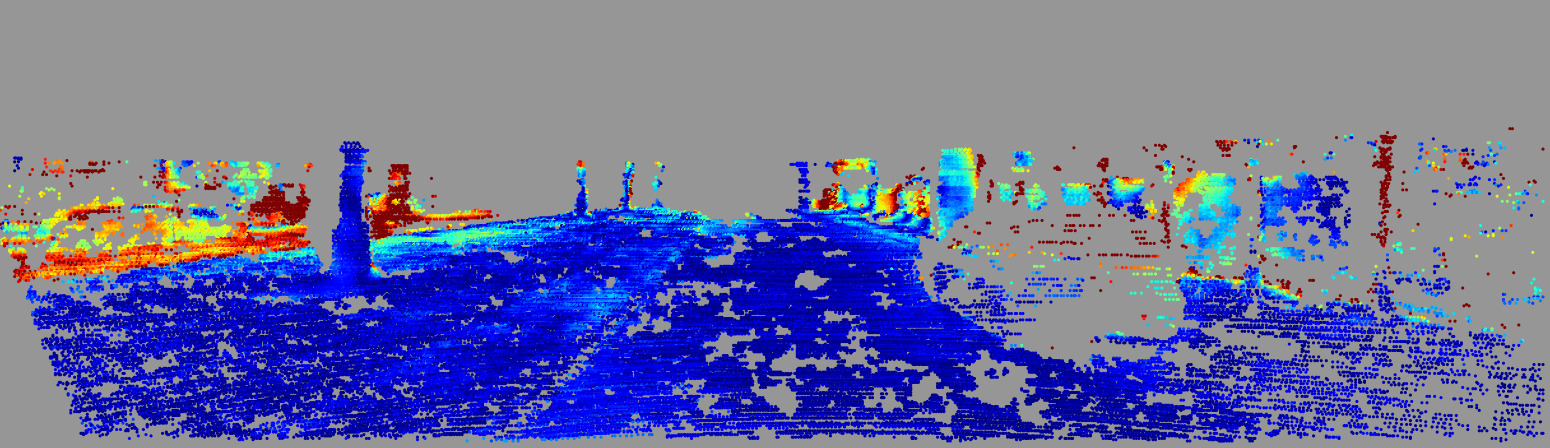}
    \end{minipage}
    
    \begin{minipage}{0.19\linewidth}
    	\centering
    	+PPE
    \end{minipage}
    \begin{minipage}{0.39\linewidth}
    	\centering
    	\includegraphics[width=1\textwidth]{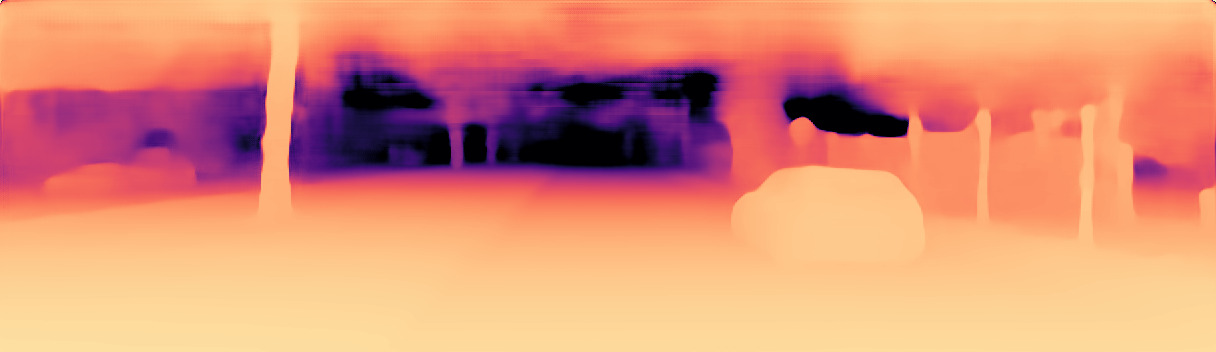}
    \end{minipage}
    \begin{minipage}{0.39\linewidth}
    	\centering
    	\includegraphics[width=1\textwidth]{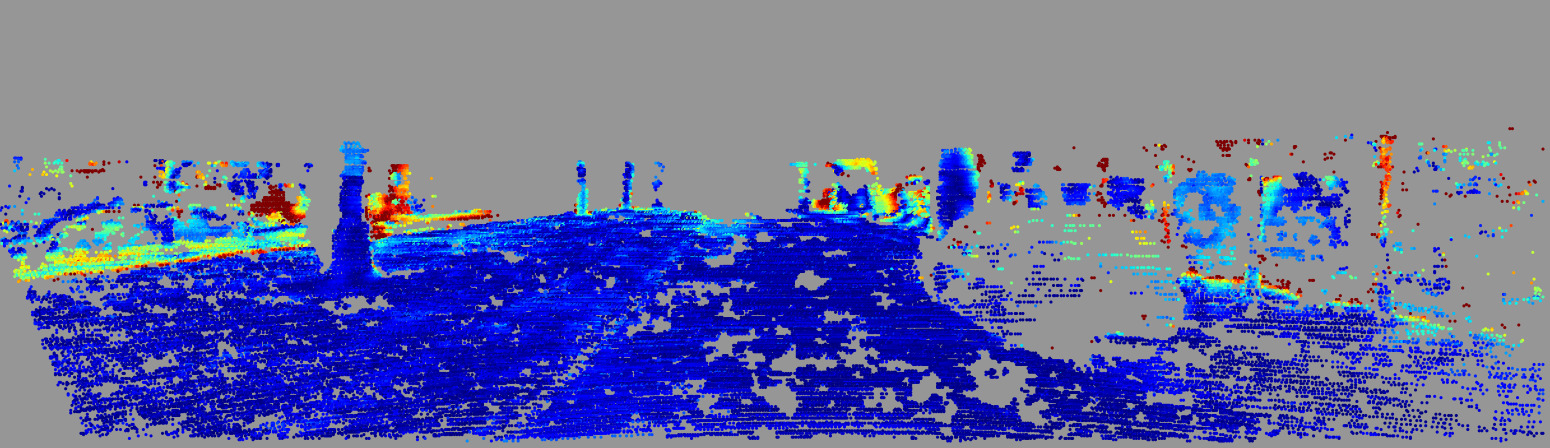}
    \end{minipage}
    
    \begin{minipage}{0.19\linewidth}
    	\centering
    	+SFB
    \end{minipage}
    \begin{minipage}{0.39\linewidth}
    	\centering
    	\includegraphics[width=1\textwidth]{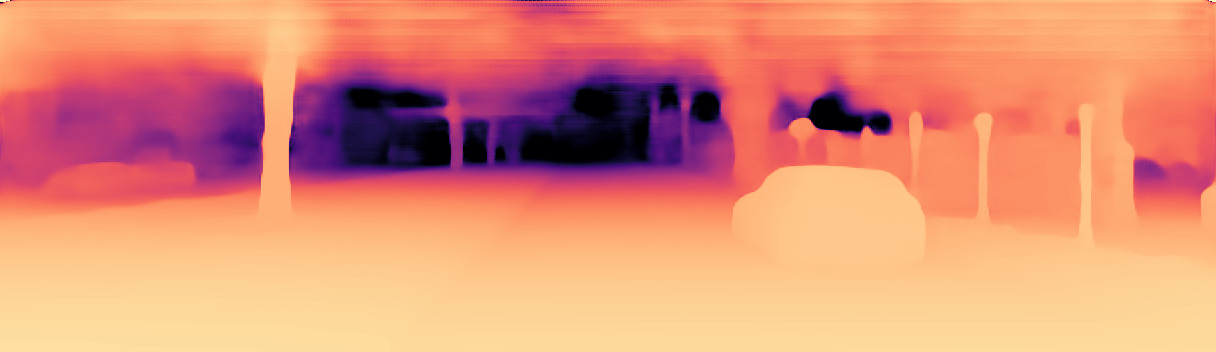}
    \end{minipage}
    \begin{minipage}{0.39\linewidth}
    	\centering
    	\includegraphics[width=1\textwidth]{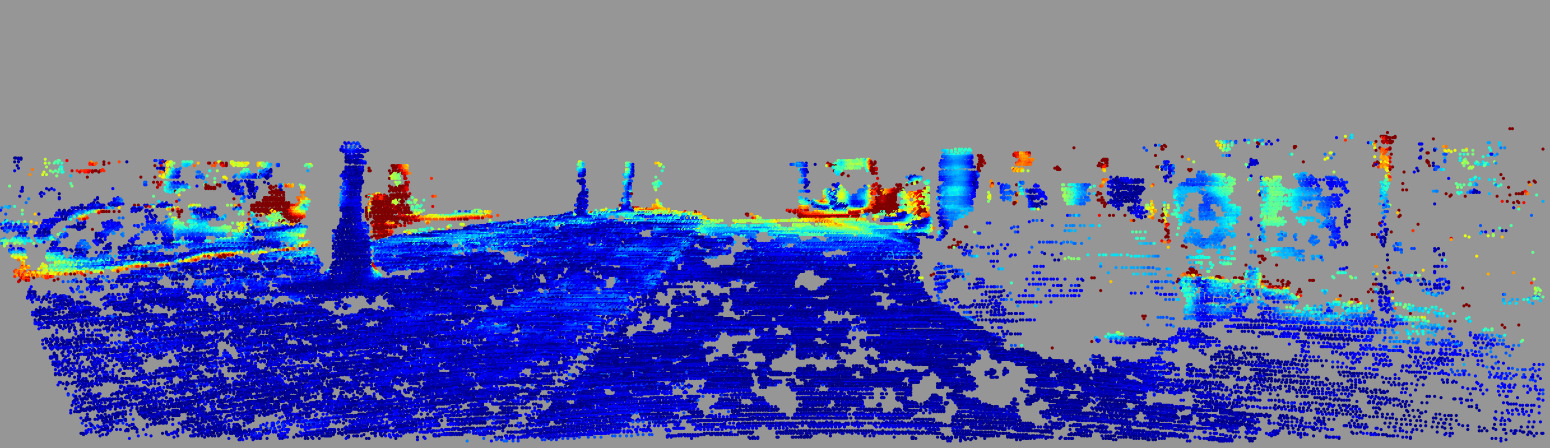}
    \end{minipage}
    
    \begin{minipage}{0.19\linewidth}
    	\centering
    	+DL
    \end{minipage}
    \begin{minipage}{0.39\linewidth}
    	\centering
    	\includegraphics[width=1\textwidth]{imgs_appendix/results/58result.jpg}
    \end{minipage}
    \begin{minipage}{0.39\linewidth}
    	\centering
    	\includegraphics[width=1\textwidth]{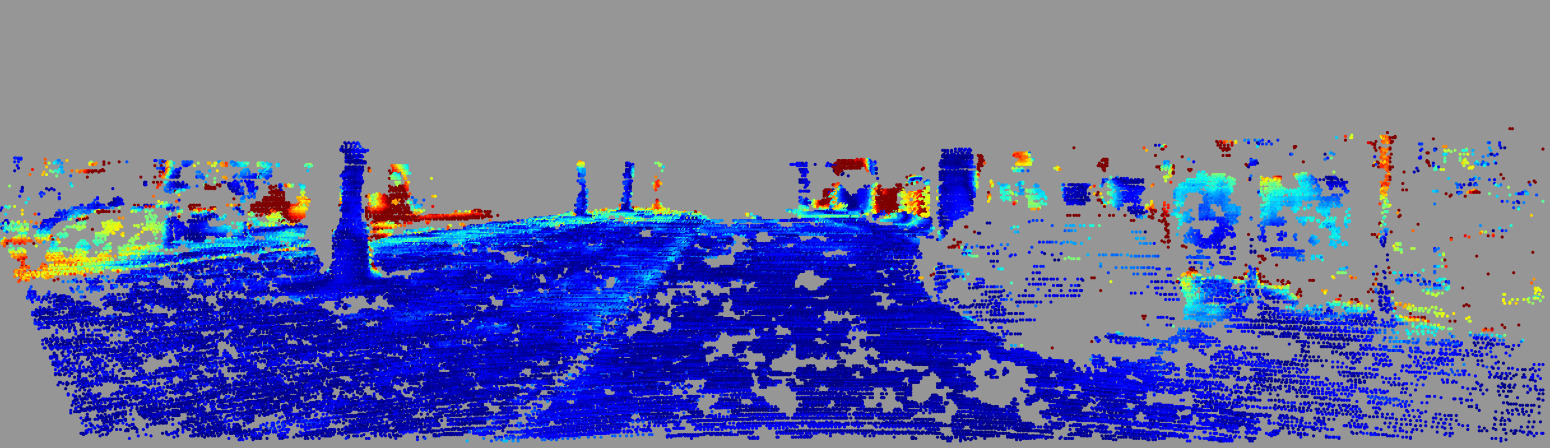}
    \end{minipage}
    
    \caption{Qualitative results for components. The components is added incrementally. Flow pretrain is shown as FP. SFB means single frame branch. DL is the depth loss. PPE is the proposed Planar Position Embedding. The first row shows the target image and the plane-aligned image pairs. The rest rows shows the predicted depth map and the corresponding error map separately.}
    \label{appendix_fig:ablation}
\end{figure*}

\begin{figure*}[tb]
    \centering
    \begin{minipage}{0.19\linewidth}
    	\centering
    	Input Image
    \end{minipage}
    \begin{minipage}{0.39\linewidth}
    	\centering
    	\includegraphics[width=1\textwidth]{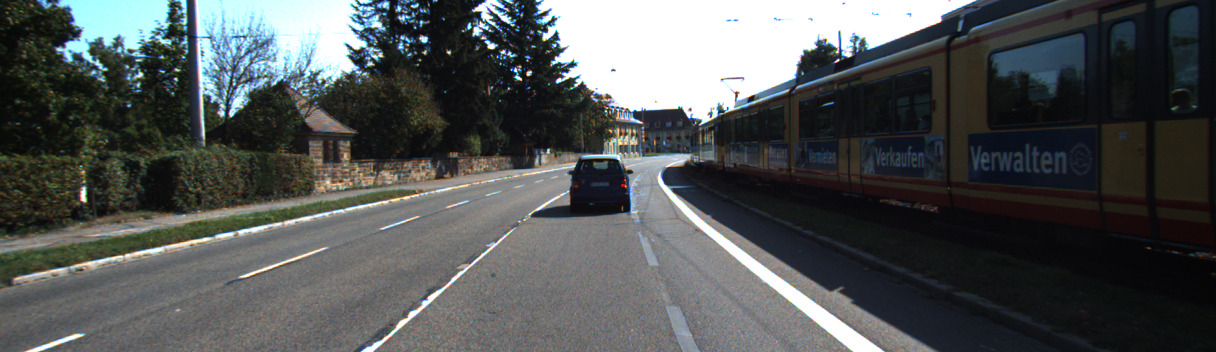}
    \end{minipage}
    \begin{minipage}{0.39\linewidth}
    	\centering
    	\includegraphics[width=1\textwidth]{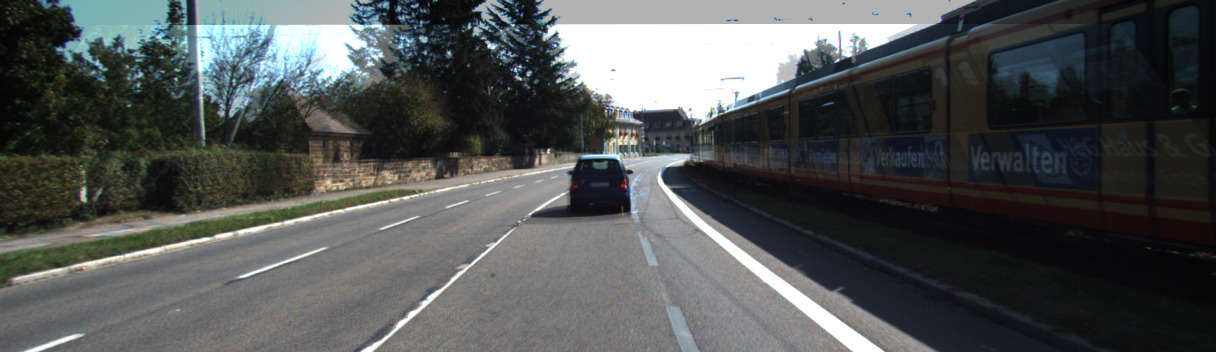}
    \end{minipage}
    
    \begin{minipage}{0.19\linewidth}
    	\centering
    	Baseline
    \end{minipage}
    \begin{minipage}{0.39\linewidth}
    	\centering
    	\includegraphics[width=1\textwidth]{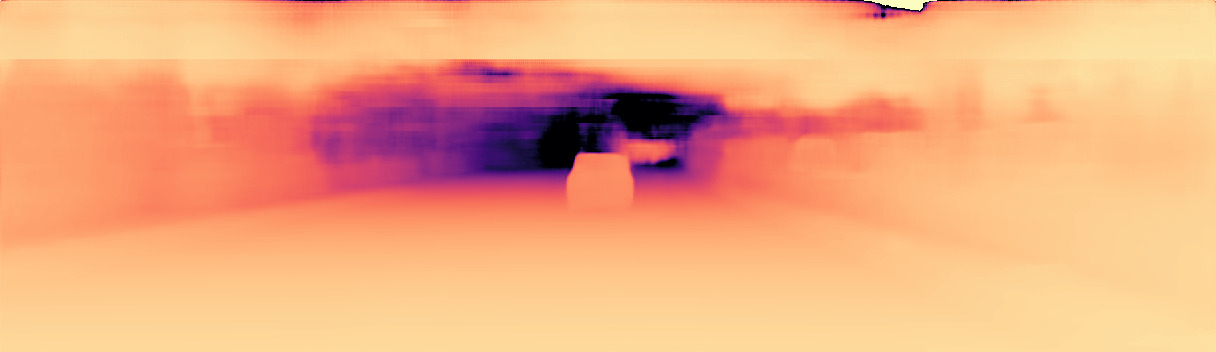}
    \end{minipage}
    \begin{minipage}{0.39\linewidth}
    	\centering
    	\includegraphics[width=1\textwidth]{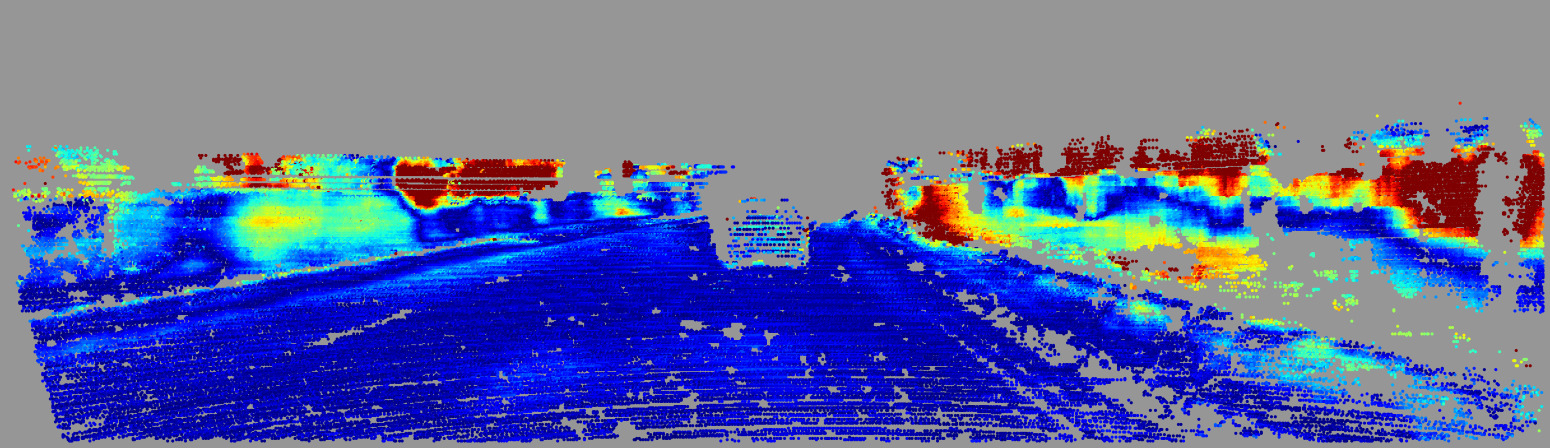}
    \end{minipage}
    
    \begin{minipage}{0.19\linewidth}
    	\centering
    	+FP
    \end{minipage}
    \begin{minipage}{0.39\linewidth}
    	\centering
    	\includegraphics[width=1\textwidth]{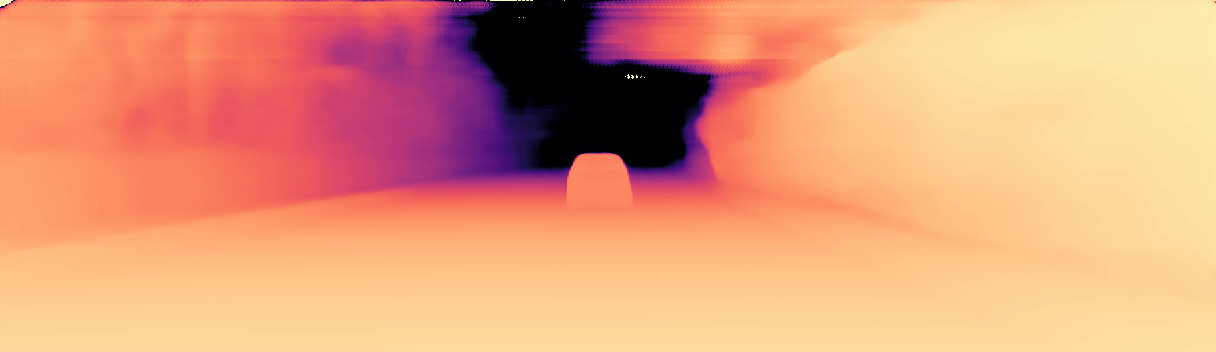}
    \end{minipage}
    \begin{minipage}{0.39\linewidth}
    	\centering
    	\includegraphics[width=1\textwidth]{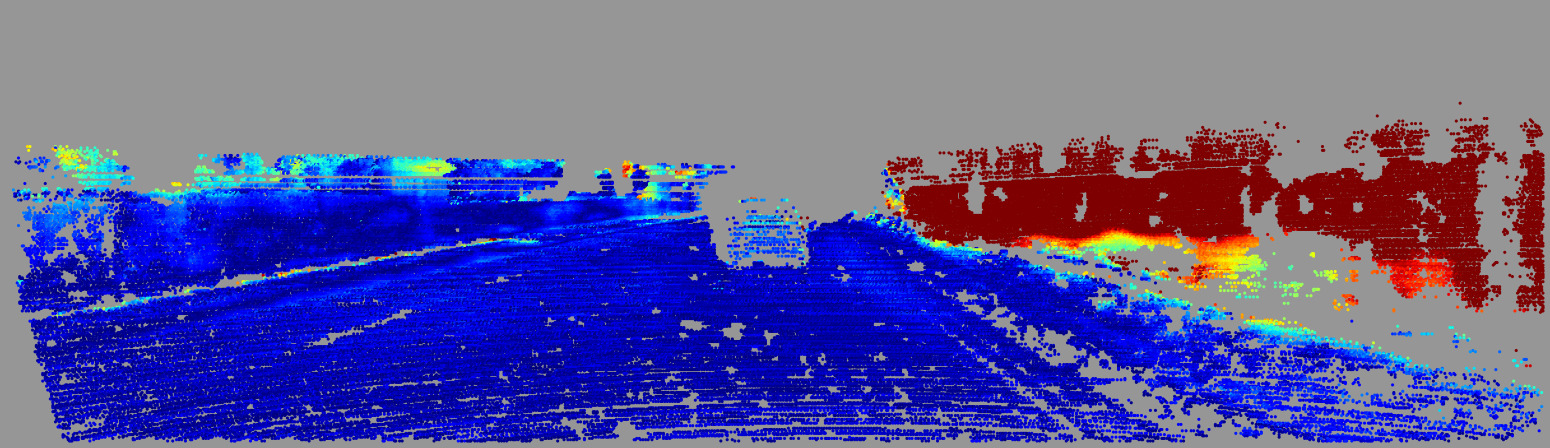}
    \end{minipage}

    \begin{minipage}{0.19\linewidth}
    	\centering
    	Input Image
    \end{minipage}
    \begin{minipage}{0.39\linewidth}
    	\centering
    	\includegraphics[width=1\textwidth]{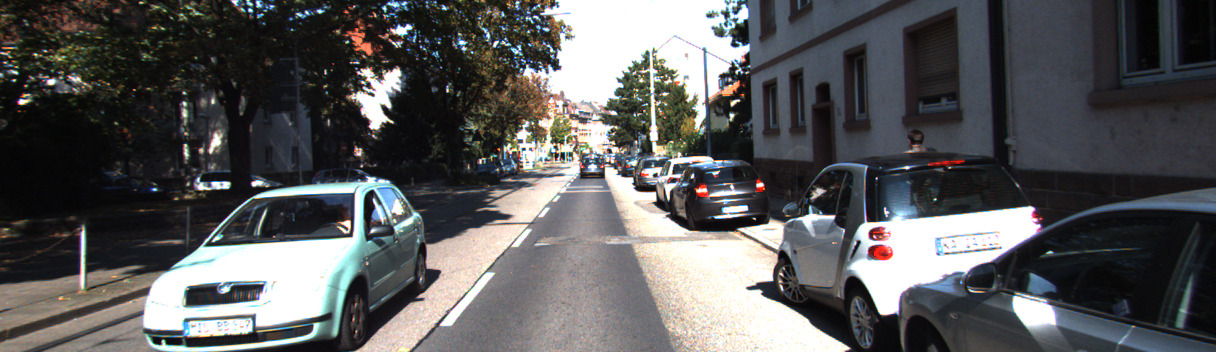}
    \end{minipage}
    \begin{minipage}{0.39\linewidth}
    	\centering
    	\includegraphics[width=1\textwidth]{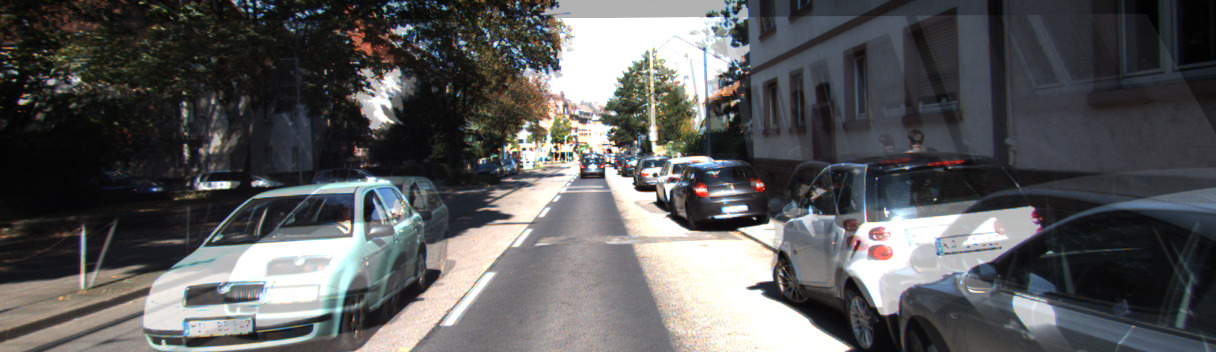}
    \end{minipage}
    
    \begin{minipage}{0.19\linewidth}
    	\centering
    	Baseline
    \end{minipage}
    \begin{minipage}{0.39\linewidth}
    	\centering
    	\includegraphics[width=1\textwidth]{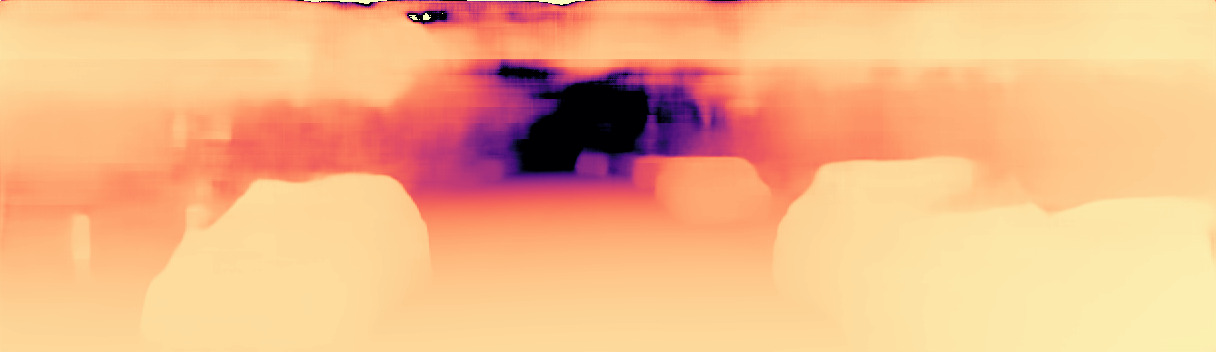}
    \end{minipage}
    \begin{minipage}{0.39\linewidth}
    	\centering
    	\includegraphics[width=1\textwidth]{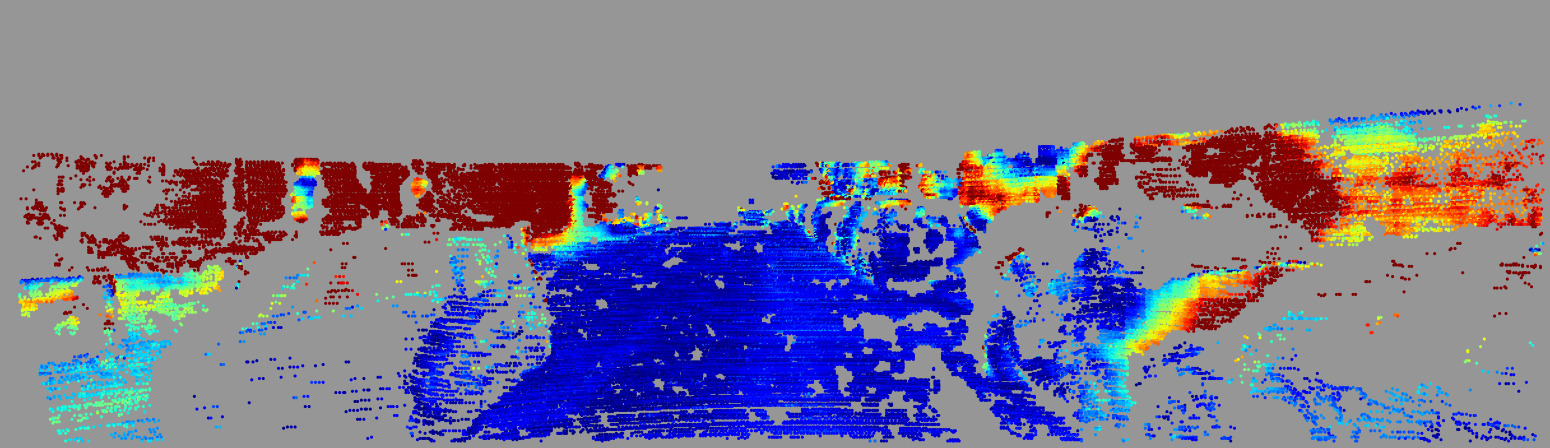}
    \end{minipage}
    
    \begin{minipage}{0.19\linewidth}
    	\centering
    	+FP
    \end{minipage}
    \begin{minipage}{0.39\linewidth}
    	\centering
    	\includegraphics[width=1\textwidth]{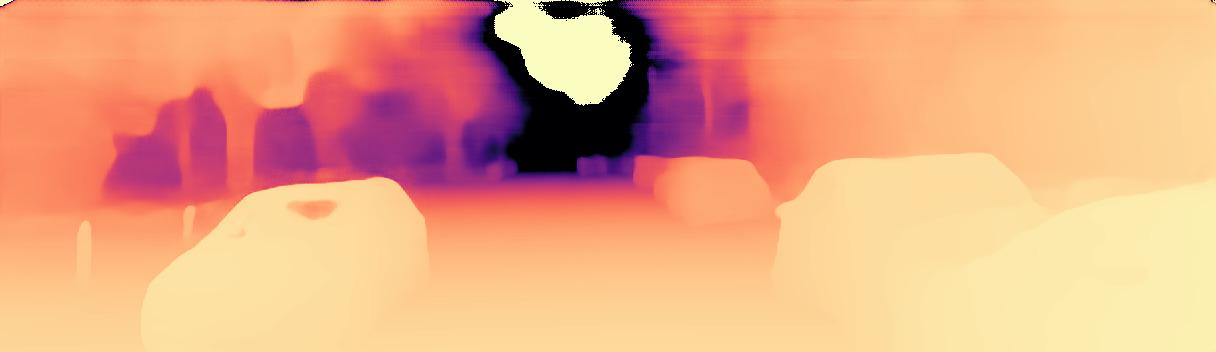}
    \end{minipage}
    \begin{minipage}{0.39\linewidth}
    	\centering
    	\includegraphics[width=1\textwidth]{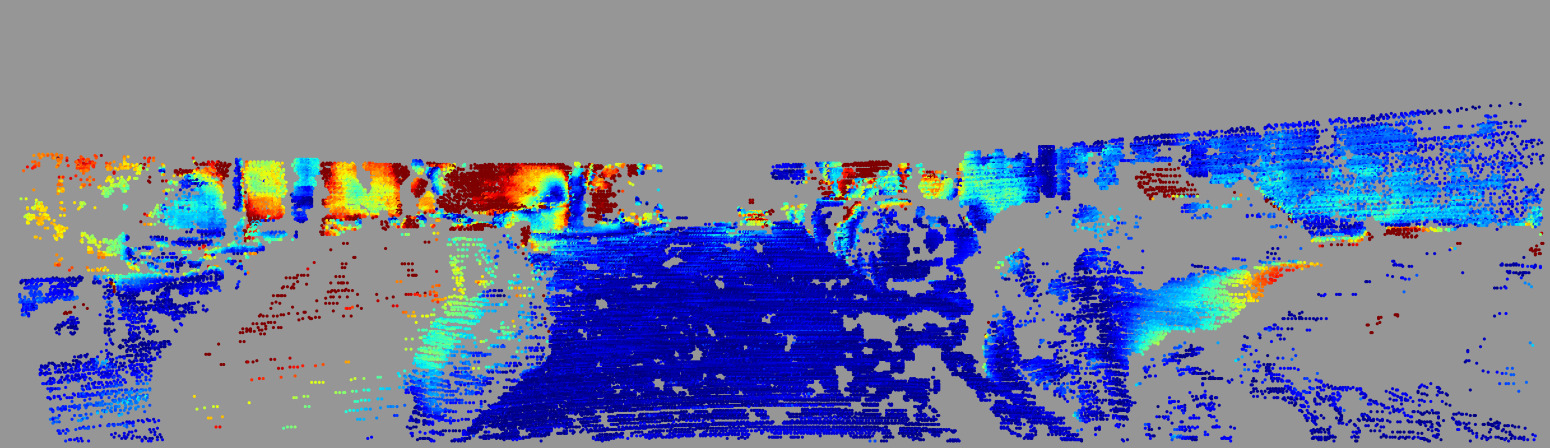}
    \end{minipage}

    \begin{minipage}{0.19\linewidth}
    	\centering
    	Input Image
    \end{minipage}
    \begin{minipage}{0.39\linewidth}
    	\centering
    	\includegraphics[width=1\textwidth]{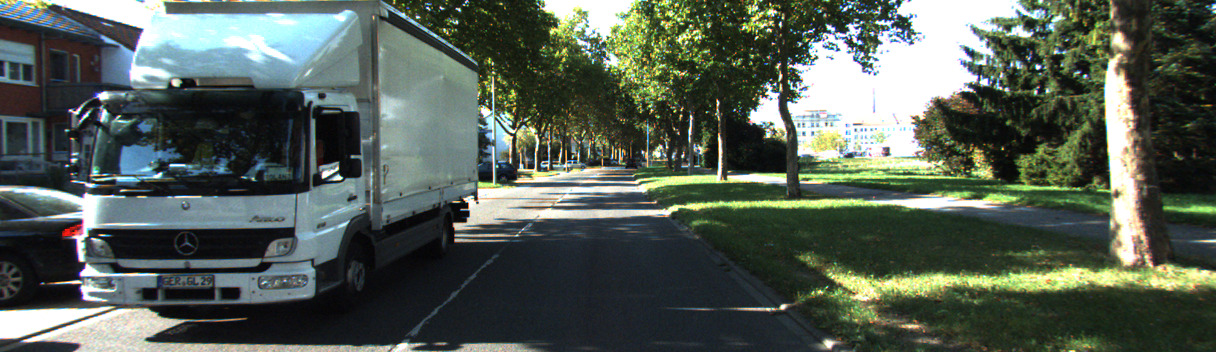}
    \end{minipage}
    \begin{minipage}{0.39\linewidth}
    	\centering
    	\includegraphics[width=1\textwidth]{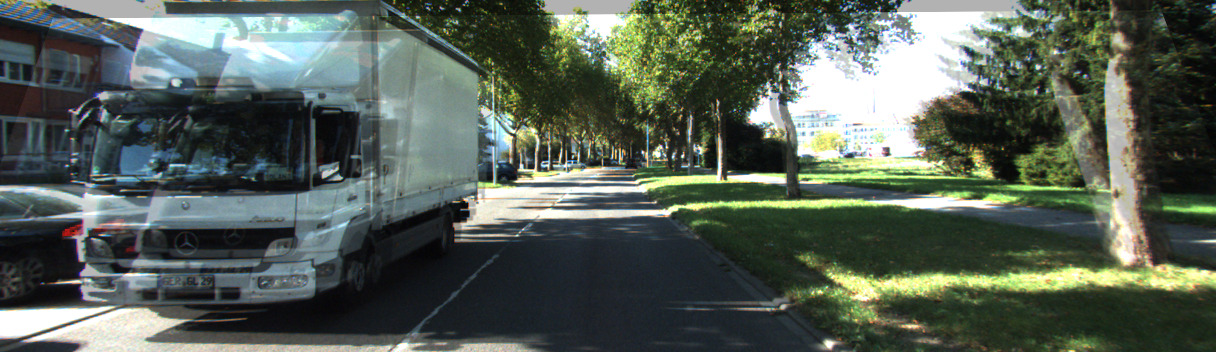}
    \end{minipage}
    
    \begin{minipage}{0.19\linewidth}
    	\centering
    	Baseline
    \end{minipage}
    \begin{minipage}{0.39\linewidth}
    	\centering
    	\includegraphics[width=1\textwidth]{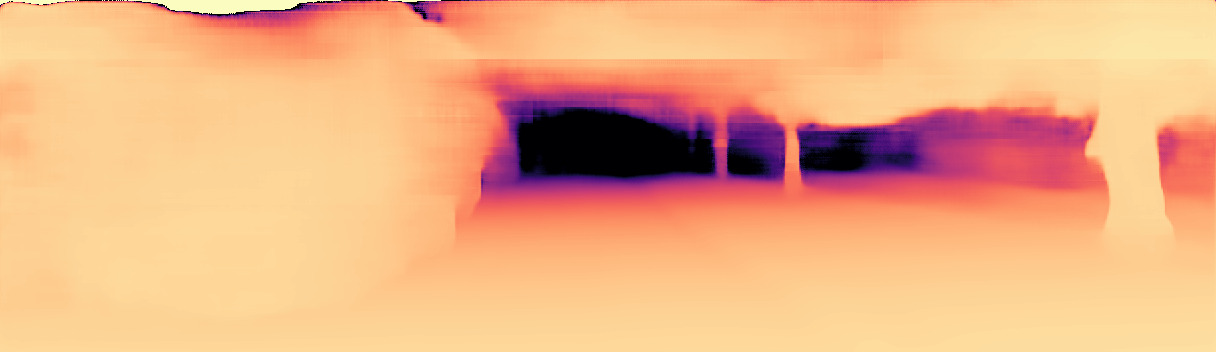}
    \end{minipage}
    \begin{minipage}{0.39\linewidth}
    	\centering
    	\includegraphics[width=1\textwidth]{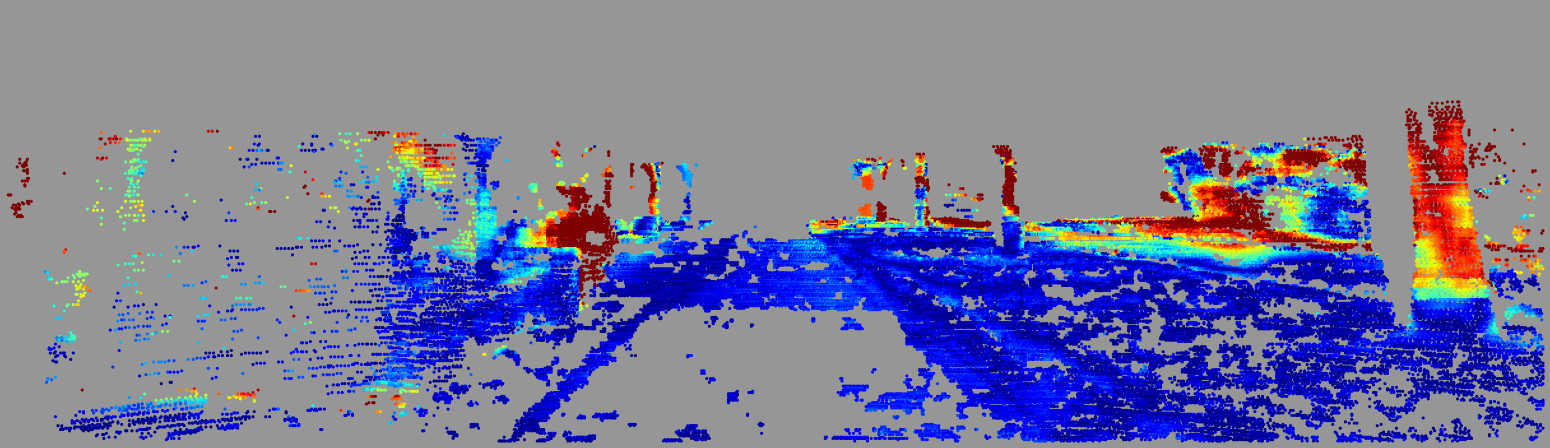}
    \end{minipage}
    
    \begin{minipage}{0.19\linewidth}
    	\centering
    	+FP
    \end{minipage}
    \begin{minipage}{0.39\linewidth}
    	\centering
    	\includegraphics[width=1\textwidth]{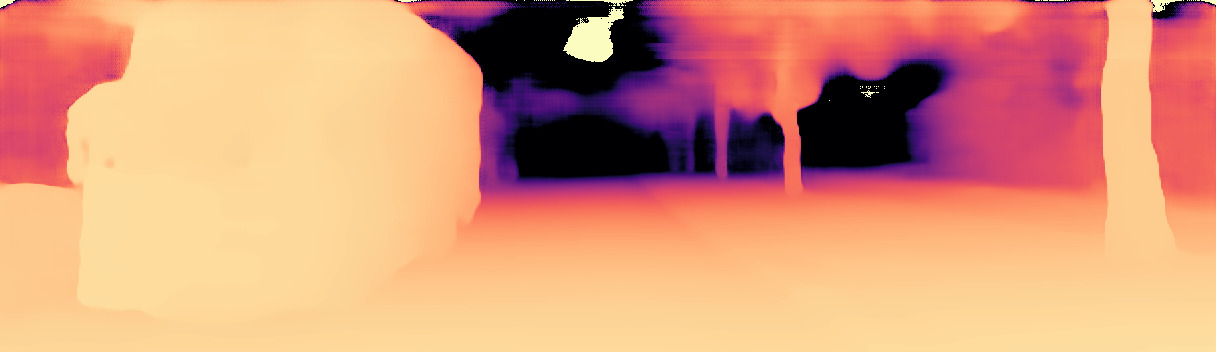}
    \end{minipage}
    \begin{minipage}{0.39\linewidth}
    	\centering
    	\includegraphics[width=1\textwidth]{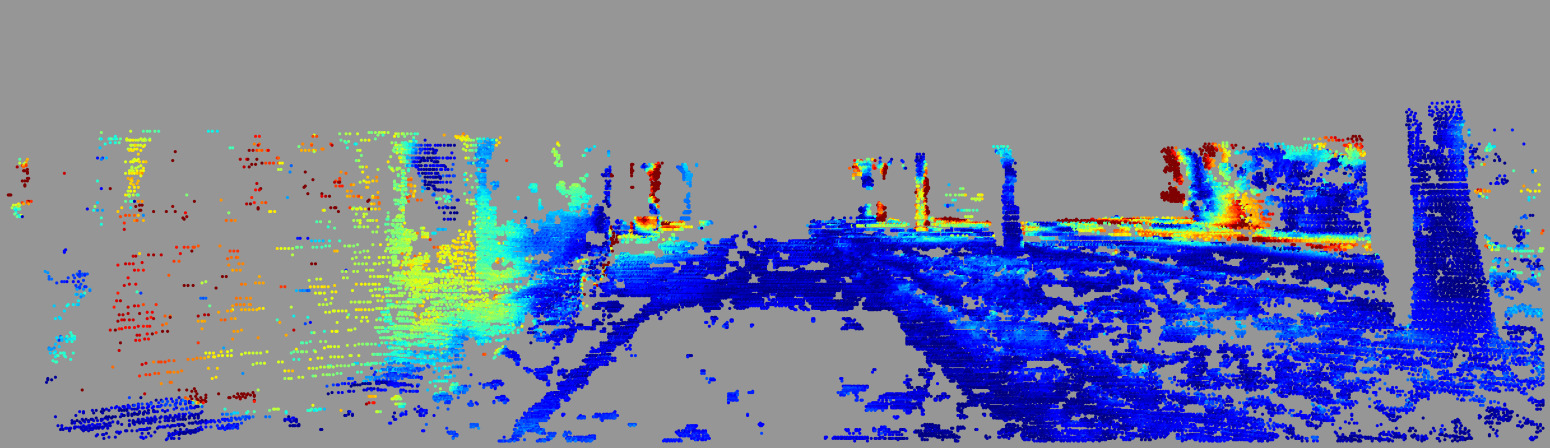}
    \end{minipage}
    
    \caption{Comparison between baseline and +FP. For each sample, the first row shows the target image and the plane-aligned image pairs. The rest rows shows the predicted depth map and the corresponding error map separately.}
    \label{appendix_fig:flow_ablation}
\end{figure*}

In Fig.~\ref{appendix_fig:ablation}, we show an example of how each component affects the result. By adding flow pre-training, the performance is improved significantly on static scenes but worsened on dynamic objects that violate the static assumption in planar parallax geometry. It suggests that, without flow pre-training, models may not take advantage of the consecutive frame. More examples are shown in Fig.\ref{appendix_fig:flow_ablation}. By adding Planar Position Embedding, the unreasonable errors have been restrained. Finally, the single frame branch and depth supervision improves the performance on dynamic objects and leads the error to the final level.

\end{document}